\documentclass[authoryear,a4paper,11pt]{elsarticle_WP}

\newif\ifarxiv
\arxivtrue

\usepackage{clipboard}
\newclipboard{./quotes}
\usepackage{graphicx}
\usepackage{amssymb,amsmath,amsthm}
\usepackage{bm}
\usepackage{color}
\usepackage[margin=1in]{geometry}
\usepackage{natbib}

\usepackage{booktabs}
\usepackage{setspace}

\usepackage{array}
\usepackage[normalem]{ulem}
\usepackage{tikz}

\usepackage{url}

\usepackage{mathtools}
\usepackage{multicol}

\usepackage{framed}
\usepackage[colorlinks,citecolor=blue,linkcolor=blue]{hyperref}
\usepackage{algorithm,algpseudocode}
\usepackage[noabbrev]{cleveref}
\usepackage{pgfplots}
\usepackage{catchfile}
\usetikzlibrary{decorations.pathreplacing}

\usepackage[font=small,aboveskip=5pt,belowskip=0pt,labelfont=bf,format=hang]{caption}
\usepackage[aboveskip=5pt, belowskip=5pt]{subcaption}
\usepackage{float} 
\usepackage{xparse}
\usepackage{xr}

\journal{Spatial Statistics}

\newcolumntype{x}[1]{>{\centering\arraybackslash\hspace{0pt}}p{#1}}

\newcommand{\zerob} {{\bf 0}}

\newcommand{\expect} {{\mathbb{E}}}

\newcommand{\thetab} {{\boldsymbol{\theta}}}
\newcommand{\alphab} {{\boldsymbol{\alpha}}}

\newcommand{\sigmab} {{\boldsymbol{\sigma}}}
\newcommand{\nub} {{\boldsymbol{\nub}}}
\newcommand{\gammab} {{\boldsymbol{\gamma}}}
\newcommand{\mub} {{\boldsymbol{\mu}}}

\newcommand{\Thetab} {{\boldsymbol{\Theta}}}

\newcommand{\lambdab} {\boldsymbol{\lambda}}

\newcommand{\xib} {\boldsymbol{\xi}}
\newcommand{\psib} {\boldsymbol{\psi}}

\newcommand{\Sigmamat} {{\bm \Sigma}}

\newcommand{\Amat} {\textbf{A}}
\newcommand{\Bmat} {\textbf{B}}

\newcommand{\Wmat} {\textbf{W}}

\newcommand{\Cmat} {\mathbf{C}}

\newcommand{\Smat} {\textbf{S}}

\newcommand{\Xvec} {\mathbf{X}}

\newcommand{\bvec} {\textbf{b}}

\newcommand{\svec} {\textbf{s}}

\newcommand{\fvec} {\textbf{f}}
\newcommand{\rvec} {\textbf{r}}
\newcommand{\muvec} {\boldsymbol{\mu}}

\newcommand{\betab} {\boldsymbol {\beta}}
\newcommand{\rhob} {\boldsymbol{\rho}}

\renewcommand{\zerob}{\mathbf{0}}

\newcommand{\x}{\mathbf{x}}

\newcommand{\Yvec}{\mathbf{Y}} 

\newcommand{\Zvec}{\mathbf{Z}}

\newcommand{\cov}{\mathrm{cov}}

\newcommand{\Var}{\mathrm{var}}

\newcommand{\appropto}{\mathrel{\vcenter{
  \offinterlineskip\halign{\hfil$##$\cr
    \propto\cr\noalign{\kern2pt}\sim\cr\noalign{\kern-2pt}}}}}

\DeclareMathOperator*{\argmin}{arg\,min}

\let\originalleft\left
\let\originalright\right
\renewcommand{\left}{\mathopen{}\mathclose\bgroup\originalleft}
\renewcommand{\right}{\aftergroup\egroup\originalright}

\ifarxiv
\else
\fi

\begin{document}

\begin{frontmatter}

\title{\bf Spatial Bayesian Neural Networks}
 
\author[Wollongong]{Andrew Zammit-Mangion\corref{correspondingauthor}}
\cortext[correspondingauthor]{Corresponding author}
\ead{azm@uow.edu.au}
 
\author[Wollongong]{Michael D. Kaminski}
\ead{mdk201@uowmail.edu.au}

\author[Huawei]{Ba-Hien Tran}
\ead{ba.hien.tran@huawei.com}

\author[KAUST]{Maurizio Filippone}
\ead{maurizio.filippone@kaust.edu.sa}

\author[Wollongong]{Noel Cressie}
\ead{ncressie@uow.edu.au}

\address[Wollongong]{School of Mathematics and Applied 
                        Statistics, University of Wollongong, 
                        Australia \\ \vspace{0.1in}}

\address[Huawei]{Paris Research Centre, Huawei Technologies, France \\ \vspace{0.1in}}

\address[KAUST]{Statistics Program, King Abdullah University of Science and Technology, Saudi Arabia}

\begin{abstract}

Statistical models for spatial processes play a central role in analyses of spatial data. Yet, it is the simple, interpretable, and well understood models that are routinely employed even though, as is revealed through prior and posterior predictive checks, these can poorly characterise the spatial heterogeneity in the underlying process of interest. Here, we propose a new, flexible class of spatial-process models, which we refer to as spatial Bayesian neural networks (SBNNs). An SBNN leverages the representational capacity of a Bayesian neural network; it is tailored to a spatial setting by incorporating a spatial ``embedding layer'' into the network and, possibly, spatially-varying network parameters. An SBNN is calibrated by matching its finite-dimensional distribution at locations on a fine gridding of space to that of a target process of interest. That process could be easy to simulate from or we may have many realisations from it. We propose several variants of SBNNs, most of which are able to match the finite-dimensional distribution of the target process at the selected grid better than conventional BNNs of similar complexity. We also show that an SBNN can be used to represent a variety of spatial processes often used in practice, such as Gaussian processes, lognormal processes, and max-stable processes. We briefly discuss the tools that could be used to make inference with SBNNs, and we conclude with a discussion of their advantages and limitations.

\end{abstract}


\end{frontmatter}

\section{Introduction}\label{sec:Intro}

At the core of most statistical analyses of spatial data is a spatial process model. The spatial statistician has a wide array of models to choose from, ranging from the ubiquitous Gaussian-process models \citep[e.g.,][]{rasmussen2006} to the trans-Gaussian class of spatial models \citep[e.g.,][]{deOliveira_1997}, all the way to the more complicated models for spatial extremes \citep[e.g.,][]{Davison_2015}. Each model class is itself rich, which leads to several questions: What class of covariance functions should be used? Should the model account for anisotropy or non-stationarity? Should the model be non-Gaussian? Will the model lead to computationally efficient inferences? There is a long list of modelling decisions and diagnostic checks to be made when undertaking a spatial statistical analysis \citep{Cressie_1993}, and these need to be done every time the spatial statistician is presented with new data. The typical workflow requires one to be cognisant of the large array of models available, paired with computational tools that are themselves varied and complex.

In this paper we take the first steps in a new paradigm for spatial statistical analysis that hinges on the use of a single, highly adaptive, class of spatial process models, irrespective of the nature of the data or the application. We refer to a member of this model class as a spatial Bayesian neural network (SBNN), which we \emph{calibrate} by matching its distribution at locations on a fine gridding of space, to that of a spatial process of interest. An SBNN is parameterised using weights and biases, but the parameterisation is high-dimensional and its form does not change with the context in which it is used. Consequently, the same calibration techniques and algorithms can be used irrespective of the application and the spatial process of interest. The primary purpose of modelling with an SBNN is to relieve the spatial statistician from having to make difficult modelling and computing decisions. At the core of an SBNN is a Bayesian neural network \citep[BNN,][]{neal1996}, that is, a neural network where the network parameters (i.e., the weights and the biases) are endowed with a prior distribution.

Bayesian neural networks are not new to spatial or spatio-temporal statistics; see, for example \citet{McDermott_2019} for their use in the context of spatio-temporal forecasting; \citet{Payares_2023} for their use in neurodegenerative-disease classification from magnetic resonance images; and \citet{Kirkwood_2022} for their use in geochemical mapping. However, their use to date involves models constructed through rather simple prior distributions over the weights and biases: The priors are typically user-specified, fixed, or parameter-invariant. Such a prior distribution, although straightforward to define, will likely lead to a prior spatial process model that is degenerate or, at the very least, highly uncharacteristic of the process of interest; see, for example, the related discussion of \citet{neal1996} in the context of deep neural networks and that of \citet{duvenaud2014} and \citet{Dunlop_2018} in the context of deep Gaussian processes. This degeneracy stems from the fact that neural networks are highly nonlinear, and that the prior distribution on the weights and biases has a complicated, and unpredictable, effect on the properties of the spatial process. While a poor choice of spatial process model may be inconsequential when large quantities of data are available, it is likely to be problematic when data are scarce or reasonably uninformative of the model's parameters.

The core idea that we leverage in this work is the approach of \citet{tran2022}, which we summarise as follows: If one has a sufficiently large number of realisations from a target process of interest, then one can calibrate a BNN such that its finite-dimensional distributions closely follow those of the underlying target process. Once this calibration is done, inferential methods for finding the posterior distribution of the parameters in the BNN can be used, after new data are observed; these methods generally involve stochastic gradient Hamiltonian Monte Carlo \citep[SGHMC,][]{chen2014} or variational Bayes \citep[e.g.,][]{Graves_2011,zammit2021}. This paradigm is attractive: from a modelling perspective, one does not need to look for an appropriate class of spatial process models every time a new analysis is needed, and the same inferential tool can be used irrespective of the data being considered. From a deep-learning perspective, calibration avoids the pathologies that one risks when applying BNNs with fixed priors to spatial or spatio-temporal analyses. The method of \citet{tran2022} is novel in its approach of enabling computationally- and memory-efficient calibration via a Monte Carlo approach. However, the BNNs they consider are not directly applicable to spatial problems: A novelty of our SBNNs is that they incorporate a spatial embedding layer and spatially-varying network parameters, which we show help capture spatial covariances and non-stationary behaviour of the underlying target spatial process. Specifically, we show through simulation studies that several such SBNNs are able to match a selected high- and finite-dimensional distribution of the target spatial process better than conventional BNNs of similar complexity.

In Section \ref{sec:background} we motivate and construct SBNNs, while in Section~\ref{sec:calibration} we detail an approach for calibrating SBNNs that follows closely that of \citet{tran2022}. In Section~\ref{sec:results}, we show that our SBNNs can be used as high-quality surrogate models for both stationary and highly non-stationary Gaussian processes, as well as lognormal spatial processes. In Section~\ref{sec:inference} we outline how SBNNs can be used in practice: as stochastic generators and for making inferences. Finally, in Section~\ref{sec:conclusion} we conclude with a discussion of the benefits and limitations of SBNNs, and also outline what further challenges need to be overcome before they can be applied in a real-data setting.

\section{Methodology}\label{sec:background}

\subsection{Bayesian neural networks for spatial data}

A BNN \citep[][pp.\ 10--19]{neal1996} is a composition of nonlinear random functions. Each function comprises a so-called `layer' of the network. In the context of spatial data analysis, a BNN is used to model a spatial process $Y(\cdot)$ over a spatial domain $D \subset \mathbb{R}^{d}$, where $\svec \in D$ is the input to the BNN and the spatial dimension $d$ is small; typically, $d \in \{1,2, 3\}$. We define a BNN as follows:
\begin{align}\label{eq:composition}
Y(\svec) &\equiv f(\svec; \thetab) = f_L (\fvec_{L-1} (\cdots \fvec_1(\fvec_0(\svec; \theta_0); \thetab_1) \cdots; \thetab_{L-1}); \thetab_L), \quad \svec \in D,
\end{align}

\noindent where, for $l = 0,\dots, L$, $\fvec_l(\cdot\,; \thetab_l) \in \mathbb{R}^{d_l}$, $d_l$ is the output dimension of the $l$-th layer, for the $L$-th layer $d_L = 1$, and the output is $Y(\cdot)$. We parameterise each function $f_l(\cdot\,; \thetab_l)$  by $\thetab_l$, for $l = 0,\dots,L$. We will assume that $\theta_0$ is known by design, and we collect the remaining parameters across the whole BNN into ${\bm{\theta}\equiv\{\bm{\theta}_l: l=1,\dots,L\}}$. For what we refer to as a \emph{vanilla} BNN, we let $\fvec_{0}(\svec; \theta_0) \equiv \svec,$ for $\svec \in D$ (so that $d_0 = d$), and we let the remaining layers have the form,

\begin{equation}
  \mathbf{f}_l(\cdot\,; \thetab_l) = \frac{1}{\sqrt{d_{l-1}}} \mathbf{W}_l \bm{\varphi}_{l-1}(\cdot) + \mathbf{b}_l,
  \qquad l=1,\ldots, L,
  \label{eq:forwardpass}
\end{equation}

\noindent where $\thetab_l\equiv\{\thetab_l^w, \thetab_l^b\}$; $\thetab_l^w \equiv \textrm{vec}(\Wmat_l)$; $\thetab_l^b \equiv \bvec_l$; $\Wmat_l \in\mathbb{R}^{d_{l}\times d_{l-1}}$ and $\bvec_l \in\mathbb{R}^{d_{l}}$ are random weights and biases associated with the $l$-th layer, respectively, to which prior distributions will be assigned; and $\textrm{vec}(\cdot)$ is the vectorisation operator. Further, in \eqref{eq:forwardpass}, $\bm{\varphi}_{l-1}(\cdot)\in\mathbb{R}^{d_{l-1}}$ are so-called activation functions that are evaluated element-wise over the output of layer $l-1$; popular activation functions include the hyperbolic tangent, the rectified linear unit (ReLU), and the softplus functions \citep{Goodfellow_2016}. Finally, the division by $\sqrt{d_{l-1}}$ in \eqref{eq:forwardpass} gives rise to the neural-tangent-kernel (NTK) parameterisation; see \citet{jacot2018} for its motivation and further details.

\begin{figure}[t!]

    \begin{center}
    
        \includegraphics{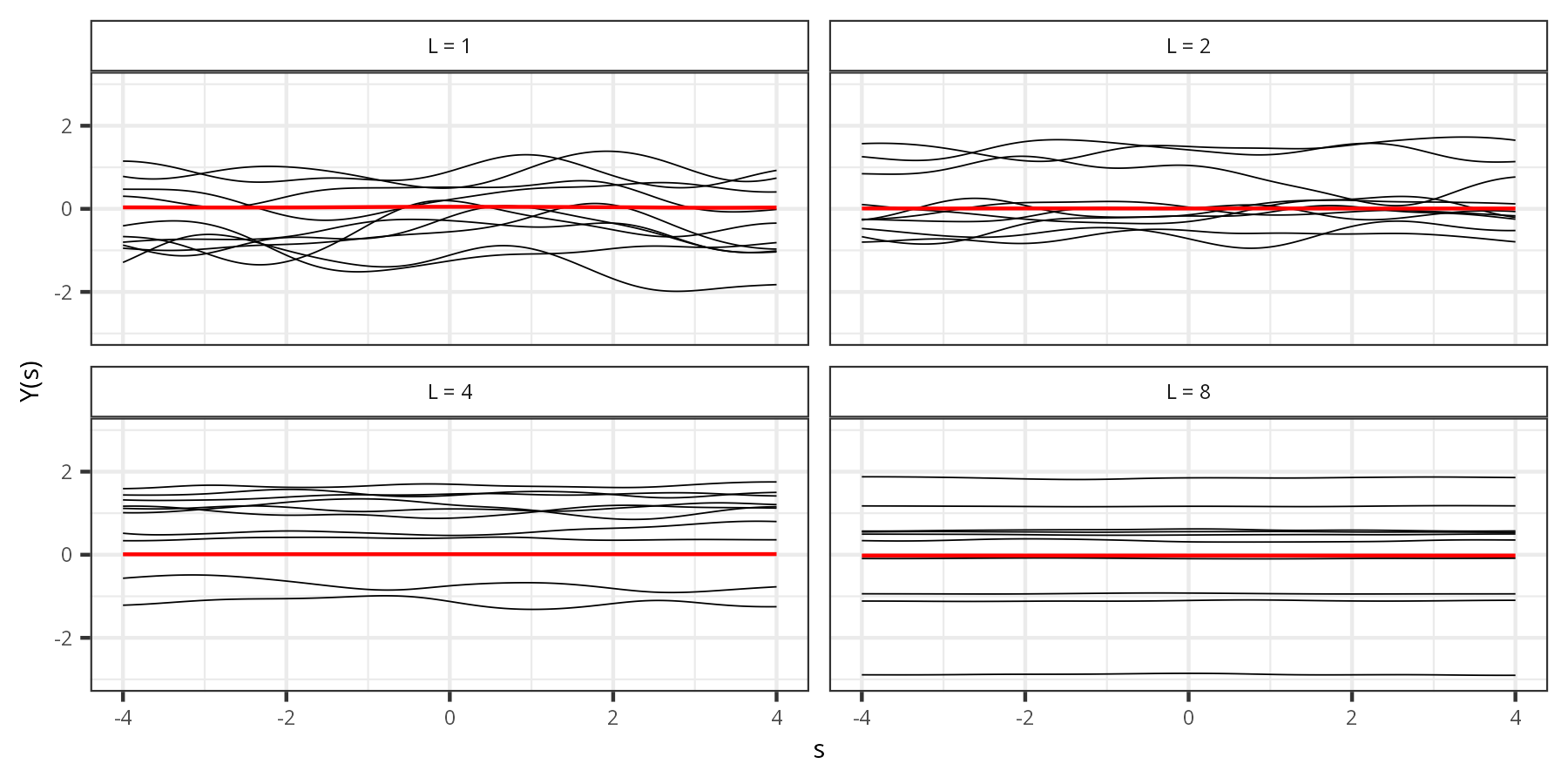}
    
    \end{center}
        \caption{Sample paths (black) of the vanilla Bayesian neural network of the form \eqref{eq:composition} and \eqref{eq:forwardpass} with $\fvec_{0}(\svec; \theta_0) \equiv \svec$, where here each hidden layer is of dimension $d_l  = 40,\, l = 1,\dots,(L-1)$, the activation functions in $\bm{\varphi}_{l-1}(\cdot), l = 1,\dots,L,$ are $\textrm{tanh}(\cdot)$ functions,  $D \equiv [-4,4]$, and  the weights and biases collected in $\thetab$ are all drawn independently from a standard normal distribution with zero mean and unit variance. The empirical-process mean computed from 4000 sample paths is also shown (red). (Top-left panel) $L = 1$. (Top-right panel) $L = 2$. (Bottom-left panel) $L = 4$. (Bottom-right panel) $L = 8$.}
        \label{fig:bnn_pathology}
      \end{figure}

The finite-dimensional distributions of the process $Y(\cdot)$ are fully determined by the prior distribution of the weights and biases that define $\thetab$. A natural question is then: What distribution should we have for the weights and biases? There is no straightforward answer to this question, in large part because of the nonlinearities inherent to BNNs, which make the relationship between the prior distribution of $\thetab$ and that of the process $Y(\cdot)$ unintuitive. Yet, the choice is important: For example, one might model the weights and biases as independent and equip them with a $\textrm{Gau}(0,1)$ distribution, but this choice leads to a seemingly degenerate stochastic process. For illustration, \Cref{fig:bnn_pathology} shows sample paths drawn from the process $Y(\cdot)$ over $D \equiv [-4, 4]$ when all weights and biases are simulated independently from a $\textrm{Gau}(0,1)$ distribution. Observe that, as the number of layers increases from $L=1$ (top-left panel) to $L=8$ (bottom-right panel), sample paths from the process tend to flatten out over $\svec \in D$. Clearly, the stochastic process with $L=8$ is an unreasonable model for applications involving spatial data.

In Section~\ref{sec:calibration} we show how one may \emph{calibrate} these prior distributions (i.e., estimate the hyper-parameters that parameterise the prior means and variances) so that a selected high- and finite-dimensional distribution from $Y(\cdot)$ -- for example one at locations on a fine gridding of $D$ -- matches closely that of another, user-specified, spatial process. As is shown in Section~\ref{sec:results}, we had difficulty calibrating the vanilla BNN given by \eqref{eq:composition} to spatial models that are routinely employed (e.g., the Gaussian process), in the sense that we were unable to obtain a BNN with a finite-dimensional distribution that was close to that of the target spatial model. This was likely because the vanilla BNN in \eqref{eq:composition} is not tailored for spatial data. Our SBNNs modify \eqref{eq:composition} in two ways that specifically aim at modelling spatial dependence. These modifications, which are discussed in the following subsections, lead to a class of SBNNs that can match target spatial processes more closely than BNNs of similar complexity.

\subsection{Spatial Bayesian neural networks}\label{sec:SBNN}

As shown below in Section~\ref{sec:results}, we often found difficulty capturing spatial covariances when using vanilla BNNs (i.e., \eqref{eq:composition} and \eqref{eq:forwardpass} with $\fvec_{0}(\svec; \theta_0) \equiv \svec,~ \svec \in D$). This corroborates the findings of \citet{chen2022}, who argue that classical neural networks cannot easily incorporate spatial dependence between the inputs, when employed for spatial prediction. In their paper, they mitigated this issue by using a set of spatial basis functions in the first layer of the network, in a process they called \emph{deepKriging}. We also found that the inclusion of this ``embedding layer'' greatly improved the ability of SBNNs to express realistic covariances (Section~\ref{sec:embedding}). However, we found that, even with the embedding layer, our SBNNs were prone to not capturing complex non-stationary behaviour. To deal with this, we made the parameters appearing in our SBNNs spatially varying (Section~\ref{sec:varying-weights}).

\subsubsection{The embedding layer in an SBNN}\label{sec:embedding}

The first modification we make to the classical BNN is to follow \citet{chen2022} and replace the zeroth layer in \eqref{eq:composition} with an embedding layer comprising a set of $K$ radial basis functions \citep[RBF, ][]{Buhmann_2003} centred on a regularly-spaced grid  in $D$. Specifically, we define $\fvec_0(\svec;\theta_0) \equiv \rhob(\svec; \tau) \equiv (\rho_1(\svec;\tau),\ldots,\rho_{K}(\svec;\tau))',$ where $\theta_0 = \tau$, $\rho_k(\cdot\,;\tau),~k = 1,\dots,K,$ are $K$ radial basis functions with centroids $\{\xib_k \in D : k = 1,\dots, K\}$ that together make up the spatial embedding layer, and $\tau > 0$ is a length-scale parameter. Here, we use the Gaussian bell-shaped radial basis functions; specifically, we let 
\begin{equation}
    \rho_k(\svec;\tau) = \exp\left[-\left(\frac{\|\svec - \xib_k\|}{\tau}\right)^2\right],\quad k = 1,\dots,K,
    \label{eq:rbf}
\end{equation}
\noindent where $\|\cdot\|$ is the Euclidean norm. We set $\tau$ so that the radial basis functions have a suitable level of overlap, as is often done in low-rank spatial modelling \citep[e.g.,][]{cressie2008, nychka2018,zammitmangion2018}; see Figure~\ref{fig:basis_functions} in the Supplementary Material for an example. 
Given a spatial location $\mathbf{s}\in\mathbb{R}^{d}$, the output of the embedding layer is given by $\bm{\rho}(\svec;\tau) \in\mathbb{R}^{d_0}$ with dimension $d_0 = K$. The spatial embedding layer therefore represents a vector $\bm{\rho}(\svec;\tau)\in(0,1]^{K}$, indexed by spatial location $\svec$, which encodes the proximity of the input  $\svec$ to each RBF centroid. A value for $\rho_k(\svec;\tau)$ close to one indicates proximity of the point $\svec$ to the centroid $\xib_k$ and, as $\svec$ moves further away from $\xib_k$, its value decreases rapidly. If the weights and biases of an SBNN are drawn from a prior distribution that is spatially invariant, we refer to the resulting network as an SBNN with spatially-invariant parameters (SBNN-I).

\subsubsection{Spatially varying network parameters}\label{sec:varying-weights}

In Section~\ref{sec:results} we show that the embedding layer is generally important for an SBNN to model covariances, but we also show that to model non-stationarity one needs something more. One natural way to introduce additional flexibility to capture spatially heterogeneous behaviour better, is to change the weights and biases of the SBNN-I to make them spatially varying. We do this by defining the distribution for each weight and bias as Gaussian with some mean and variance that both vary smoothly over $D$. As a practical matter, we employ the same basis used in the embedding layer, $\bm{\rho}(\mathbf{\cdot}; \tau)$, to model the smoothly-varying means and variances.

For ease of notation, consider a single, now spatially-varying, weight or bias parameter $\theta(\cdot)$. We model the prior mean of $\theta(\cdot)$ as
\begin{equation}
    \mu(\mathbf{\cdot}) = \rho_1(\mathbf{\cdot}) \alpha_1 + \cdots + \rho_K(\mathbf{\cdot}) \alpha_{K},
    \label{eq:mean_ns}
\end{equation}
where $\alpha_k\in\mathbb{R}$ for $k = 1,\dots,K$, and the prior standard deviation as
\begin{equation}
    \sigma(\mathbf{\cdot}) = \mathrm{softplus}(\rho_1(\cdot) \beta_1 + \cdots + \rho_{K}(\cdot) \beta_{K}),
    \label{eq:stdev_ns}
\end{equation}
where $\beta_k\in\mathbb{R}$ for $k = 1,\dots,K$ and $\mathrm{softplus}:t\mapsto \ln(1+e^t)$ enforces positivity. It remains then to establish the spatial covariance of $\theta(\cdot)$  which, for $\eta \sim \textrm{Gau}(0,1)$, we define as follows:
\begin{equation}\label{eq:generative_theta}
    \theta(\cdot) = \mu(\cdot) + \sigma(\cdot)\eta.
\end{equation}
Under this model, for any two locations $\svec \in D$ and $\rvec \in D$, $\textrm{cov}(\theta(\svec),\theta(\rvec)) = \sigma(\svec)\sigma(\rvec)$ and hence $\textrm{corr}(\theta(\svec),\theta(\rvec)) = 1$. This is a rather inflexible prior (spatial) model, but it comes with the advantage of not introducing additional covariance hyper-parameters that would otherwise need to be estimated. All smoothness in $\theta(\cdot)$ for a given weight or bias is induced by that in its $\mu(\cdot)$ and $\sigma(\cdot)$, which are smooth by construction. Since there are many weights and biases (i.e., many $\theta(\cdot)$'s), there are many $\eta$'s (one for each weight and bias), which we model as mutually independent.  Rather than estimating scalar means and standard deviations when calibrating, we instead estimate the coefficients $\alpha_k$ and $\beta_k$, for $k= 1,\dots,K$, for each weight and bias parameter, or groups thereof. We refer to the resulting network as an SBNN with spatially-varying parameters (SBNN-V).

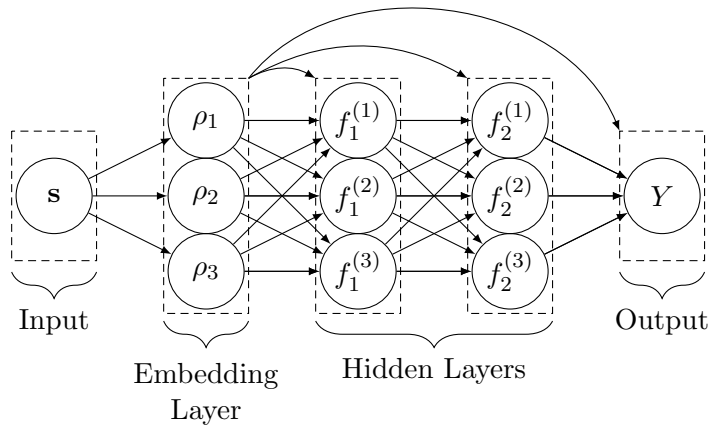
\begin{figure}
    \begin{center}
    \begin{tikzpicture}[>=latex]

    \node[circle, draw, minimum size=1cm, inner sep=0pt] (x) at (0,0) {$\svec$};
    
    \node[circle, draw, minimum size=1cm, inner sep=0pt] (rho1) at (2,1) {$\rho_1$};
    \node[circle, draw, minimum size=1cm, inner sep=0pt] (rho2) at (2,0) {$\rho_2$};
    \node[circle, draw, minimum size=1cm, inner sep=0pt] (rho3) at (2,-1) {$\rho_3$};
    
    \node[circle, draw, minimum size=1cm, inner sep=0pt] (f11) at (4,1) {$f_1^{(1)}$};
    \node[circle, draw, minimum size=1cm, inner sep=0pt] (f12) at (4,0) {$f_1^{(2)}$};
    \node[circle, draw, minimum size=1cm, inner sep=0pt] (f13) at (4,-1) {$f_1^{(3)}$};
    
    \node[circle, draw, minimum size=1cm, inner sep=0pt] (f21) at (6,1) {$f_2^{(1)}$};
    \node[circle, draw, minimum size=1cm, inner sep=0pt] (f22) at (6,0) {$f_2^{(2)}$};
    \node[circle, draw, minimum size=1cm, inner sep=0pt] (f23) at (6,-1) {$f_2^{(3)}$};
    
    \node[circle, draw, minimum size=1cm, inner sep=0pt] (f3) at (8,0) {$Y$};

    \draw[densely dashed] ([shift={(-0.2,0.5)}]x.north west) rectangle ([shift={(0.2,-0.5)}]x.south east);
    \draw[densely dashed] ([shift={(-0.2,0.2)}]rho1.north west) rectangle ([shift={(0.2,-0.2)}]rho3.south east);
    \draw[densely dashed] ([shift={(-0.2,0.2)}]f11.north west) rectangle ([shift={(0.2,-0.2)}]f13.south east);
    \draw[densely dashed] ([shift={(-0.2,0.2)}]f21.north west) rectangle ([shift={(0.2,-0.2)}]f23.south east);
    \draw[densely dashed] ([shift={(-0.2,0.5)}]f3.north west) rectangle ([shift={(0.2,-0.5)}]f3.south east);

    \foreach \i in {1,2,3} {
        \draw[->] (x) to (rho\i);
        \draw[->] (rho\i) to (f1\i);
        \draw[->] (f1\i) to (f2\i);
        \draw[->] (f2\i) to (f3);
    }

    \foreach \i in {1,2,3} {
        \foreach \j in {1,2,3} {
            \draw[->] (rho\i) to (f1\j);
        }
    }

    \foreach \i in {1,2,3} {
        \foreach \j in {1,2,3} {
            \draw[->] (f1\i) to (f2\j);
        }
    }

    \foreach \i in {1,2,3} {
        \draw[->] (f2\i) to (f3);
    }

    \coordinate (rhoboxNE) at ([shift={(0.2,0.2)}]rho1.north east);
    \coordinate (f11boxNW) at ([shift={(-0.2,0.2)}]f11.north west);
    \coordinate (f21boxNW) at ([shift={(-0.2,0.2)}]f21.north west);
    \coordinate (f3boxNW) at ([shift={(-0.2,0.5)}]f3.north west);
    
    \draw[->, bend left] (rhoboxNE) to (f11boxNW);
    \draw[->, bend left=30] (rhoboxNE) to (f21boxNW);
    \draw[->, bend left=60] (rhoboxNE) to (f3boxNW);

    \draw[decorate,decoration={brace,amplitude=10pt,mirror}] 
    ([shift={(-0.2,-0.65)}]x.south west) -- ([shift={(0.2,-0.65)}]x.south east) 
    node[midway,below=10pt] {Input};

    \draw[decorate,decoration={brace,amplitude=10pt,mirror}] 
    ([shift={(-0.2,-0.3)}]rho3.south west) -- ([shift={(0.2,-0.3)}]rho3.south east) 
    node[midway,below=10pt] {\begin{tabular}{c} Embedding \\ Layer \end{tabular}};

    \draw[decorate,decoration={brace,amplitude=10pt,mirror}] 
    ([shift={(-0.2,-0.3)}]f13.south west) -- ([shift={(0.2,-0.3)}]f23.south east) 
    node[midway,below=10pt] {Hidden Layers};

    \draw[decorate,decoration={brace,amplitude=10pt,mirror}] 
    ([shift={(-0.2,-0.65)}]f3.south west) -- ([shift={(0.2,-0.65)}]f3.south east) 
    node[midway,below=10pt] {Output};

    \end{tikzpicture}
    \end{center}
    \caption{Schematic of an SBNN with $L = 3$ layers, and with spatially-varying parameters (SBNN-V). The SBNN-V contains an embedding layer and skip connections that feed the output of the embedding layer into each subsequent layer of the network.}
        \label{fig:augmented_architecture}
    \end{figure}

We illustrate the architecture of the SBNN-V in \Cref{fig:augmented_architecture}.
Note that by setting the means and standard deviations of the neural-network parameters to be of the form \eqref{eq:mean_ns} and \eqref{eq:stdev_ns}, we are introducing so-called \emph{skip connections} into the network architecture, which feed the output of the embedding layer $\bm{\rho}(\cdot\,; \tau)$ directly into each subsequent layer. This use of skip connections is similar to the way they are used in the popular architecture \emph{ResNet} for feature re-usability \citep{He_2016}. This SBNN-V can also be viewed as a simple \emph{hyper-network}, since the prior means and standard deviations of the weights and biases are themselves the outputs of shallow (one-layer) networks \citep[e.g.,][]{Malinin_2020}.
\\

\subsubsection{Model specification of BNNs and SBNNs}\label{sec:architecture}

An SBNN comprises many (potentially several thousand) weights and biases which, as is typical with BNNs, we model as mutually independent. However, a choice must be made on whether to equip all of these independent parameters with different prior distributions, or whether to assume that the parameters are independent but identically distributed within groups. In this work we will consider both options. The former `prior-per-parameter' scheme comes with the advantage that it leads to a highly flexible SBNN, with the downside that many hyper-parameters need to be stored and estimated during calibration. For the latter option we will group the parameters by layer in what we refer to as a `prior-per-layer' scheme \citep[note that other groupings are possible; see, for example][]{Mackay_1992}. The `prior-per-layer' scheme comes with the advantage that the number of hyper-parameters that need to be estimated during calibration is much smaller; an SBNN constructed under this scheme is therefore easier and faster to calibrate. However, we find in Section~\ref{sec:results} that while this `prior-per-layer' scheme is sufficiently flexible to model some stochastic processes of interest, there are cases where SBNNs with a `prior-per-parameter' scheme may calibrate better to the target process. In what follows we detail our SBNNs under both schemes, the `prior-per-layer' scheme (``SBNN-IL'' and ``SBNN-VL'') and the `prior-per-parameter' scheme (``SBNN-IP'' and ``SBNN-VP''). For completeness we also outline the vanilla BNN variants, whose parameters are spatially invariant by definition (``BNN-IL'' and ``BNN-IP''). \\

\noindent {\bf SBNN-IL:} The SBNN-I with a `prior-per-layer' scheme (SBNN-IL) is given by the following hierarchical spatial statistical model:

\begin{minipage}{\textwidth}

\vspace{0.1in}

 For $\svec \in D,~ l = 1,\dots, L$,  
    \begin{align*}
        Y(\svec) \equiv f(\svec; \thetab) &= f_L (\fvec_{L-1} (\cdots \fvec_1(\fvec_0(\svec; \theta_0); \thetab_1) \cdots; \thetab_{L-1}); \thetab_L),\\
        \mathbf{f}_l(\cdot\,; \thetab_l) &= \frac{1}{\sqrt{d_{l-1}}} \mathbf{W}_l \bm{\varphi}_{l-1}(\cdot) + \mathbf{b}_l, \\
        \mathbf{f}_0(\svec; \theta_0) &= \rhob(\svec;\tau) \equiv (\rho_1(\svec;\tau),\ldots,\rho_{K}(\svec;\tau))',
      \end{align*}
      
      \vspace{-0.4in}
      
      \begin{multicols}{2}  
        \begin{align*}
            \theta^w_{l,i} & \sim \textrm{Gau}(\mu^w_l, (\sigma^w_l)^2), \quad i = 1,\dots, n^w_l,\\ 
            \sigma^w_l &= \textrm{softplus}(\gamma^w_l), 
        \end{align*}

        \columnbreak  
        
        \begin{align}
            \theta^b_{l,i} & \sim \textrm{Gau}(\mu^b_l, (\sigma^b_l)^2), \quad i = 1,\dots, n^b_l,\label{eq:theta_SBNN-IL}\\ 
            \sigma^b_l &= \textrm{softplus}(\gamma^b_l), \nonumber
        \end{align}
    \end{multicols}
    \end{minipage}

\vspace{0.2in}

\noindent where $\theta_0 = \tau$. In \eqref{eq:theta_SBNN-IL}, $\thetab_l^w \equiv (\theta_{l,1}^w,\dots,\theta_{l,n_l^w}^w)' = \textrm{vec}(\Wmat_l)$ are the weight parameters at layer $l$; $\mu^w_l \in \mathbb{R}$ and $\sigma^w_l\in \mathbb{R}^+$  are the prior mean and the prior standard deviation of the weights at layer $l$; $\gamma_l^w \in \mathbb{R}$; and $n_l^w$ is the number of weights in layer $l$. The superscript $b$ replaces the superscript $w$ to denote the corresponding quantities for the bias parameters $\thetab_l^b \equiv (\theta_{l,1}^b,\dots,\theta_{l,n_l^b}^b)' = \bvec_l$. The hyper-parameters that need to be estimated when calibrating the SBNN-IL are ${\psib \equiv \{\{\mu_l^w, \mu_l^b, \gamma_l^w, \gamma_l^b\}: l = 1,\dots,L\}}$. The spatial process $Y(\cdot)$ is the output of the final layer. 

\vspace{0.2in}

\noindent {\bf SBNN-IP:} For the SBNN-I with a `prior-per-parameter' scheme (SBNN-IP), the weights and biases are instead distributed as follows:

\begin{multicols}{2}  
    \textrm{For $l = 1,\dots,L$, $ i = 1,\dots, n^w_l$,}
    \begin{align*}
        \theta^w_{l,i} & \sim \textrm{Gau}(\mu^w_{l,i}, (\sigma^w_{l,i})^2),\\ 
        \sigma^w_{l,i} &= \textrm{softplus}(\gamma^w_{l,i}),
    \end{align*}

    \columnbreak  
    \textrm{For $l = 1,\dots,L$, $ i = 1,\dots, n^b_l$,}
    \begin{align}
        \theta^b_{l,i} & \sim \textrm{Gau}(\mu^b_{l,i}, (\sigma^b_{l,i})^2),\label{eq:theta_SBNN-IP}\\ 
        \sigma^b_{l,i} &= \textrm{softplus}(\gamma^b_{l,i}). \nonumber
         \end{align}
\end{multicols}

\noindent In \eqref{eq:theta_SBNN-IP}, $\mub^w_l \equiv (\mu^w_{l,1},\dots, \mu^w_{l,n_l^w})'$ and $\sigmab^w_l \equiv (\sigma^w_{l,1},\dots, \sigma^w_{l,n_l^w})'$ are the prior means and the prior standard deviations of the weights at layer $l$, and $\gammab_l^w \equiv  (\gamma^w_{l,1},\dots, \gamma^w_{l,n_l^w})'$ is a vector of real-valued elements. The superscript $b$ replaces the superscript $w$ to denote the corresponding quantities for the bias parameters. The hyper-parameters that need to be estimated when calibrating the SBNN-IP are $\psib \equiv \{\{\mub_l^w, \mub_l^b, \gammab_l^w, \gammab_l^b\}: l = 1,\dots,L\}$. 

\vspace{0.2in}

\noindent {\bf BNN-IL / BNN-IP:} The two variants of the vanilla BNN, BNN-IL and BNN-IP, are practically identical to SBNN-IL and SBNN-IP, respectively, with the same sets of hyper-parameters $\psib$ that need to be calibrated but with $\fvec_0(\svec; \theta_0) = \svec$ instead.

\vspace{0.2in}

\noindent {\bf SBNN-VL:}  Denote the spatially-varying parameters in layer $l$ of the SBNN-V at location $\svec \in D$ as $\thetab_l(\svec) \equiv \{\thetab^w_l(\svec),\thetab^b_l(\svec)\}$, where $\thetab^w_l(\cdot)$ and $\thetab^b_l(\cdot)$ are the spatially-varying weight and bias parameters, respectively, and let $\thetab(\svec) \equiv \{\thetab_l(\svec): l = 1,\dots,L\}$ denote all the SBNN-V parameters at location $\svec \in D$. The SBNN-V with a `prior-per-layer' scheme (SBNN-VL) is given by the following hierarchical spatial statistical model:

\begin{minipage}{\textwidth}
   
\vspace{0.1in}

For $\svec \in D,~ l = 1,\dots, L$,  
\begin{align*}
    Y(\svec) \equiv f(\svec; \thetab(\svec)) &= f_L (\fvec_{L-1} (\cdots \fvec_1(\fvec_0(\svec; \theta_0); \thetab_1(\svec)) \cdots; \thetab_{L-1}(\svec)); \thetab_L(\svec)),\\
    \mathbf{f}_l(\cdot\,; \thetab_l(\svec)) &= \frac{1}{\sqrt{d_{l-1}}} \mathbf{W}_l(\svec) \bm{\varphi}_{l-1}(\cdot) + \mathbf{b}_l(\svec), \\
    \mathbf{f}_0(\svec; \theta_0) &= \rhob(\svec;\tau) \equiv (\rho_1(\svec;\tau),\ldots,\rho_{K}(\svec;\tau))',     
  \end{align*}
  
  \vspace{-0.4in}
  
  \begin{multicols}{2}  
    \begin{align*}
        \theta^w_{l,i}(\svec) & = \mu^w_l(\svec) + \sigma^w_l(\svec)\eta^w_{l,i}, \quad i = 1,\dots, n^w_l,\\
    \mu^w_l(\svec) &= (\alphab^{w}_l)'\rhob(\svec),\\
    \sigma^w_l(\svec) &= \mathrm{softplus}((\betab^{w}_l)'\rhob(\svec)), \\
    \eta^w_{l,i} &\stackrel{\mathrm{iid}}{\sim} \textrm{Gau}(0, 1), \quad i = 1,\dots, n^w_l,
    \end{align*}

    \columnbreak  
    
    \begin{align}
        \theta^b_{l,i}(\svec) & = \mu^b_l(\svec) + \sigma^b_l(\svec)\eta^b_{l,i}, \quad i = 1,\dots, n^b_l,\nonumber\\
        \mu^b_l(\svec) &= (\alphab^{b}_l)'\rhob(\svec),\label{eq:theta_SBNN-VL} \\
        \sigma^b_l(\svec) &= \mathrm{softplus}((\betab^{b}_l)'\rhob(\svec)), \nonumber \\
        \eta^b_{l,i} &\stackrel{\mathrm{iid}}{\sim} \textrm{Gau}(0, 1), \quad i = 1,\dots, n^b_l,\nonumber 
    \end{align}
\end{multicols}
\end{minipage}

\vspace{0.2in}

\noindent where $\theta_0 = \tau$. In \eqref{eq:theta_SBNN-VL}, $\thetab_l^w(\cdot) \equiv (\theta_{l,1}^w(\cdot),\dots,\theta_{l,n_l^w}^w(\cdot))' = \textrm{vec}(\Wmat_l(\cdot))$ 
are the weight parameters at layer $l$; $\alphab_l^w \equiv (\alpha_{l,1}^w,\dots,\alpha^w_{l,K})'$ and $\betab_l^w \equiv (\beta_{l,1}^w,\dots,\beta_{l,K}^w)'$ are vectors of real-valued basis-function coefficients for the prior mean and the prior standard deviation of the weights at layer $l$; and $n_l^w$ is the number of weights in layer $l$. The superscript $b$ replaces the superscript $w$  to denote the corresponding quantities for the spatially-varying bias parameters ${\thetab_l^b(\cdot) \equiv (\theta_{l,1}^b(\cdot),\dots,\theta_{l,n_l^b}^b(\cdot))' = \bvec_l(\cdot)}$. The hyper-parameters that need to be estimated when calibrating the SBNN-VL are $\psib \equiv \{\{\alphab_l^w, \alphab_l^b, \betab_l^w, \betab_l^b\}: l = 1,\dots,L\}$. As with the SBNN-IL, the spatial process $Y(\cdot)$ is the output of the final layer. 

\vspace{0.2in}

\noindent {\bf SBNN-VP:} For the SBNN-V with a `prior-per-parameter' scheme (SBNN-VP), the weights and biases are instead given by:

\begin{multicols}{2}  
    \textrm{For $\svec \in D$, $l = 1,\dots,L$, $ i = 1,\dots, n^w_l$,}
    \begin{align*}
        \theta^w_{l,i}(\svec) & = \mu^w_{l,i}(\svec) + \sigma^w_{l,i}(\svec)\eta^w_{l,i}, \\
    \mu^w_{l,i}(\svec) &= (\alphab^{w}_{l,i})'\rhob(\svec),\\
    \sigma^w_{l,i}(\svec) &= \mathrm{softplus}((\betab^{w}_{l,i})'\rhob(\svec)), \\
    \eta^w_{l,i} &\stackrel{\mathrm{iid}}{\sim} \textrm{Gau}(0, 1),
    \end{align*}

    \columnbreak  
    
    \textrm{For $\svec \in D$, $l = 1,\dots,L$, $ i = 1,\dots, n^b_l$,}
    \begin{align}
        \theta^b_{l,i}(\svec) & = \mu^b_{l,i}(\svec) + \sigma^b_{l,i}(\svec)\eta^b_{l,i},\nonumber\\
        \mu^b_{l,i}(\svec) &= (\alphab^{b}_{l,i})'\rhob(\svec),\label{eq:theta_SBNN-VP}\\
        \sigma^b_{l,i}(\svec) &= \mathrm{softplus}((\betab^{b}_{l,i})'\rhob(\svec)), \nonumber \\
        \eta^b_{l,i} &\stackrel{\mathrm{iid}}{\sim} \textrm{Gau}(0, 1).\nonumber 
    \end{align}
\end{multicols}

\noindent In \eqref{eq:theta_SBNN-VP}, $\alphab_{l,i}^w \equiv (\alpha_{l,i,1}^w,\dots,\alpha^w_{l,i,K})'$ and $\betab_{l,i}^w \equiv (\beta_{l,i,1}^w,\dots,\beta_{l,i,K}^w)'$ are real-valued basis-function coefficients for the prior mean and the prior standard deviation of the $i$th weight at layer $l$. The superscript $b$ replaces the superscript $w$  to denote the corresponding quantities for the spatially-varying bias parameters. The hyper-parameters that need to be estimated when calibrating the SBNN-VP are $\psib \equiv \{\{\Amat_l^w, \Amat_l^b, \Bmat_l^w, \Bmat_l^b\}: l = 1,\dots,L\}$, where matrix $\Amat_l^w \equiv (\alphab_{l,1}^w, \dots, \alphab_{l,n^w_l}^w)$, with $\Amat_l^b, \Bmat_l^w$ and $\Bmat_l^b$ similarly defined.

\section{SBNN calibration using the Wasserstein distance}\label{sec:calibration}

Assume now that we have access to realisations from a second stochastic process over $D$, $\widetilde{Y}(\cdot)$, which we refer to as the target process. These realisations from $\widetilde Y(\cdot)$ could be simulations from a stochastic simulator or data from a re-analysis product, say. Assume further that we want an SBNN to have a selected finite-dimensional distribution that `matches' that of the target's, in a sense detailed below. In this section we outline an approach to  adjusting the SBNN hyper-parameters $\psib$ in order to achieve this objective; we refer to the procedure of choosing $\psib$ such that a given finite-dimensional distribution of the two processes are closely matched as `calibration.' Calibration is a difficult task as it involves exploring the space of high-dimensional distribution functions, and it was not considered computationally feasible until very recently. The method we adopt for calibration, detailed in \citet{tran2022}, makes the problem computationally tractable. Their method is based on minimising a Wasserstein distance \citep[see, e.g., ][]{Panaretos_2019} between the given finite-dimensional distributions of the two processes via Monte Carlo approximations.

Consider a $d \times n$ matrix of locations $\Smat \equiv (\svec_1,\dots,\svec_n)$ where $\svec_1,\dots,\svec_n \in D$ with $n\ge 1$ (typically $\Smat$ is a fine gridding of $D$). For the SBNN-V, $\Thetab \equiv \{\thetab(\svec_1),\dots,\thetab(\svec_n)\}$ are the parameters at the set of locations. From \eqref{eq:composition}, {$\Yvec \equiv (Y(\svec_1),\dots,Y(\svec_n))' = (f(\svec_1; \thetab(\svec_1)),\dots,f(\svec_n; \thetab(\svec_n)))' \equiv \fvec_\Thetab$} denotes a vector whose elements are the process evaluated at these locations. The same definitions hold for the SBNN-I variants, but with $\thetab(\svec_i)$ replaced with $\thetab$, for $i = 1,\dots,n$. The distribution of $\Yvec$ is analytically intractable, but it is straightforward to simulate from, since $p(\Thetab ; \psib)$ (the prior distribution of the weights and biases) is straightforward to simulate from (see, e.g., \eqref{eq:generative_theta}) and $\Yvec = \fvec_\Thetab$ is a deterministic function of $\Thetab$ (see \eqref{eq:composition}). We collect the corresponding evaluations of the second, or target, process $\widetilde{Y}(\cdot)$ in $\widetilde\Yvec \equiv (\widetilde{Y}(\svec_1),\dots,\widetilde{Y}(\svec_n))'$. We then  match the distribution of $\Yvec$ to the empirical distribution of $\widetilde\Yvec$ by minimising the dissimilarity between these two distributions. The natural choice of dissimilarity measure to minimise is the Kullback-Leibler divergence; however, this divergence term contains an entropy term that is analytically intractable and computationally difficult to approximate \citep{flam-shepherd2017,delattre2017}. On the other hand, the Wasserstein distance leads to no such difficulty.

The {Wasserstein distance} is a measure of dissimilarity between two probability distributions. As in \citet{tran2022}, we consider a special case of the \emph{Wasserstein-1 distance} that is given by
\begin{equation}
  W_1(\psib) \equiv \sup_{\phi: \| \nabla \phi \| \le 1} ~ \mathbb{E} [\phi(\Yvec) \mid \psib] - \mathbb{E} [\phi(\widetilde\Yvec)],
  \label{eq:wasserstein1_dual}
\end{equation} 
where the first expectation is taken with respect to the distribution derived from the SBNN; the second expectation is taken with respect to the distribution derived from the target process; $\phi(\cdot)$ is differentiable; and where the constraint on the gradient norm $ \| \nabla \phi \| \le 1$ ensures that $\phi(\cdot)$ is also 1-Lipschitz. A Monte Carlo approximation to \eqref{eq:wasserstein1_dual} yields the empirical expectations,
\begin{equation}
  W_1(\psib) \approx \sup_{\phi: \| \nabla \phi \| \le 1} ~ \frac{1}{N}\sum_{i = 1}^N \left(\phi(\Yvec^{(i)}_{\psib}) - \phi(\widetilde\Yvec^{(i)})\right),
  \label{eq:wasserstein1_dual_MC}
\end{equation}
where $\Yvec_{\psib}^{(i)} \equiv \fvec_{\Thetab^{(i)}}$ for $\Thetab^{(i)} \sim p(\Thetab ; \psib)$, $\widetilde\Yvec^{(i)}$ is a sample from the target process, and $N$ is the number of samples from each process. Note that one could choose different sample sizes for the Monte Carlo approximation of the two expectations in~\eqref{eq:wasserstein1_dual} if needed; for ease of exposition we assume that both sample sizes are equal to $N$. Calibration proceeds by minimising this approximate Wasserstein distance. Specifically, we seek $\psib^*$, where
\begin{equation}\label{eq:psi}
\psib^* = \argmin_{\psib} W_1(\psib).
\end{equation}
Since all the functions in the BNN are differentiable with respect to $\psib$, we use gradient descent to carry out the optimisation in a computationally efficient way. 

The Wasserstein distance in \eqref{eq:wasserstein1_dual} is a supremum over the space of 1-Lipschitz functions. There is no computationally feasible way to explore this function space and, therefore, we adopt the same approach as in \citet{gulrajani2017} and \citet{tran2022} and model $\phi(\cdot)$ using a neural network $\phi_{NN}(\cdot\,; \lambdab)$ parameterised by some parameters $\lambdab$, which is differentiable by construction. This leads to the following approximation of the Wasserstein-1 distance \eqref{eq:wasserstein1_dual},
$$
W_1(\psib) \approx \mathbb{E} [\phi_{NN}(\Yvec; \lambdab^*) \mid \psib] - \mathbb{E} [\phi_{NN}(\widetilde\Yvec; \lambdab^*)],
$$
where the optimised neural network parameters are given by
\begin{align}
    \bm{\lambda}^*
    &= \mathop{\arg\max}_{\bm{\lambda}} \mathbb{E} [\phi_{NN}(\Yvec; \lambdab) \mid \psib] - \mathbb{E} [\phi_{NN}(\widetilde\Yvec; \lambdab)] \label{eq:opt_hyperparams_exact} \\
    &\approx \mathop{\arg\max}_{\bm{\lambda}} \frac{1}{N}\sum_{i=1}^N \left(\phi_{NN}(\Yvec^{(i)}_{\psib}; \lambdab) - \phi_{NN}(\widetilde{\Yvec}^{(i)}; \lambdab)\right),  \label{eq:opt_hyperparams}
\end{align}
where we have again chosen to use the sample size $N$ for the approximation of both expectations for notational convenience.

So far we have not ensured that $\|\nabla\phi_{NN}(\cdot\,; \lambdab^*)\| \le 1$ (i.e., 1-Lipschitz continuity), just that $\phi_{NN}(\cdot\,; \lambdab^*)$ is differentiable (by construction). To ensure that $\phi_{NN}(\cdot\,; \lambdab^*)$ is at least approximately 1-Lipschitz, we make use of a remarkable result by \citet{gulrajani2017}, which we summarise as follows: For the weighted average $\overline{\Yvec} \equiv \delta\Yvec + (1 - \delta)\widetilde\Yvec, ~ {0\le \delta \le 1}$, the 1-Lipschitz function that maximises \eqref{eq:wasserstein1_dual}, $\phi^*(\cdot)$ say, has gradient-norm $\|\nabla\phi^*(\overline{\Yvec})\| = 1$ with probability one. Therefore, when exploring the space of differentiable functions, one need only consider those functions that have a gradient-norm at $\overline{\Yvec}$ equal to one. In practice, we can use this result to suggest a penalisation term to add to \eqref{eq:opt_hyperparams}; specifically, we alter \eqref{eq:opt_hyperparams_exact} and \eqref{eq:opt_hyperparams} as follows:
\begin{align}
    \bm{\lambda}^*
    &= \mathop{\arg\max}_{\bm{\lambda}} \mathbb{E} [\phi_{NN}(\Yvec; \lambdab) \mid \psib] - \mathbb{E} [\phi_{NN}(\widetilde\Yvec; \lambdab)] - \zeta \mathbb{E}\left[\left(\|\nabla \phi_{NN}(\overline{\Yvec}; \lambdab)\| - 1\right)^2 \right]\label{eq:opt_hyperparams_exact2} \\
    &\approx \mathop{\arg\max}_{\bm{\lambda}} \frac{1}{N}\sum_{i=1}^N \left(\phi_{NN}(\Yvec^{(i)}_{\psib}; \lambdab) - \phi_{NN}(\widetilde{\Yvec}^{(i)}; \lambdab) - \zeta\left(\|\nabla \phi_{NN}(\overline{\Yvec}^{(i)}; \lambdab)\| - 1\right)^2 \right),  \label{eq:opt_hyperparams2}
\end{align}
where, for $i = 1,\dots,N$, $\overline{\Yvec}^{(i)} = \delta^{(i)}\Yvec^{(i)} + (1 - \delta^{(i)})\widetilde\Yvec^{(i)}$, $\delta^{(i)}$ is indepentally drawn from the bounded uniform distribution on $[0,1]$, and $\zeta$ is a tuning parameter. Following the recommendation of \citet{gulrajani2017} we set $\zeta = 10$. Note that this ``soft-constraint'' approach does not guarantee that $\phi_{NN}(\cdot\,;\lambdab^*)$ will be 1-Lipschitz. However, our results in Section~\ref{sec:results} suggest that this approach is effective. Neural networks that enforce 1-Lipschitz continuity by design are available, and we discuss them in Section~\ref{sec:conclusion}.

Note that \eqref{eq:psi} and \eqref{eq:opt_hyperparams_exact2} together form a two-stage optimisation problem, which we solve using gradient methods. In the first stage, which we refer to as the inner-loop optimisation, we optimise $\lambdab$ using gradient ascent while keeping $\psib$ fixed, in order to establish the Wasserstein distance for the fixed value of $\psib$. In the second stage, which we refer to as the outer-loop optimisation, we do a single gradient descent step to find a new $\psib$ (conditional on $\lambdab$) that reduces the Wasserstein distance. We only do one step at a time in the outer-loop optimisation stage since the Wasserstein distance needs to be re-established for every new value of $\psib$. We generate $N$ samples from $\Yvec, \widetilde\Yvec$, and $\overline{\Yvec}$, once for each inner-loop optimisation and once for each outer-loop step.  We iterate both stages until the Wasserstein distance ceases to notably decrease after several outer-loop optimisation steps.

The calibration procedure outlined above optimises $\psib$ such that the distribution of $\Yvec$ (i.e., $Y(\cdot)$ evaluated over $\Smat$) is close, in a Wasserstein-1 sense, to that of $\widetilde\Yvec$ (i.e., $\widetilde Y(\cdot)$ evaluated over the same matrix of locations $\Smat$). The choice of $\Smat$ determines the finite-dimensional distribution being compared. In applications where we have access to the realisations $\widetilde{\Yvec}$ but where we cannot simulate from $\widetilde{Y}(\cdot)$ at arbitrary locations, $\Smat$ is fixed by the application. When we can freely simulate from $\widetilde{Y}(\cdot)$, then, since $d$ is typically small in spatial applications, one may define $\Smat$ as a fine gridding over $D$, and this is the approach we present throughout this paper.  The optimised hyper-parameters $\psib^*$ lead to an SBNN, $Y(\cdot)$, that approximates the target process  $\widetilde Y(\cdot)$ well over the locations in $\Smat$.

\section{Simulation studies} \label{sec:results}

\Copy{TrueProcess}{In all the simulation studies given in this section, we consider the special case that the target process $\widetilde{Y}(\cdot)$ is known and that it can be easily simulated from, so that we can compare the (S)BNNs to the process to which they are calibrated. In practice, the target process may be unknown, so in that case we would need realisations from it that are used to calibrate the (S)BNNs. These could come from a remote sensing instrument or from a stochastic simulator.} We consider three processes for $\widetilde{Y}(\cdot)$: a stationary Gaussian spatial process (Section~\ref{sec:StatGP}); a non-stationary Gaussian spatial process (Section~\ref{sec:NonStatGP}); and a stationary lognormal spatial process (Section~\ref{sec:Lognormal}). We define all these processes on a two-dimensional spatial domain $D = [-4,4] \times [-4,4]$, and we set $\svec_1,\dots,\svec_n$ as the centroids of a $64 \times 64$ gridding of $D$ (so that $\Smat$ is a matrix of size $2 \times 4096$). For the activation functions in ${\bm \varphi}_l(\cdot),~l =0,\dots,L-1,$ we use the $\tanh(\cdot)$ function, where ${\mathrm{tanh}:t\mapsto (e^{2t}-1)/(e^{2t}+1)}$. We use $L = 4$ layers and set each hidden layer to have dimension $40$; that is, we set $d_1 = d_2 = d_3 = 40$. The input and output dimensions are $d = 2$ and $d_4 = 1$, while $d_0 = 2$ for vanilla BNNs, and $d_0 = K$ for SBNNs, where recall that $K$ is the number of basis functions in the embedding layer.  We let $\phi_{NN}(\cdot\,; \lambdab)$ be a neural network with two hidden layers each of the form~\eqref{eq:forwardpass} with dimension 200 and with softplus activation functions. The weights and biases of the two layers, which we collect in $\lambdab$, are initialised by simulating from the bounded uniform distribution on $[-\sqrt{z}, \sqrt{z}]$, where $z$ is given by the reciprocal of the input dimension of the respective layer \citep{He_2015}.

\begin{table}\caption{Summary of the models used in the experiments of Sections~\ref{sec:results} and \ref{sec:inference}. \label{tab:model_definition}}
  \begin{tabular}{lp{14cm}}
    \hline  Model & Summary \\ \hline
    BNN-IL & BNN  with spatially-invariant parameters and with a `prior-per-layer' scheme \\
    BNN-IP & BNN  with spatially-invariant parameters and with a `prior-per-parameter' scheme \\
    SBNN-IL & Spatial BNN with an embedding layer, spatially invariant parameters, and with a `prior-per-layer' scheme \\
    SBNN-IP & Spatial BNN with an embedding layer, spatially invariant parameters, and with a `prior-per-parameter' scheme \\
    SBNN-VL & Spatial BNN with an embedding layer, spatially varying parameters, and with a `prior-per-layer' scheme \\
    SBNN-VP & Spatial BNN with an embedding layer, spatially varying parameters, and with a `prior-per-parameter' scheme \\\hline
\end{tabular}
\end{table}

Throughout the simulation studies we consider the six (S)BNNs discussed in Section~\ref{sec:architecture}, and summarised in Table~\ref{tab:model_definition}. For the SBNNs we set $\fvec_0(\cdot\,; \theta_0) = \rhob(\cdot\,;\tau)$ as in \eqref{eq:rbf} with length scale $\theta_0 = \tau = 1$, and we use $K = 15^2$ radial basis functions arranged on a $15 \times 15$ grid in $D$. In Figure~\ref{fig:basis_functions} in the Supplementary Material we show evaluations of a subset of the basis functions whose centroids vary over $\svec = (0,s_2)'$, where $s_2 \in [-4,4]$. We did not conduct a detailed experiment to analyse how the results change with $\tau$; provided there is reasonable overlap between the basis functions, as in our case, we do not expect the results to substantially change with $\tau$.

As initialisation for our six models, we set all the $\mu$'s equal to zero, and all the $\gamma$'s equal to one for the BNN-I and SBNN-I variants. For the SBNN-V variants, we set all the $\alpha$'s equal to zero, while we simulate all the $\beta$'s independently from a normal distribution with zero mean and unit variance. When calibrating, we sample $N$ parameter vectors, ${\{\Thetab^{(i)}: i = 1,\dots,N\}}$, from the prior on the weights and biases, $N$ corresponding realisations from the (S)BNN $\{\Yvec^{(i)}: i = 1,\dots,N\}$, and $N$ realisations from the target process $\{ \widetilde\Yvec^{(i)}: i = 1,\dots,N\}$, where
$N = 1024$ for the BNN-I and SBNN-I variants, and where $N = 512$ for the SBNN-V variants, in order to reduce memory requirements. Using these simulations, we then carry out 50 gradient steps when optimising $\lambdab$ and keeping $\psib$ fixed in \eqref{eq:opt_hyperparams} (inner-loop optimisation). We then re-simulate to obtain $N$ realisations of $Y(\cdot)$ and $\widetilde{Y}(\cdot)$ before carrying out a single gradient step when optimising $\psib$ and keeping $\lambdab$ fixed in \eqref{eq:wasserstein1_dual_MC} (outer-loop optimisation). Recall that we only do a single gradient step when optimising $\psib$ as the Wasserstein distance in \eqref{eq:wasserstein1_dual} depends on $\psib$, and thus needs to be re-established (i.e., $\lambdab$ needs to be re-estimated) for every update of $\psib$. We repeat this two-stage procedure of iteratively optimising $\lambdab$ and $\psib$ (always using the most recently updated values of $\lambdab$ and $\psib$ as initial conditions) until the Wasserstein distance stabilises. Since we are generating data `on-the-fly' when calibrating, there is little risk of over-fitting \citep{Chan_2018}; see Figure~\ref{fig:wass_distances} in the Supplementary Material for plots of the Wasserstein distances for all models and simulation experiments as a function of the outer-loop optimisation step. Note that these Wasserstein distances are necessarily approximate since the 1-Lipschitz functions are approximated using a neural network, as outlined in Section~\ref{sec:calibration}. As in \citet{tran2022}, we use Adagrad and RMSprop \citep{kochenderfer2019} strategies for adjusting the gradient step sizes in the inner- and outer-loop optimisations, respectively.

In Table~\ref{tab:results} we show the Wasserstein-1 distance averaged over the final 100 outer-loop iterations for the six (S)BNNs we consider. In this table we also list the number of hyper-parameters associated with each model. In the case of the `prior-per-layer' BNN-IL and SBNN-IL, the number of hyper-parameters is 16 (two mean and two variance hyper-parameters for each layer). In the case of the `prior-per-layer' SBNN-VL, the number of hyper-parameters in each of the $L = 4$ layers is $4K = 900$, for a total of $4 \times 900 = 3600$ hyper-parameters. The `prior-per-parameter' models are considerably more parameterised since, for these models, the number of hyper-parameters grows linearly with the number of weights and biases in the model. The number of weights and biases is 3441 for BNN-IL and BNN-IP, and it is 12361 for all the SBNNs with an embedding layer. Multiplying the number of parameters by two for the BNN-I and the SBNN-I variants, and by $2K$ for the SBNN-V variant, gives the total number of hyper-parameters associated with each of these `prior-per-parameter' models. A cursory glance at this table reveals that a larger number of hyper-parameters generally, but not always, leads to a lower Wasserstein distance, and that the SBNN variants generally outperform the BNN-I variants, sometimes by a substantial margin. We explore the nuances in more detail in the following sections.

Reproducible code, which contains additional details on the simulation studies in this section, is available from \url{https://github.com/andrewzm/SBNN}.

\begin{table}[!t]
\centering
\caption{Number of parameters, number of hyper-parameters, and Wasserstein distance on convergence for each simulation experiment and (S)BNN combination. The reported Wasserstein distance is computed as the average over the distances in the final 100 outer-loop optimisation steps. \label{tab:results}} 
\begin{tabular}{lp{1.5cm}p{1.75cm}p{1.85cm}p{2.6cm}p{1.75cm}}
  \hline
Model & Num. par. & Num. hyper-par. & $W_1(\psib)$ (Stat. GP) & $W_1(\psib)$ \newline (Non-stat. GP) & $W_1(\psib)$ (Log-GP) \\ 
  \hline
BNN-IL & 3441 & 16 & 107.14 & 40.37 & 28.18 \\ 
  BNN-IP & 3441 & 6882 & 122.33 & 60.35 & 9.09 \\ 
  SBNN-IL & 12361 & 16 & 4.22 & 95.64 & 10.67 \\ 
  SBNN-IP & 12361 & 24722 & 6.72 & 2.65 & 0.69 \\ 
  SBNN-VL & 12361 & 3600 & 7.86 & 6.70 & 8.37 \\ 
  SBNN-VP & 12361 & 5562450 & 2.00 & 0.95 & 1.03 \\ 
   \hline
\end{tabular}
\end{table}

\subsection{Calibration to stationary Gaussian spatial process}\label{sec:StatGP}

In this simulation study we consider a mean zero stationary and isotropic Gaussian spatial process  with unit variance and squared-exponential covariance function as our target process $\widetilde{Y}(\cdot)$. Hence, $\widetilde{\Yvec} \sim \textrm{Gau}(\zerob, \Sigmamat)$, where $\Sigmamat \equiv (\cov(\widetilde{Y}(\svec_k), \widetilde{Y}(\svec_l)): k,l = 1,\dots, n)$. We model the covariances through a squared-exponential covariogram $C^o(\cdot)$,
\begin{equation}
     \cov(\widetilde{Y}(\svec),\widetilde{Y}(\rvec)) \equiv C^o(\|\svec - \rvec\|) = \exp\left(-\frac{\|\svec-\rvec\|^2}{2\ell^2}\right),\quad \svec,\rvec \in D, \label{eq:sqexp_covariogram}
\end{equation}
where we set the length scale $\ell = 1$. From Table~\ref{tab:results} we see that all the SBNN variants perform similarly in this case, and considerably outperform the BNN-I variants. For this reason, in the following discussion  we focus on comparing the calibrated BNN-IL and the calibrated SBNN-IL to the target Gaussian process; the results shown below are representative of those from their respective variants. 

\begin{figure}[t!]
        
   \includegraphics{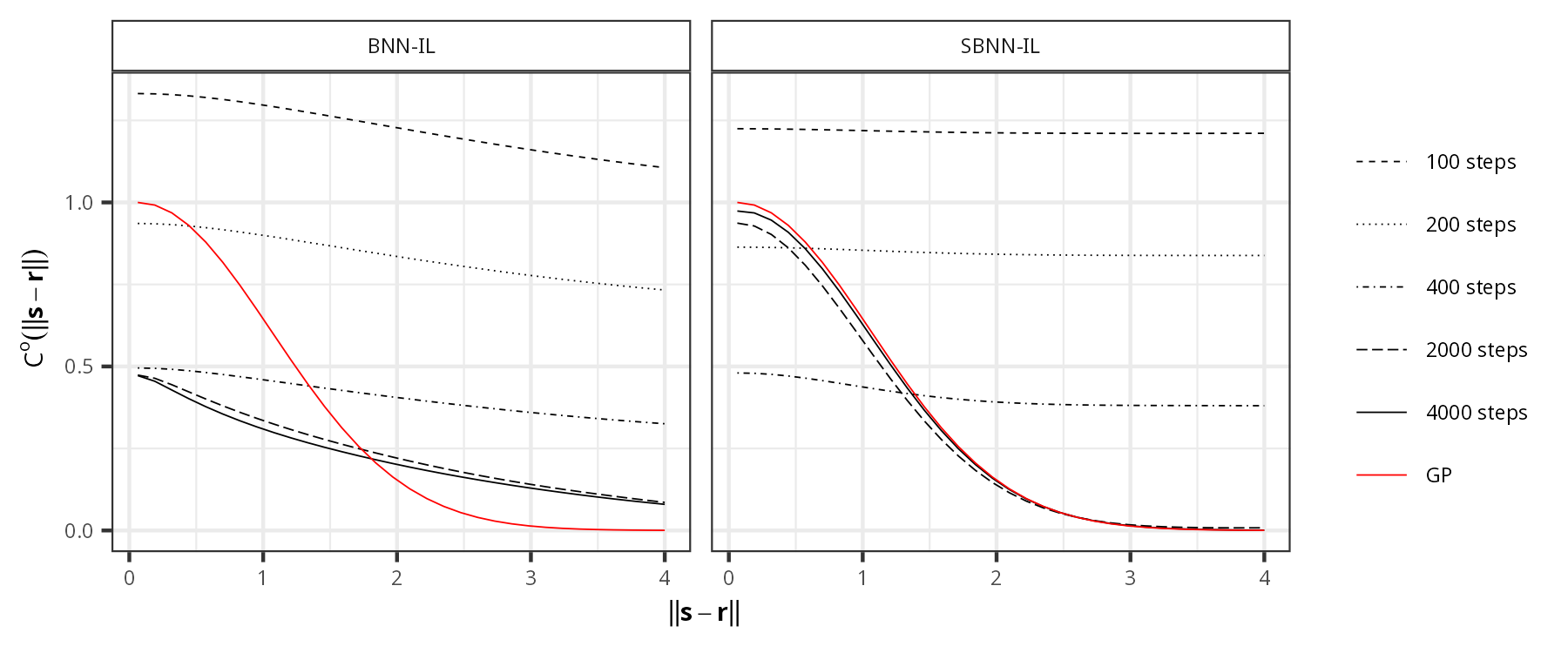}

    \caption{Empirical covariogram of the (S)BNN at different stages during the optimisation (different line-styles denote the empirical covariogram after 100, 200, 400, 2000, and 4000 gradient steps, respectively), and the true covariogram of the target Gaussian process (red). Left: BNN-IL. Right: SBNN-IL. \label{fig:gp_bnn_covs}}
\end{figure}

We first compare the two models through the empirical covariogram calculated using samples from the networks. Rather than only computing the empirical covariogram at the final (in this case, the 4000th) optimisation step, we compute empirical estimates at a number of intermediate steps in order to monitor the adaptation of the (S)BNN to the target process during optimisation. Specifically, we compute empirical covariograms after 100, 200, 400, 2000, and 4000 outer-loop gradient steps, respectively, and compare these estimates to the true covariogram of the target Gaussian process. Figure~\ref{fig:gp_bnn_covs}, left panel, shows that the calibrated BNN-IL fails to recover the true covariogram; on convergence, BNN-IL's covariogram has smaller intercept at the origin and slower decrease as the spatial lag increases. On the other hand, Figure~\ref{fig:gp_bnn_covs}, right panel, shows that the covariogram of the SBNN-IL with an embedding layer converges (after approximately 2000 outer-loop gradient steps) to a covariogram that is very similar to that of the target process.

\begin{figure}[t!]

    \includegraphics{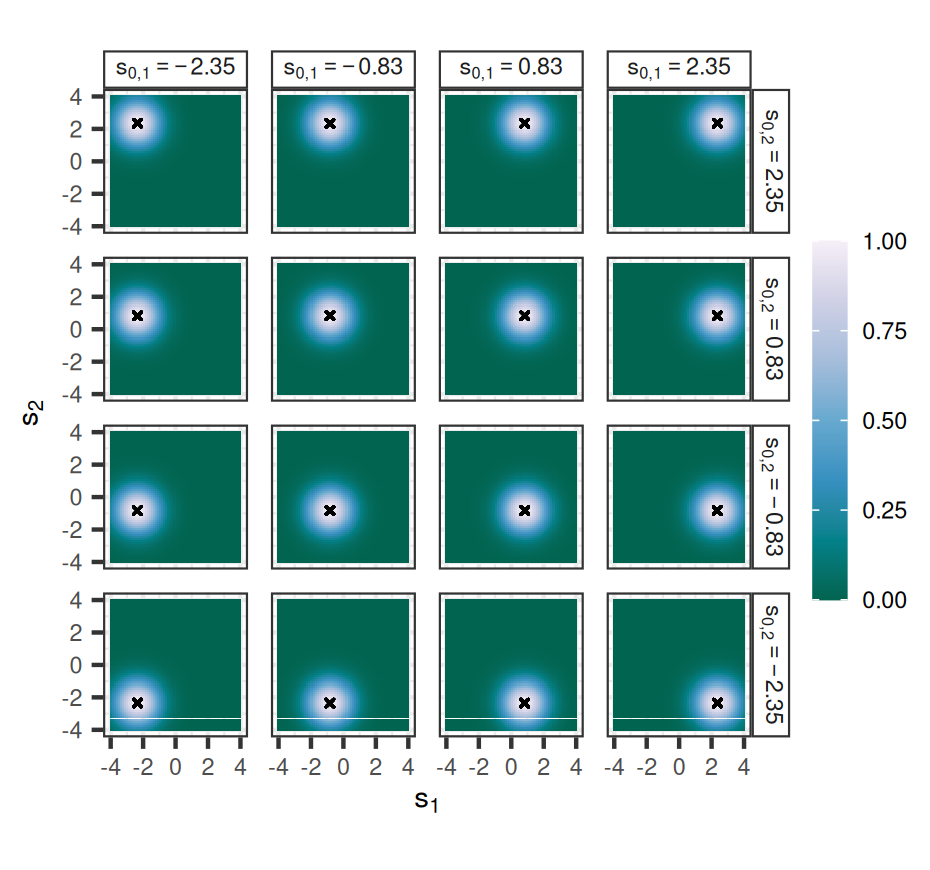}    
    \includegraphics{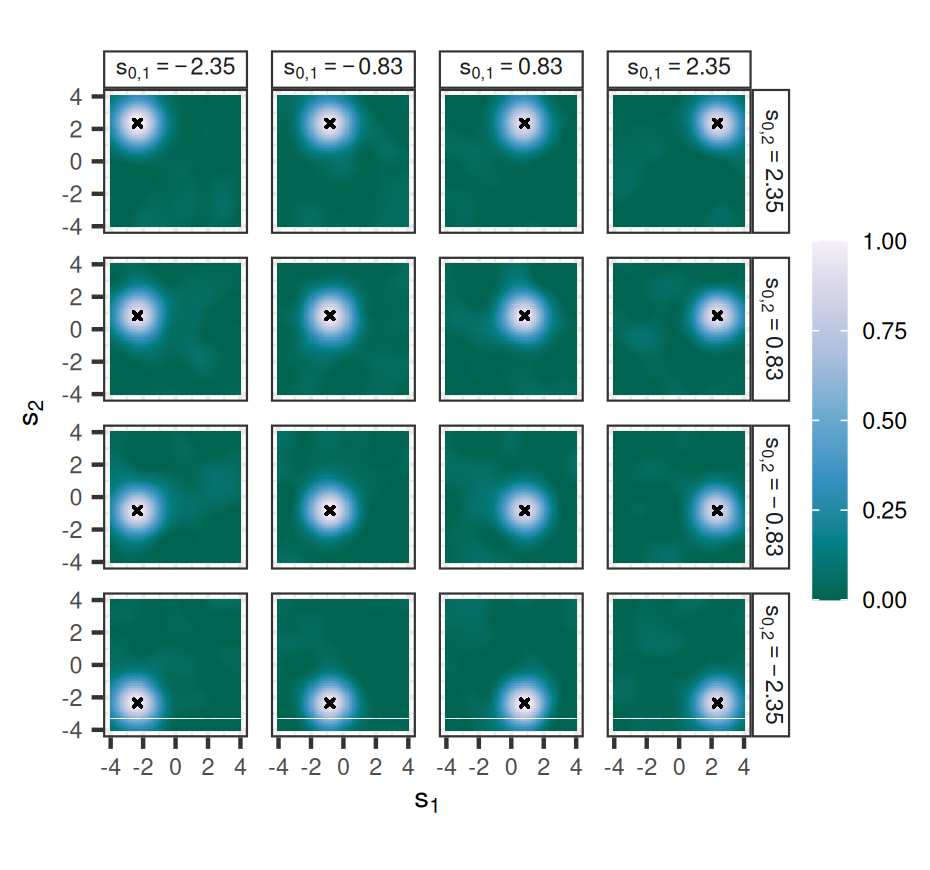}
    \includegraphics{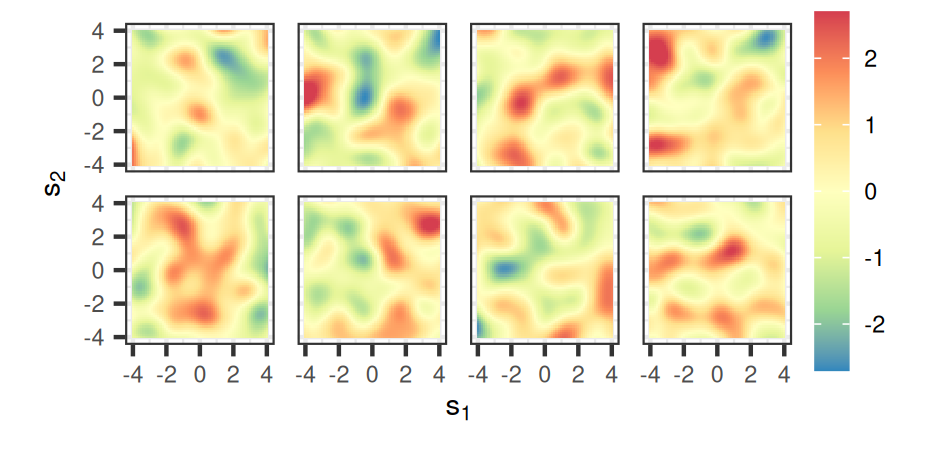}
    \includegraphics{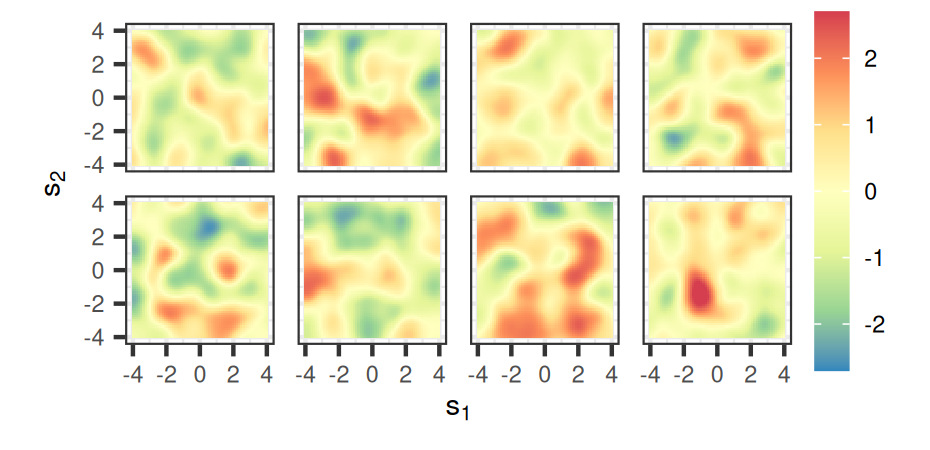}
    
    \caption{(Top-left panel) Covariance between the target process (stationary GP) at 16 grid points (crosses), with coordinates as indicated by the labels at the top and the right of the sub-panels, and the target process on a fine gridding $(64 \times 64)$ of $D = [-4, 4] \times [-4, 4]$. (Bottom-left panel) Eight realisations of the underlying target process on the same fine gridding of $D$. (Right panels) Same as left panels but for the calibrated SBNN-IL. \label{fig:gp_bnn_heatmaps_samples}}

\end{figure}

In Figure~\ref{fig:gp_bnn_heatmaps_samples}, top-left panel, we plot heat maps of the target process' covariances, ${\cov(\widetilde{Y}(\svec_0),\widetilde{Y}(\Smat))}$, for 16 values of $\svec_0$ arranged on a regular $4 \times 4$ grid in $D$, where recall that $\widetilde{Y}(\cdot)$ is stationary and $\Smat$ is made up of the grid-cell centroids of a 64 $\times$ 64 gridding of $D$. In the bottom-left panel we plot realisations of $\widetilde\Yvec$. 
In the top-right panel we show the corresponding empirical estimates of the covariances ${\cov(Y(\svec_0),Y(\Smat))}$ from the calibrated SBNN-IL.  
These covariances are indicative of stationarity and isotropy, and they are very similar to those of the target process. This is reassuring as there is nothing in the construction or training procedure of the SBNN-IL that constrains the process to be stationary or isotropic; the similarity between the covariances is another indication that the SBNN-IL is targeting the correct process. In the bottom-right panel of Figure~\ref{fig:gp_bnn_heatmaps_samples} we plot sample realisations of  $\Yvec$. 
These realisations have very similar properties to those of $\widetilde\Yvec$ (similar length scale, smoothness, and variance).

\begin{figure}[t!]
        
    \includegraphics{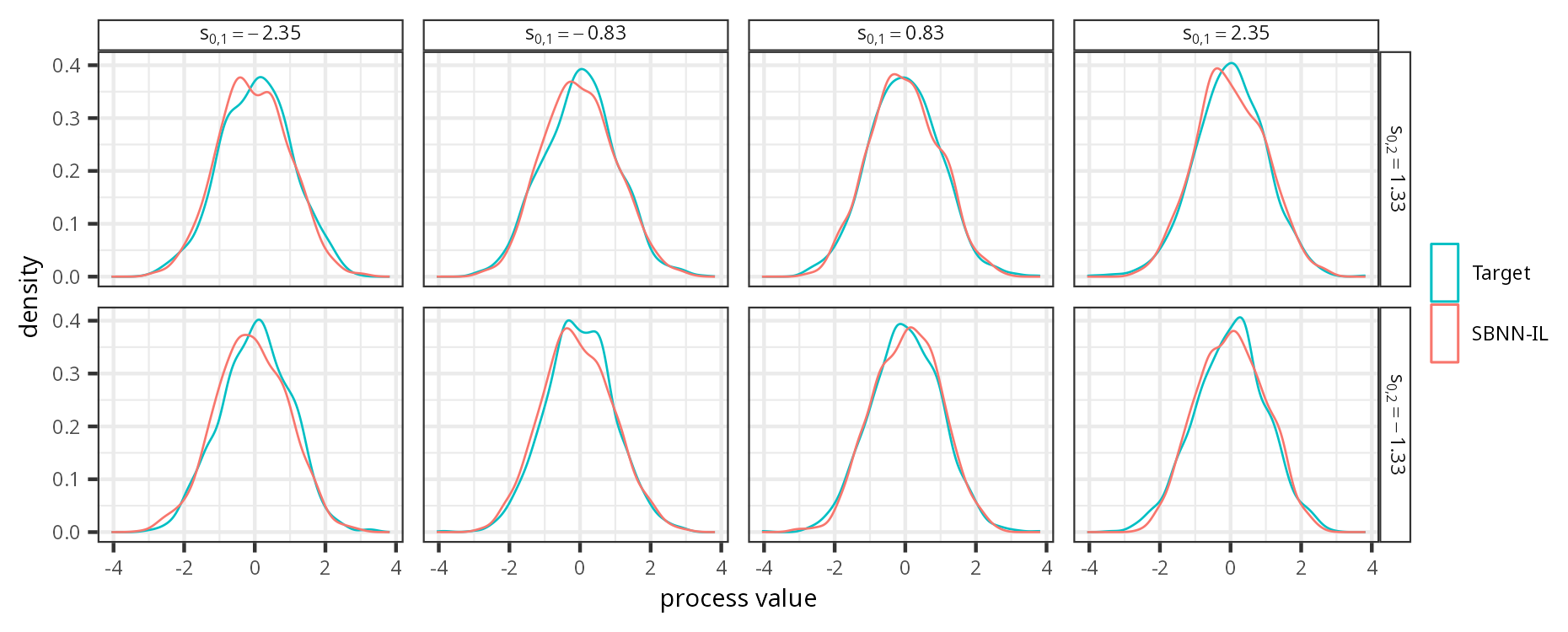} \\
    \includegraphics{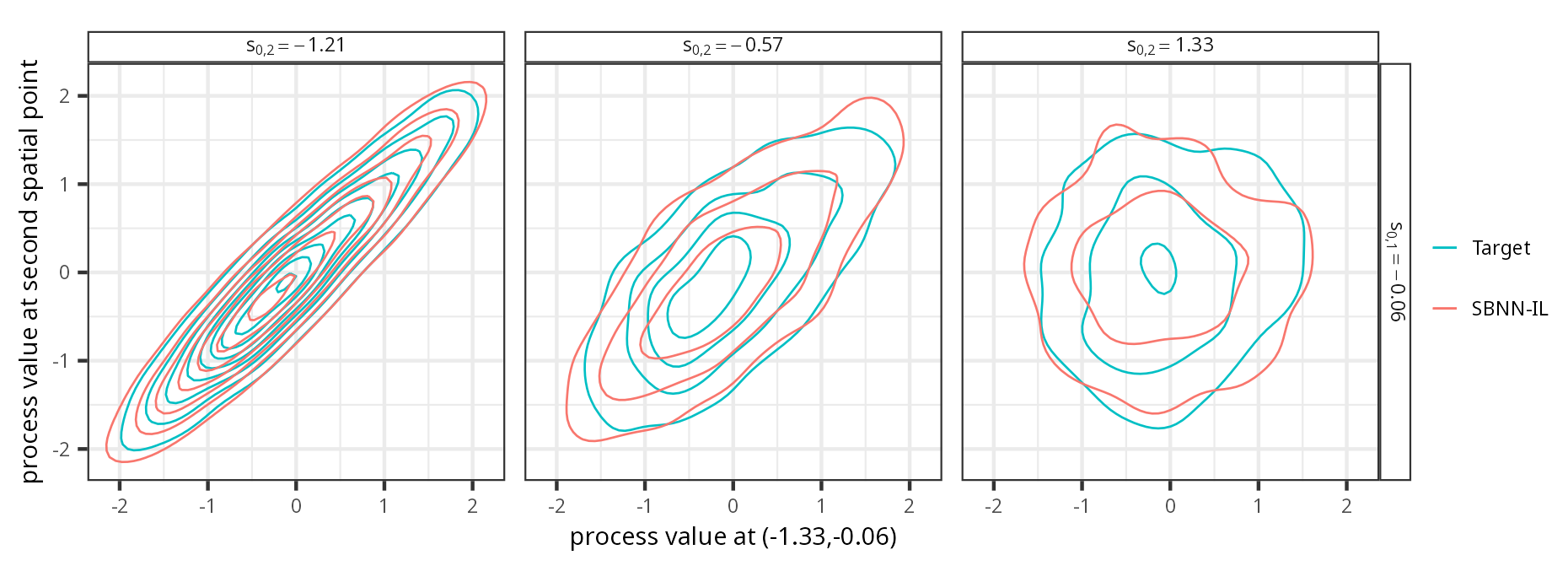} 
    \caption{Kernel density plots from 1000 samples of the SBNN-IL and the target process (stationary GP). (Top panel) Univariate densities of the two processes at eight spatial locations arranged on a $2 \times 4$ grid in $D$ (with coordinates as indicated by the labels of the sub-panels). (Bottom panel) Overlayed bivariate densities of the two processes at $\tilde\svec_0  = (-1.33, -0.06)'$ and three other locations in $D$ (with coordinates as indicated by the labels of the sub-panels). \input{./figures/Section4_3_SBNN-VP_contour_levels.tex}  \label{fig:prior_densities_stationaryGP}}
\end{figure}

Although the SBNN-IL is a highly non-Gaussian process, it has been calibrated against a Gaussian process, and thus all its finite-dimensional distributions derived from the high-dimensional one that was used for calibration should be approximately Gaussian. To illustrate that Gaussianity is well approximated, in Figure~\ref{fig:prior_densities_stationaryGP} we plot kernel density estimates from 1000 samples taken from the calibrated SBNN-IL and the true Gaussian process. The top panel shows empirical marginal densities of $Y(\svec_0)$ and $\widetilde{Y}(\svec_0)$ for eight values of $\svec_0$ arranged on a $2 \times 4$ grid in $D$, while the bottom panel shows bivariate densities corresponding to $(Y(\widetilde\svec_0), Y(\svec_0))'$ and $(\widetilde{Y}(\tilde\svec_0), \widetilde{Y}(\svec_0))'$, for $\tilde\svec_0 = (-1.33, -0.06)'$ and three choices for the location $\svec_0$: One pair of coordinates is close to $\tilde\svec_0$ (left sub-panel); one pair is far from $\tilde\svec_0$ (right sub-panel); and the last pair is in between these first two (middle sub-panel). Both the marginal and the joint densities are very similar and suggest that Gaussianity of the finite-dimensional distribution has been well-approximated during calibration. Overall, the evidence points to the calibrated SBNN-IL being a very good approximation to the underlying Gaussian process. 

In Figures~\ref{fig:gp_bnn_covs-sv}, \ref{fig:gp_bnn_heatmaps_samples-sv}, and \ref{fig:prior_densities_stationaryGP-sv} in the Supplementary Material, we show the corresponding figures for the calibrated SBNN-VL, which clearly also gives a very good approximation to the underlying process. These results are reassuring as they suggest that the SBNN-VL can model stationary processes well, despite its extra complexity that was largely introduced to model non-stationary processes. 

\newcommand{\contourcaption}{Contour levels (from the outer to the inner contours) correspond to 0.05, 0.10, 0.15, 0.20, 0.25, 0.30, 0.35, 0.40 and 0.45.
}

\subsection{Calibration to non-stationary Gaussian spatial process}\label{sec:NonStatGP}

In this simulation study we consider a non-stationary Gaussian process with mean zero, unit variance, and covariance function
\begin{displaymath}
    C(\svec,\rvec)
    =
    |\bm{\Sigma}(\svec)|^{1/4} |\bm{\Sigma}(\rvec)|^{1/4} \left|\frac{\bm{\Sigma}(\svec) + \bm{\Sigma}(\rvec)}{2}\right|^{-1/2} C^o\left(\sqrt{Q(\svec,\rvec)}\right),\quad \svec,\rvec\in D,
\end{displaymath}
where $C^o(\cdot)$ is the squared-exponential covariogram defined in \eqref{eq:sqexp_covariogram}. The matrix $\bm{\Sigma}(\cdot)\in\mathbb{R}^{d\times d}$ is the so-called kernel matrix \citep{paciorek2006}, and  $\sqrt{Q(\svec,\rvec)}$ is the inter-point Mahalanobis distance, where
\begin{displaymath}
    Q(\svec,\rvec)
    =
    (\svec - \rvec)^\top \left( \frac{\bm{\Sigma}(\svec) + \bm{\Sigma}(\rvec)}{2} \right)^{-1} (\svec - \rvec).
\end{displaymath}
We let the kernel matrix $\bm{\Sigma}(\svec) = \exp(\kappa\|\svec-\bm{\xi}\|)\mathbf{I}$, for $\svec \in D$, where $\bm{\xi}=(0.5, 1)'$ is a point in $D$, and here we set the scaling parameter $\kappa =1$. From Table~\ref{tab:results} we see that the first three models, BNN-IL, BNN-IP, and SBNN-IL, when calibrated do not match the target process as well as the other SBNN variants.

\begin{figure}[t!]
        \includegraphics{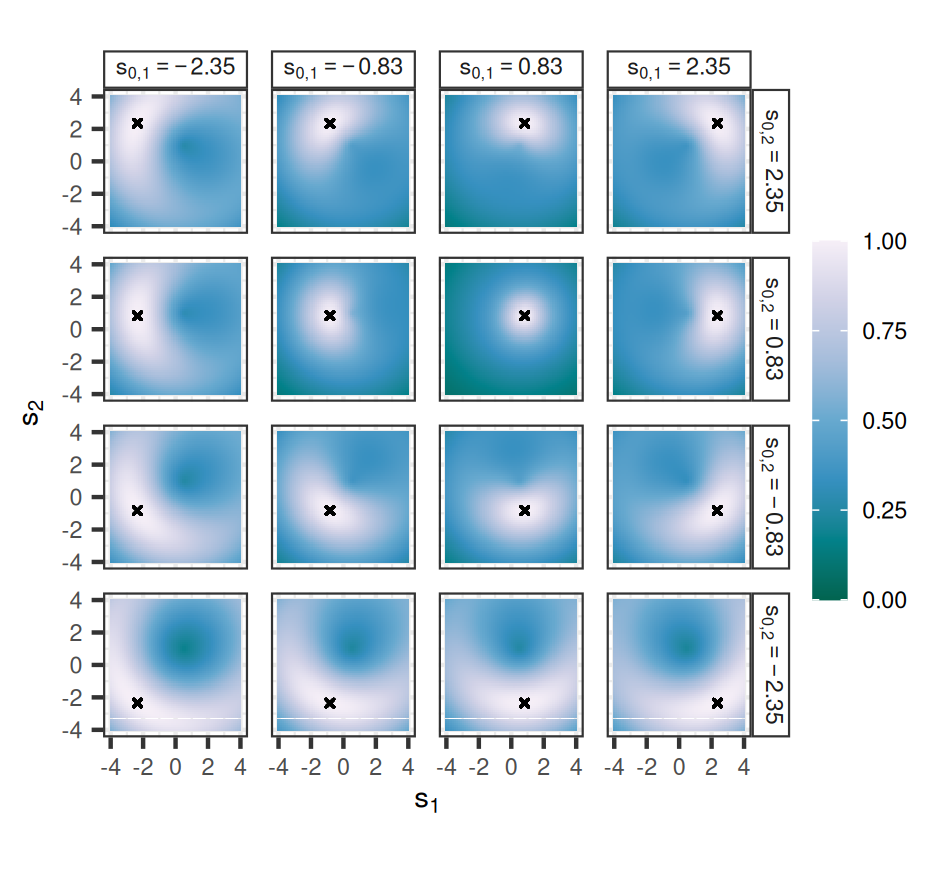}
        \includegraphics{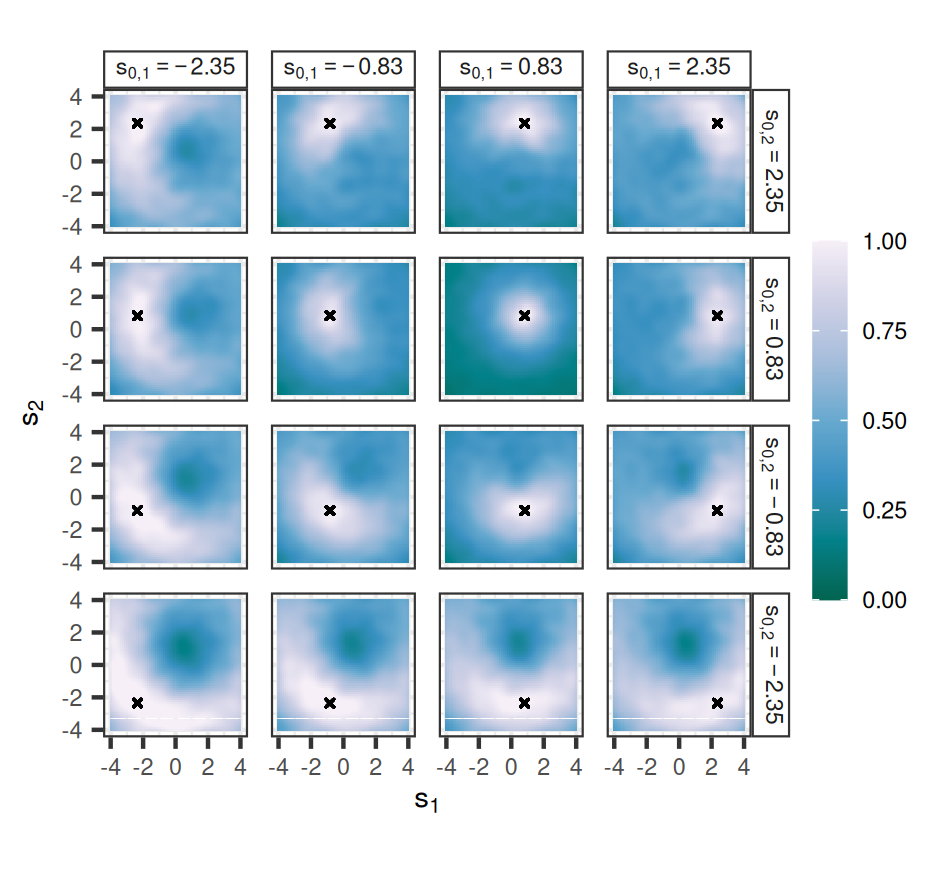}
        
    \caption{(Left panel) Covariance between the target process (non-stationary GP) at 16 grid points (crosses) arranged on a 4 × 4 grid (see Figure~\ref{fig:gp_bnn_heatmaps_samples}) and the target process on a fine gridding $(64 \times 64)$ of $D = [-4, 4] \times [-4,  4]$. (Right panel) Same as left panel, but for the calibrated SBNN-VL.}
    \label{fig:nonstat_cov} 
\end{figure}

\begin{figure}[t!]
  \makebox[0.5\linewidth]{\footnotesize \bf Target}         \\
  \includegraphics{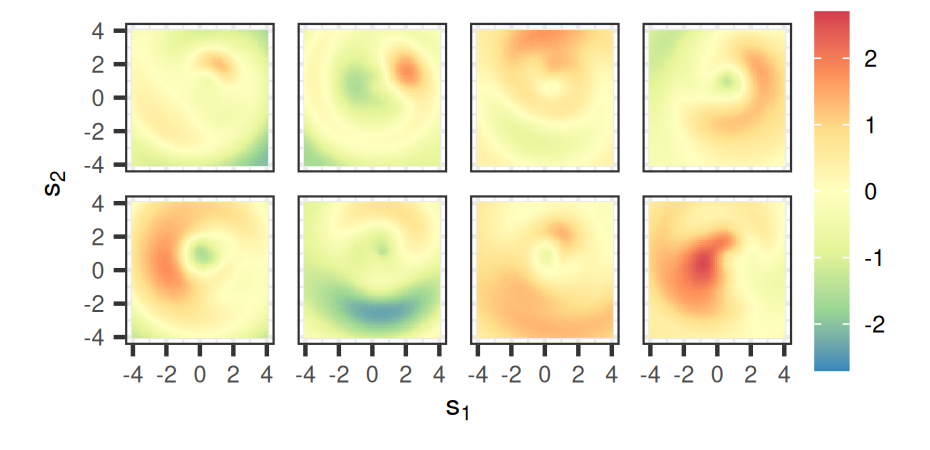}\\
         \makebox[0.5\linewidth]{\footnotesize \bf BNN-IL}          \makebox[0.5\linewidth]{\footnotesize \bf BNN-IP} 
        \includegraphics{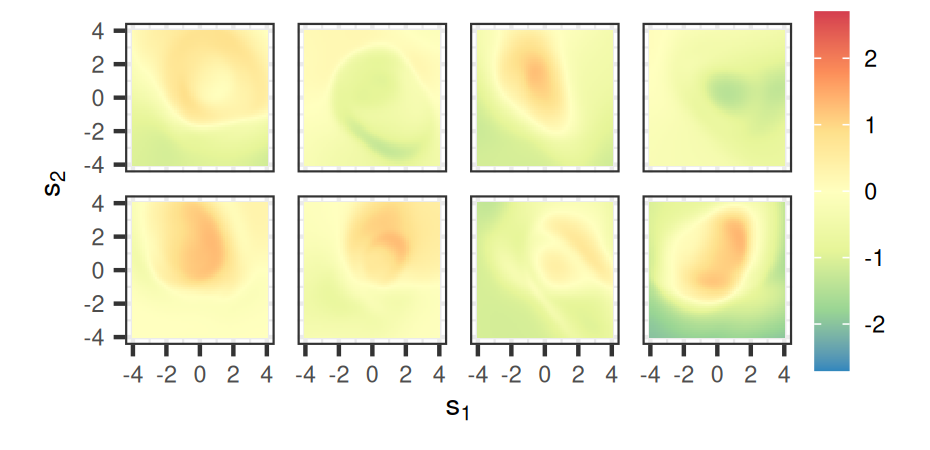}
        \includegraphics{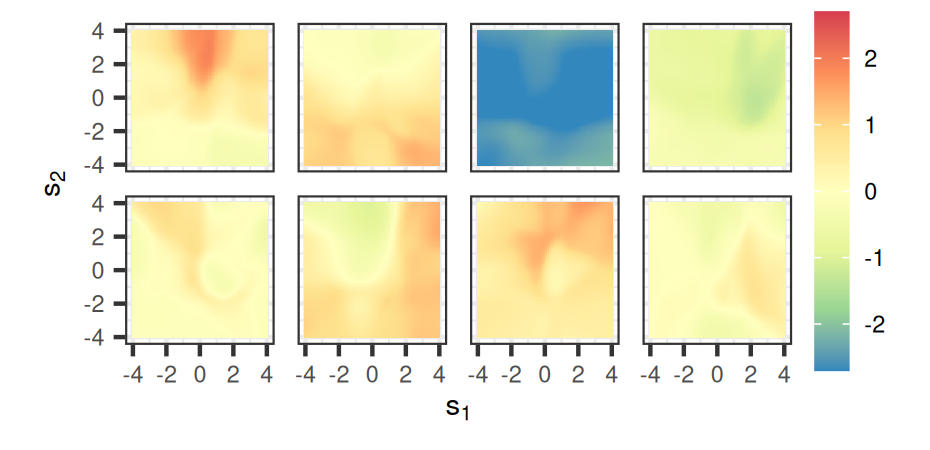}
         \makebox[0.5\linewidth]{\footnotesize \bf SBNN-IL}          \makebox[0.5\linewidth]{\footnotesize \bf SBNN-IP}         
        \includegraphics{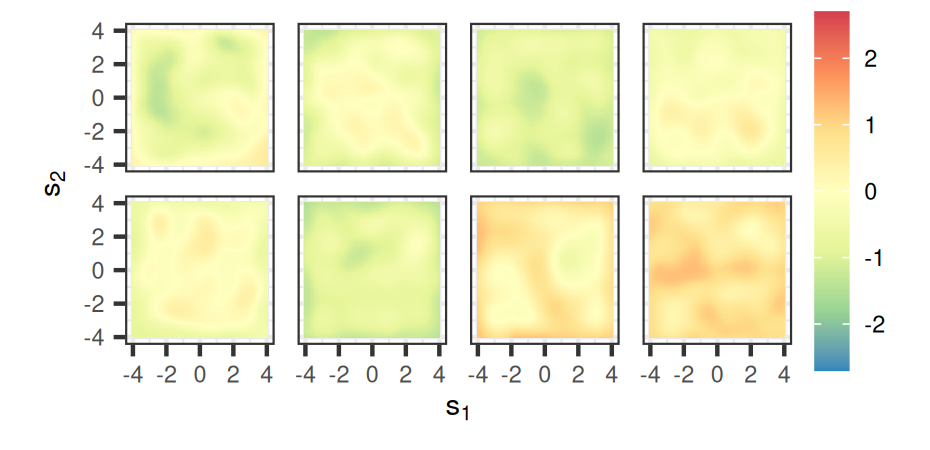}
        \includegraphics{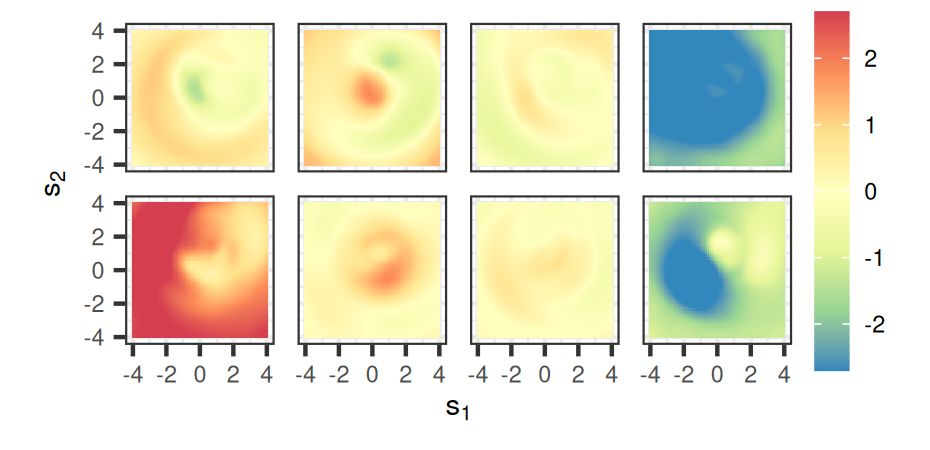}
         \makebox[0.5\linewidth]{\footnotesize \bf SBNN-VL}          \makebox[0.5\linewidth]{\footnotesize \bf SBNN-VP}                
        \includegraphics{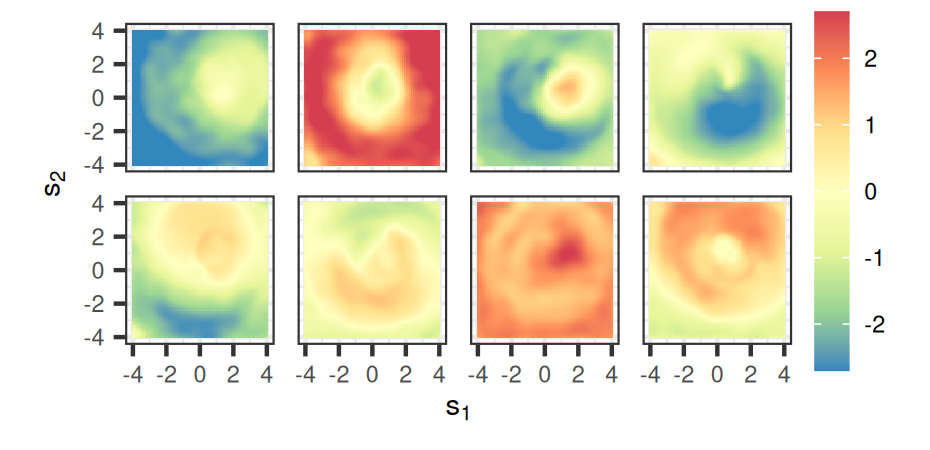}
        \includegraphics{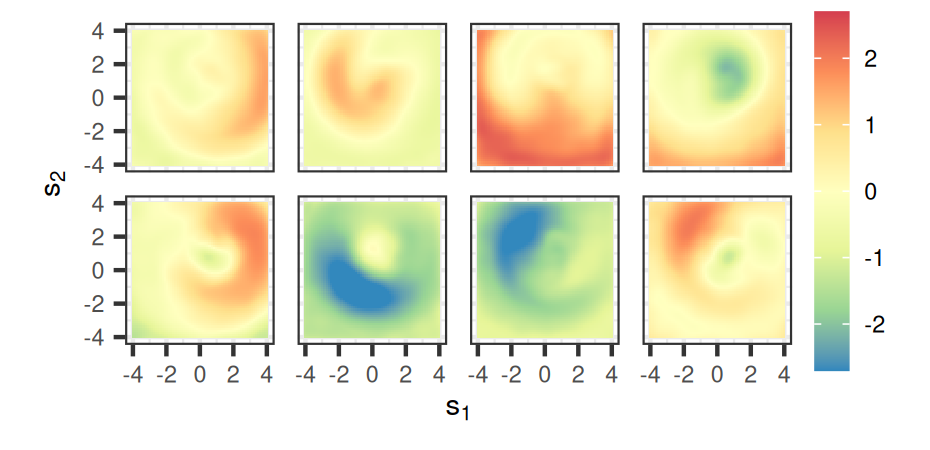}        

    \caption{Eight realisations of the target, non-stationary GP, and all six (S)BNN variants calibrated to the target process.}
    \label{fig:nonstat_samples} 
\end{figure}

In \Cref{fig:nonstat_cov} we plot the true covariance function of the target Gaussian process (left panel), and the empirical covariance function of the calibrated SBNN-VL (right panel), which is very similar to that of the calibrated SBNN-IP (not shown) and that of the calibrated SBNN-VP (not shown). Each heatmap in these panels depicts covariances of the process with respect to the process at a specific spatial location (denoted by the cross). The covariance structure ``revolves'' around the point $\bm{\xi}=(0.5,1)'$, so that it is approximately isotropic near this point, and anisotropic away from the centre. The SBNN-VL clearly captures this covariance structure. The BNN-IL and SBNN-IL, whose analogous plots are shown in Figure~\ref{fig:nonstat_cov2} in the Supplementary Material, clearly do not. The SBNN-IL provides a particularly poor fit to the covariance, suggesting that it should be only used to model stationary processes. In Figure~\ref{fig:nonstat_samples} we plot multiple sample paths from the target process and from all the calibrated models.  The sample paths from the calibrated SBNN-VP are clearly very similar to those of the target process, while those from the calibrated BNN-IL and BNN-IP and the calibrated SBNN-IL are noticeably different. Conclusions from inspecting the sample paths match those given above for the covariance functions. The message is that something `extra' is needed beyond IL for SBNNs: either IP or VL (or both, namely VP). 

\subsection{Calibration to stationary lognormal spatial process} \label{sec:Lognormal}

\begin{figure}[t!]
  \makebox[0.5\linewidth]{\footnotesize \bf Target}         \\
  \includegraphics{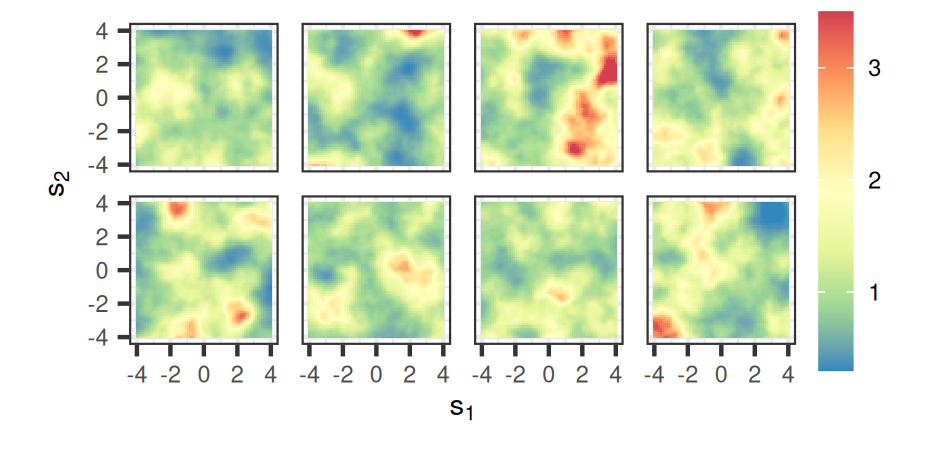}\\
         \makebox[0.5\linewidth]{\footnotesize \bf BNN-IL}          \makebox[0.5\linewidth]{\footnotesize \bf BNN-IP} 
        \includegraphics{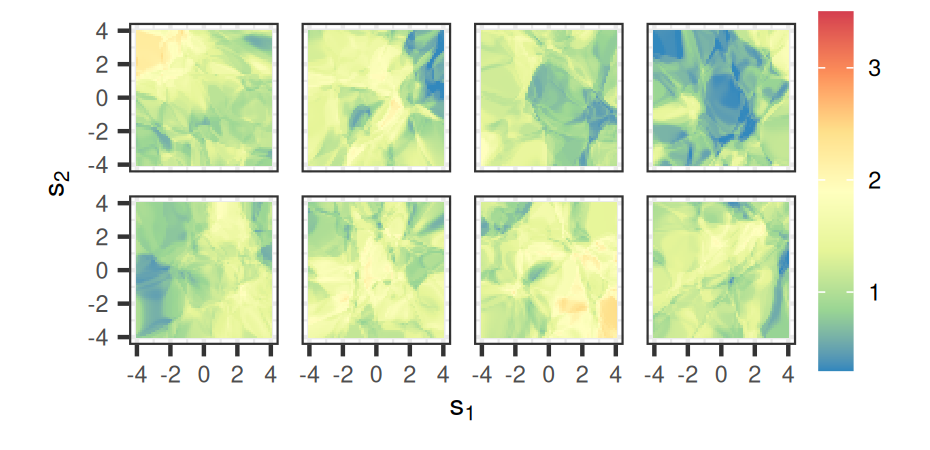}
        \includegraphics{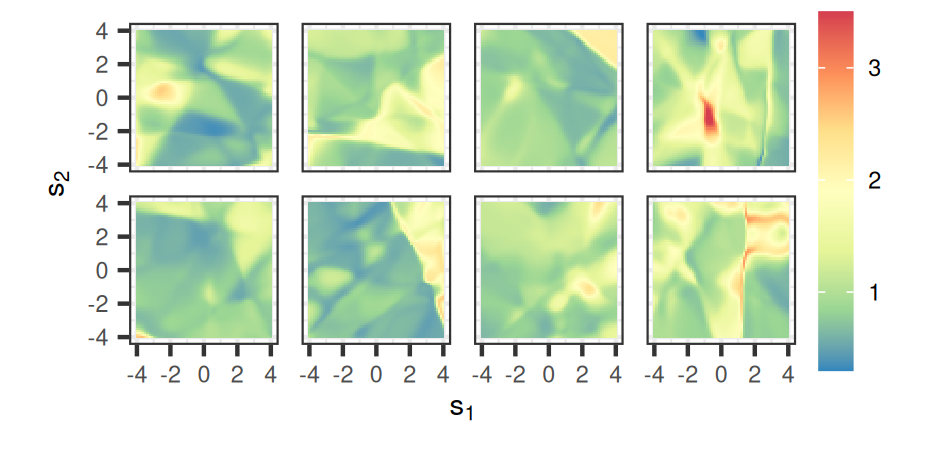}
         \makebox[0.5\linewidth]{\footnotesize \bf SBNN-IL}          \makebox[0.5\linewidth]{\footnotesize \bf SBNN-IP}         
        \includegraphics{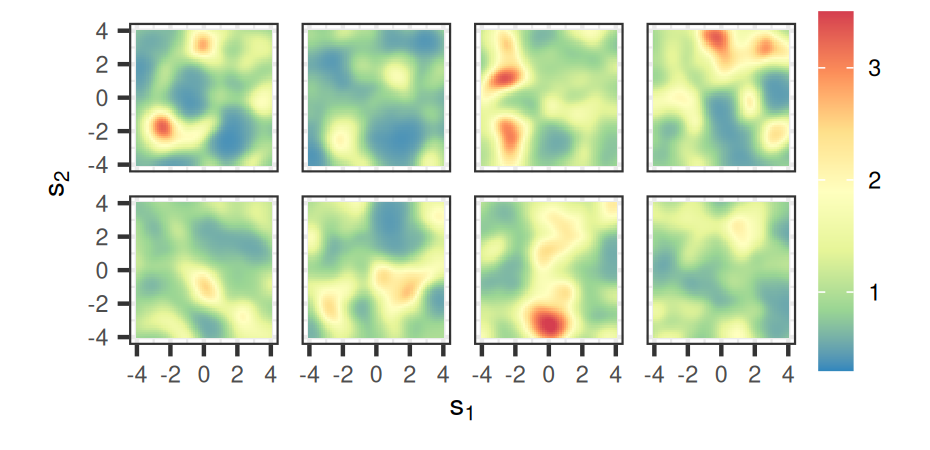}
        \includegraphics{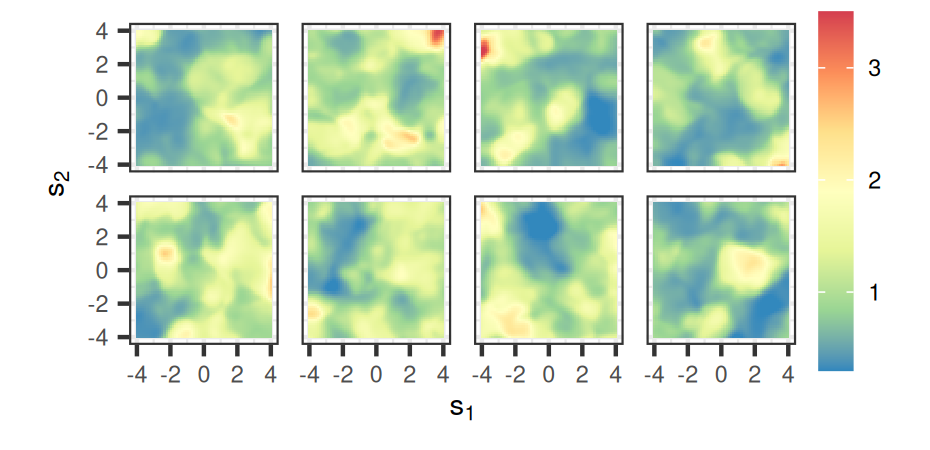}
         \makebox[0.5\linewidth]{\footnotesize \bf SBNN-VL}          \makebox[0.5\linewidth]{\footnotesize \bf SBNN-VP}                
        \includegraphics{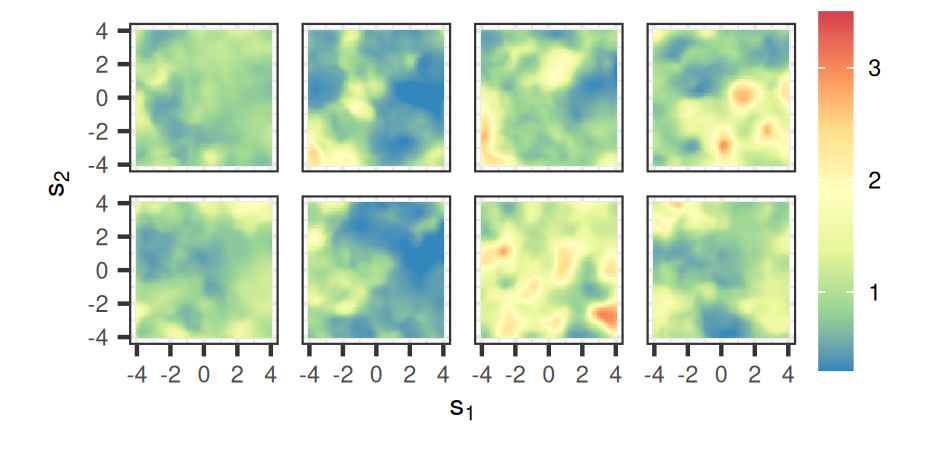}
        \includegraphics{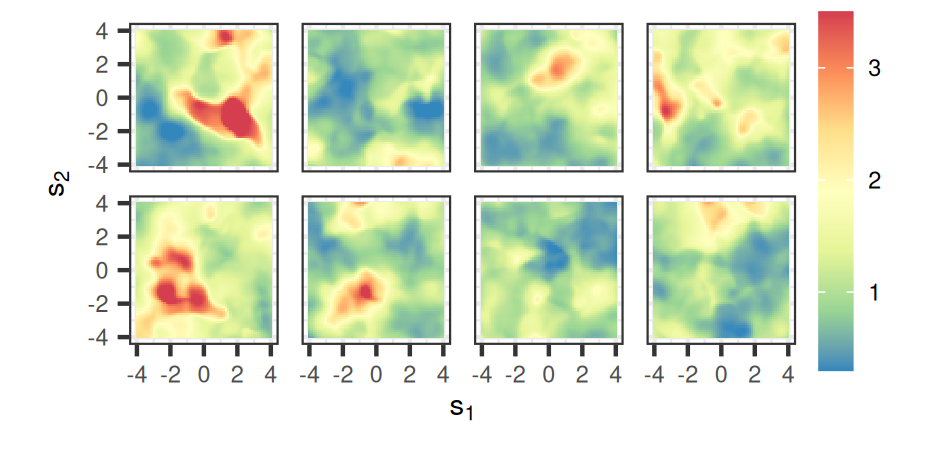}        

    \caption{Eight realisations of the target, stationary, lognormal spatial process, and all six (S)BNN variants calibrated to the target process.}
    \label{fig:lognormal_samples} 
\end{figure}

\newcommand{\contourcaption}{Contour levels (from the outer to the inner contours) correspond to 0.20, 0.40, 0.60, 0.80, 1.00, 1.20, 1.40, 1.60, 1.80, 2.00, 2.20 and 2.40.
}
\begin{figure}[t!]
        
    \includegraphics{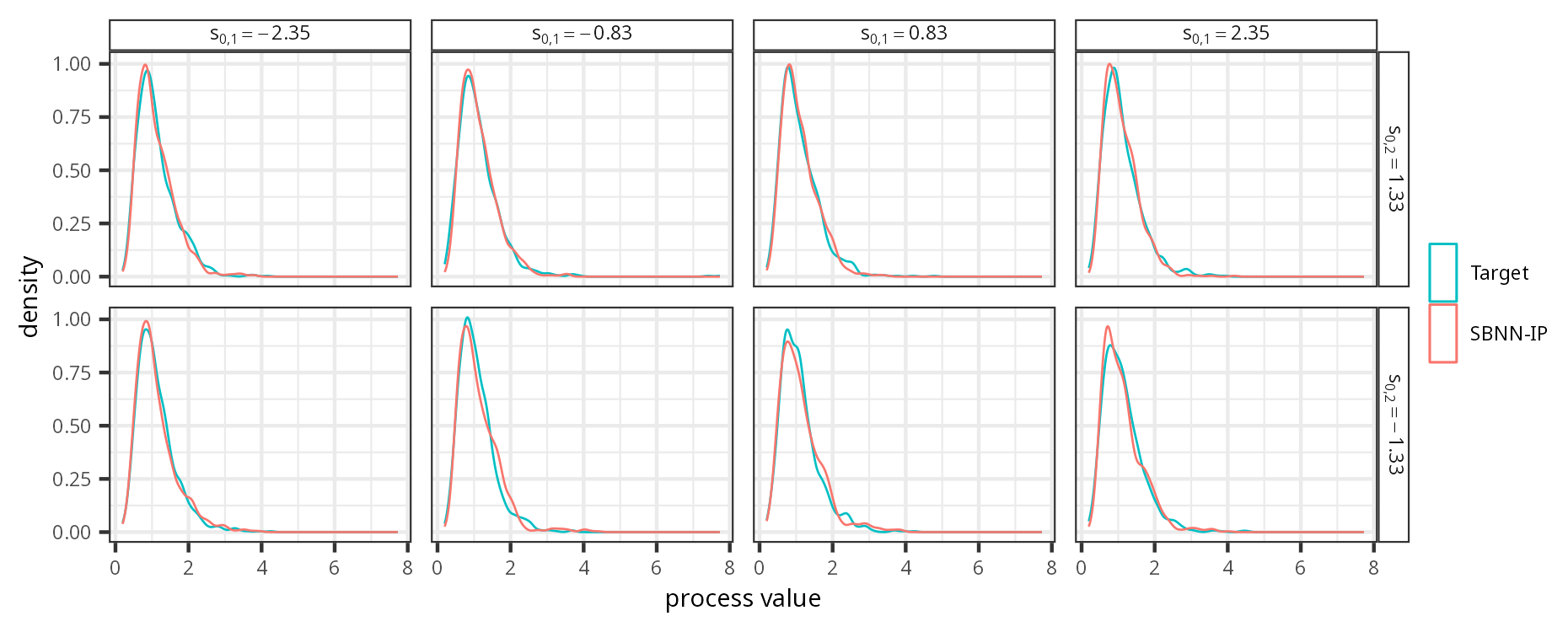} \\
    \includegraphics{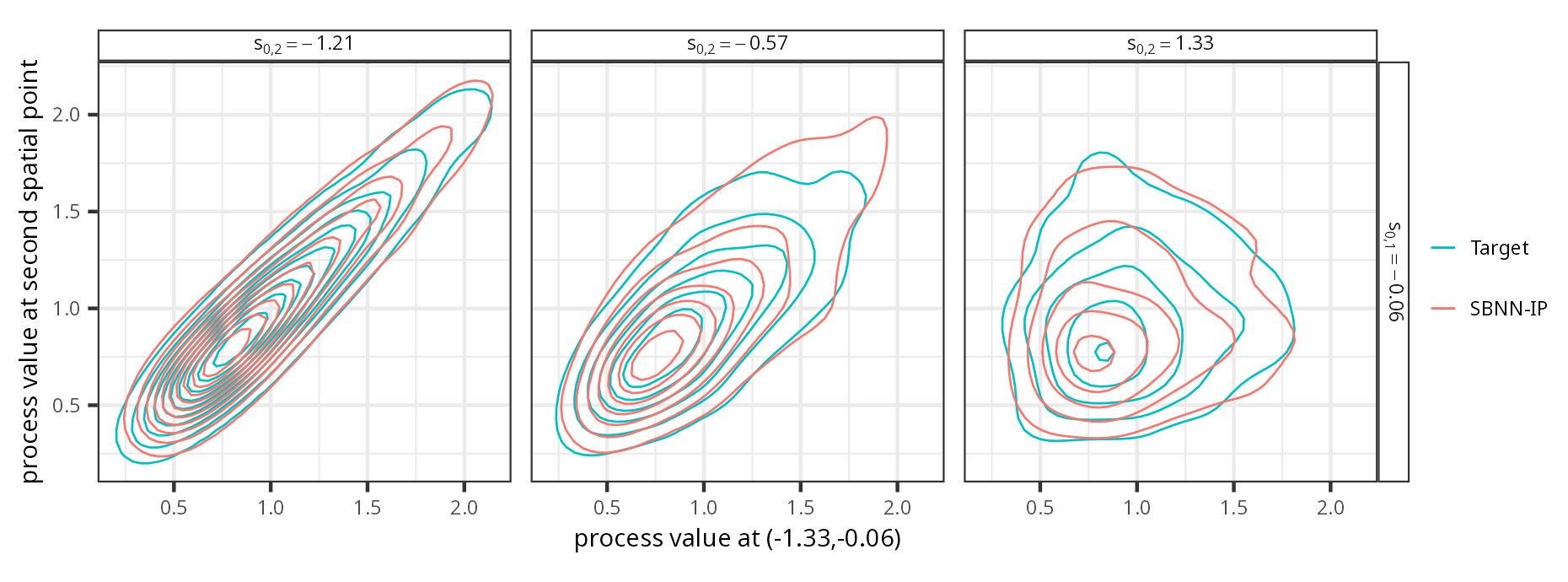} 

    \caption{Kernel density plots from 1000 samples of the SBNN-IP and the target process (stationary lognormal spatial process). (Top panel) Univariate densities of the two processes at eight spatial locations arranged on a $2 \times 4$ grid in $D$ (with coordinates as indicated by the labels of the sub-panels). (Bottom panel) Overlayed bivariate densities of the two processes at $\tilde\svec_0 = (-1.33, -0.06)'$ and three other locations in $D$ (with coordinates as indicated by the labels of the sub-panels). \input{./figures/Section4_3_SBNN-VP_contour_levels.tex}  \label{fig:prior_densities_lognormalGP}}
\end{figure}

We now consider a target stationary lognormal spatial process, specifically a process $\widetilde{Y}(\cdot)$ where $\log(\widetilde{Y}(\cdot))$ is a stationary Gaussian spatial process. Hence, $\widetilde{\Yvec} \sim \textrm{LN}(\zerob, \Sigmamat)$, where $\Xvec \sim \textrm{LN}(\muvec,\Cmat)$ denotes a multivariate lognormal distribution with $\expect[\log(\Xvec)] = \muvec$ and $\Var[\log(\Xvec)] = \Cmat$. In order to demonstrate that our SBNNs can also effectively model processes that are not smooth, we specify the target process covariances using a Mat{\'e}rn covariogram with smoothness parameter equal to 3/2. That is,
$$ \cov(\widetilde{Y}(\svec),\widetilde{Y}(\rvec)) \equiv C^o(\|\svec - \rvec\|) = \left(1 + \frac{\sqrt{3}(\|\svec - \rvec\|)}{\ell}\right)\exp\left(-\frac{\sqrt{3}(\|\svec - \rvec\|)}{\ell}\right),\quad \svec,\rvec \in D, $$
where we set the length scale $\ell = 1$. When calibrating we found it useful to use detrended zero mean realisations of the target process since, without further modification (e.g., exponentiation of the output from the neural network), our (S)BNNs are not designed for positive-only processes. After calibration, the true mean of $\widetilde{Y}(\cdot)$ was then added to the output of our (S)BNNs to ensure that realisations from the calibrated processes have the correct mean. 

We show eight realisations of the target stationary, lognormal spatial process $\widetilde{Y}(\cdot)$ and from the calibrated models in Figure~\ref{fig:lognormal_samples}. The four SBNNs are clearly able to capture the stochastic properties of the target process, with the realisations from the calibrated SBNN-IP and SBNN-VP appearing to be qualitatively more similar to those of the target process than those from the calibrated SBNN-IL and SBNN-VL. Realisations from the calibrated non-spatial BNNs are clearly very different to what one could expect from the target process. This corroborates Table~\ref{tab:results}, where we see that the `prior-per-parameter' SBNN variants have a relatively low Wasserstein distance $W_1(\cdot)$. Note that $W_1(\cdot)$ for the non-spatial BNN-IP is similar to that for the SBNN-VL, and yet the eight sample paths from the BNN-IP are clearly not similar to those from the target process, indicating the importance of following up the calibration with other diagnostics such as inspection of sample paths. 

Figure~\ref{fig:prior_densities_lognormalGP} is analogous to Figure~\ref{fig:prior_densities_stationaryGP} but for the SBNN-IP (the `best' model according to Table~\ref{tab:results}) and the lognormal process. That is, the figure shows marginal and bivariate kernel density estimates at selected points in $D$ from 1000 samples taken from the calibrated SBNN-IP and the target lognormal process. As in the Gaussian case, both the marginal and the joint densities are similar. Density plots for the SBNN-VL, shown in Figure \ref{fig:prior_densities_lognormalGP-sv} in the Supplementary Material, show that the SBNN-V with the `prior-per-layer' scheme is well-calibrated, but that it is unable to capture the lower tails well. On the other hand, the SBNN-VP has the required flexibility to model the tails well; see Figure~\ref{fig:prior_densities_lognormalGP-svp} in the Supplementary Material.

Overall, these results suggest that our SBNNs are able to model non-Gaussian processes, and that they potentially could be applied to a wider class of models than those considered here, notably models whose likelihood function is intractable but which can be simulated from relatively easily \citep[e.g., models of spatial extremes,][]{Davison_2015}. 

\section{Making inference with SBNNs}\label{sec:inference}

Once an SBNN has been calibrated, it serves two purposes: (i) to efficiently simulate realisations of an underlying stochastic process; and (ii) to make inference conditional on observational data. Unconditional simulation is straightforward to do once an SBNN is calibrated, by using one of the model specifications outlined in Section \ref{sec:architecture} with the free hyper-parameters, $\psib$,  replaced with the optimised ones, $\psib^*$. In this respect the SBNN could be used as a surrogate for a computationally-intensive stochastic simulator, such as a stochastic weather generator \citep{Semenov_1998, Kleiber_2023}. \Copy{UnCondSim}{We note that there are several other neural network architectures that are also well suited for unconditional simulation, such as variational autoencoders \citep[VAEs,][]{Kingma_2013} and generative adversarial networks \citep[GANs,][]{Goodfellow_2014}.} Inference, on the other hand, requires further computation. Specifically, given a data set $\Zvec \equiv (Z_1,\dots,Z_m)'$ of noisy measurements of $Y(\cdot)$ collected at locations $\tilde\svec_1,\dots,\tilde\svec_m \in D$, inference proceeds by evaluating, approximating, or sampling from, the posterior distribution of the weights and biases; namely, the distribution of the neural network parameters conditional on $\Zvec$ and the calibrated hyper-parameters $\psib^*$. After sampling from the posterior distribution of the weights and biases, it is straightforward to obtain samples from the posterior distribution of $Y(\cdot)$, which we refer to as the predictive distribution.  \Copy{VAELim}{An attraction of the SBNN's process model definition over other generative models such as VAEs and GANs is that it can be readily incorporated into hierarchical models typically seen in spatial statistics. As a result, issues concerning missing data, noise, or even uncertain data models, can be dealt with in a straightforward manner when making inference on the process.}

Several inferential methods have been developed for BNNs \citep[e.g.,][]{jospin2022}. These include variational inference \citep[e.g.,][]{zammit2021} and MCMC. Among MCMC techniques, Hamiltonian Monte Carlo \citep{neal1996} is the most widely used. The original HMC algorithm of \citet{neal1996} was \emph{full-batch}; that is, it used the entire dataset $\Zvec$ for generating each posterior draw. However, the gradient computations required for HMC become computationally infeasible for large datasets.
\citet{chen2014} proposed to resolve these computational limitations by approximating the gradient at each MCMC iteration using a mini-batch of the data. The resulting gradient approximation is termed a stochastic gradient, and the corresponding approximation to HMC is termed {stochastic gradient Hamiltonian Monte Carlo} (SGHMC). SGHMC and its adaptive variant \citep{springenberg2016} are well-suited for making inference with BNNs. It is straightforward to apply SGHMC to the SBNN-I variants since one can readily make use of available software for BNNs. For illustration, in this section we provide predictive distributions using SGHMC for $Y(\cdot)$ calibrated to the stationary Gaussian distribution of Section~\ref{sec:StatGP} and to a more involved stationary max-stable process where obtaining predictive distributions from many data points is extremely challenging using conventional techniques. 

\subsection{Case study 1: GP target process}\label{sec:CaseStudy1}

\begin{figure}[t!]
    \begin{center}
      \includegraphics{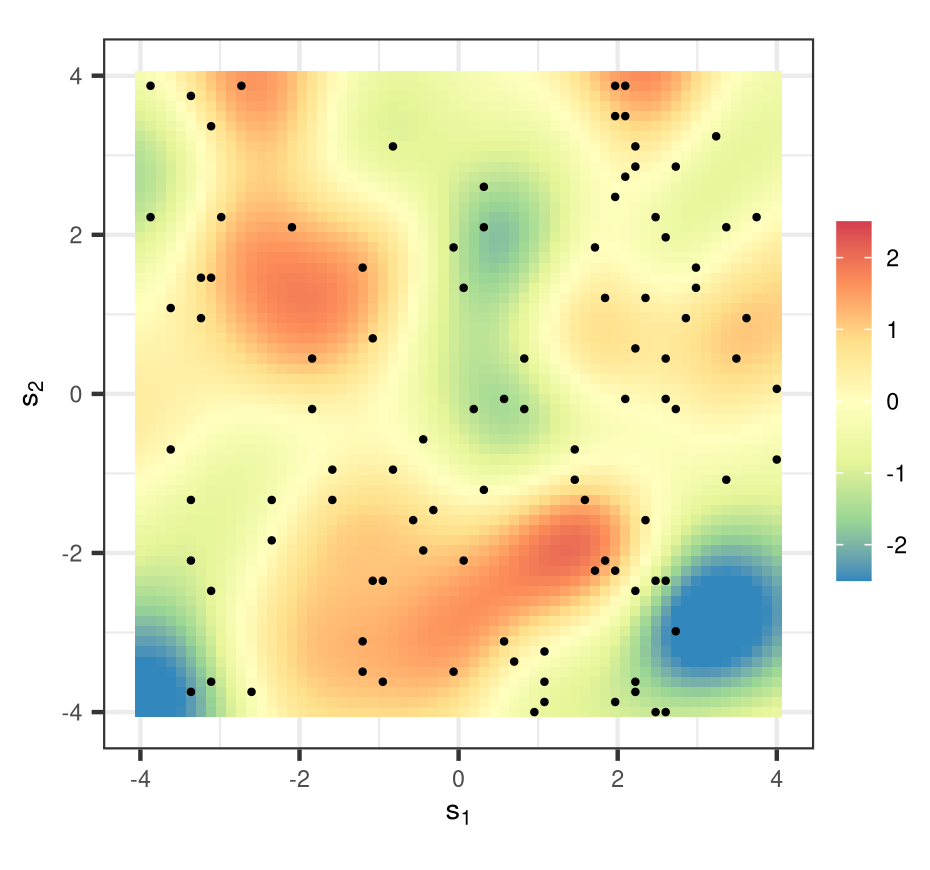}  \includegraphics{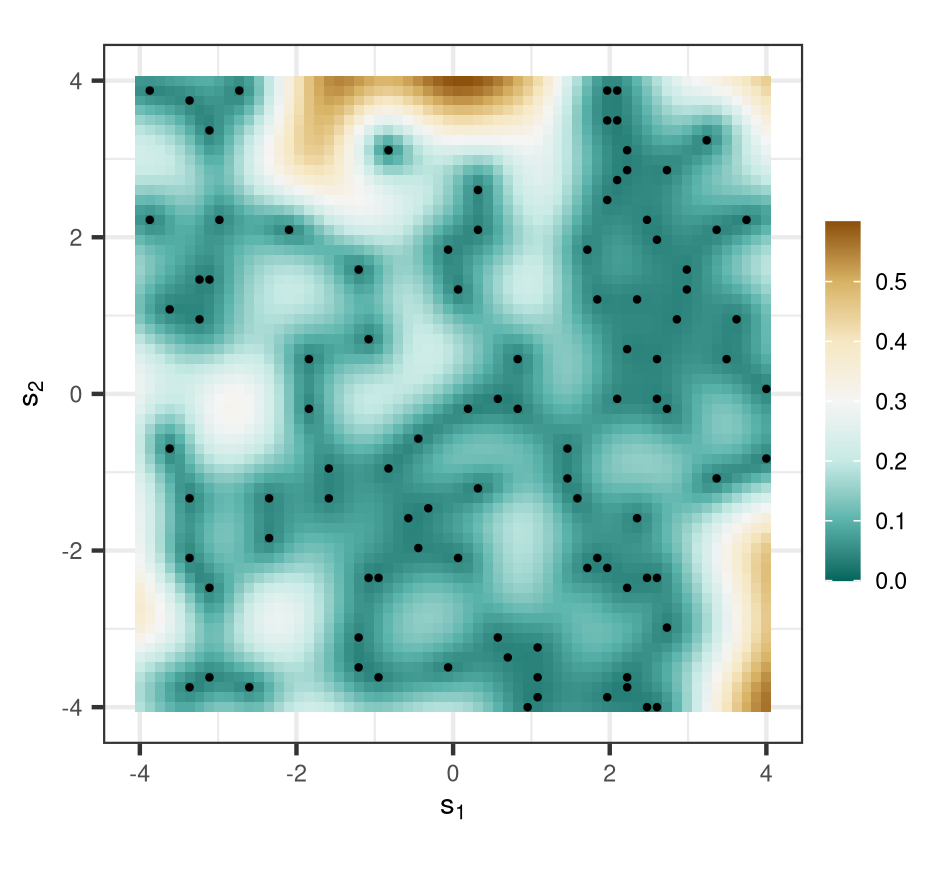} \\
   \includegraphics{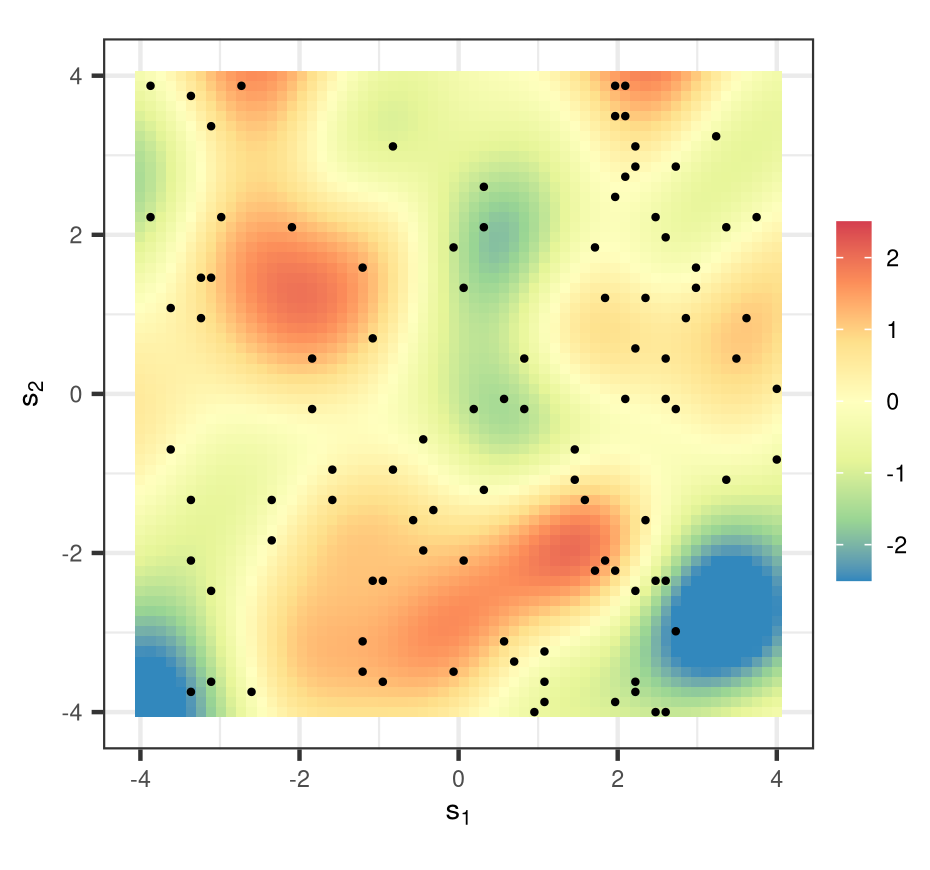}   \includegraphics{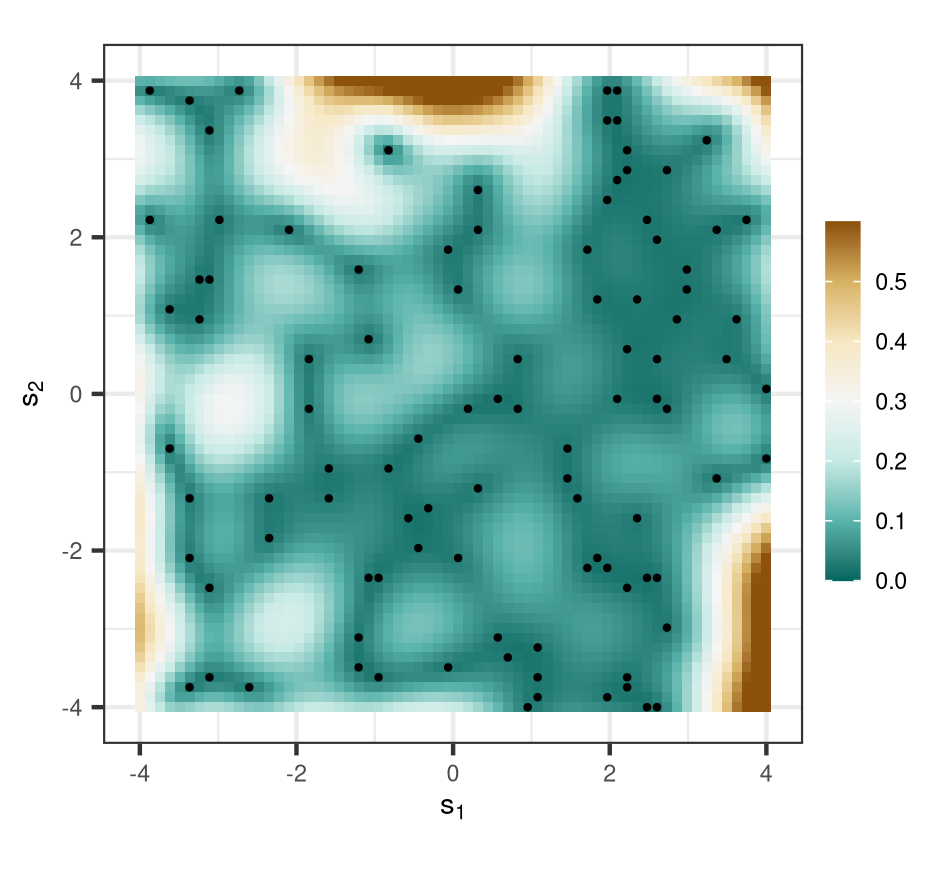}
    \end{center}
    \caption{The predictive means (left panels) and predictive standard errors (right panels) of the process evaluated at locations on a fine gridding $(64 \times 64)$ of $D = [-4, 4] \times [-4, 4]$. The data have locations denoted by the black dots. (Top panels) Predictive distribution from the calibrated SBNN-IL, found using stochastic gradient Hamiltonian Monte Carlo. (Bottom panels) The true predictive distribution evaluated from the target process (stationary GP) using multivariate Gaussian distribution formulas.}
    \label{fig:posteriors_2d}
\end{figure}

\newcommand{\contourcaption}{Contour levels (from the outer to the inner contours) correspond to 2.00, 4.00, 6.00, 8.00 and 10.00.
}
\begin{figure}[t!]
        
    \includegraphics{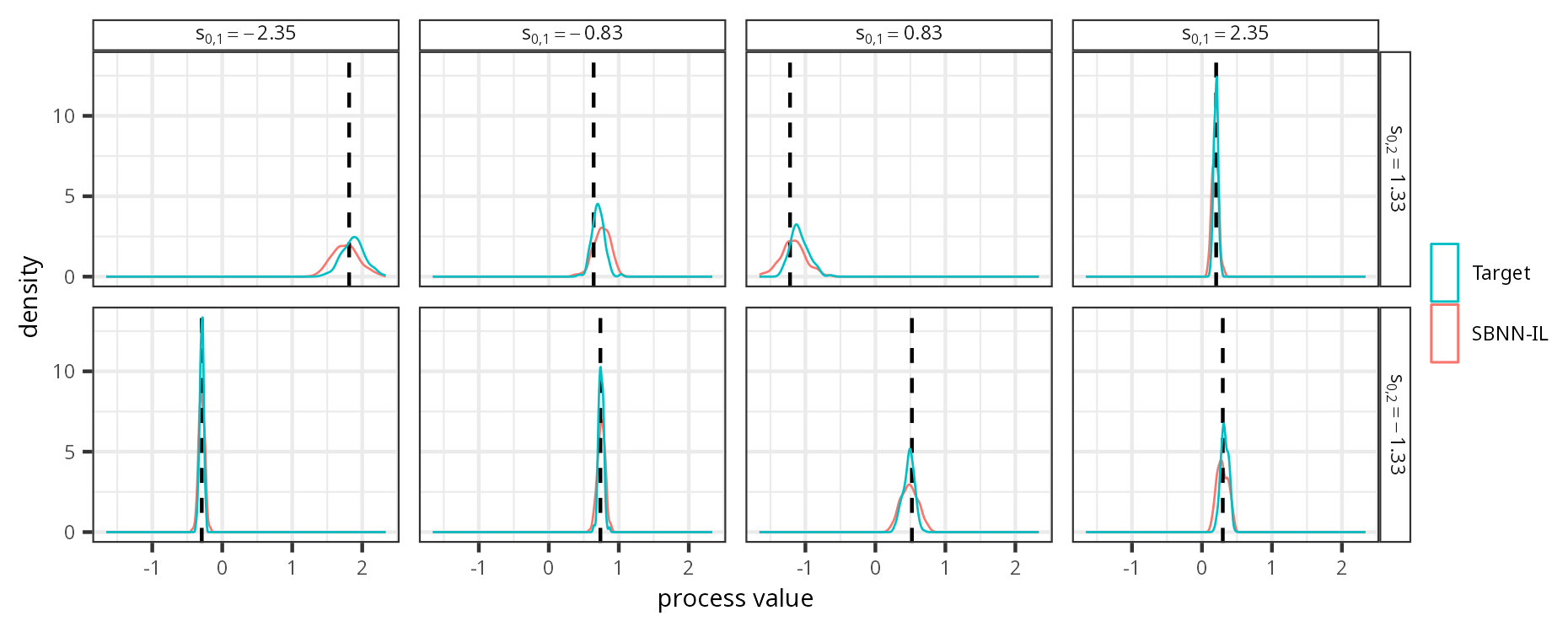} \\
    \includegraphics{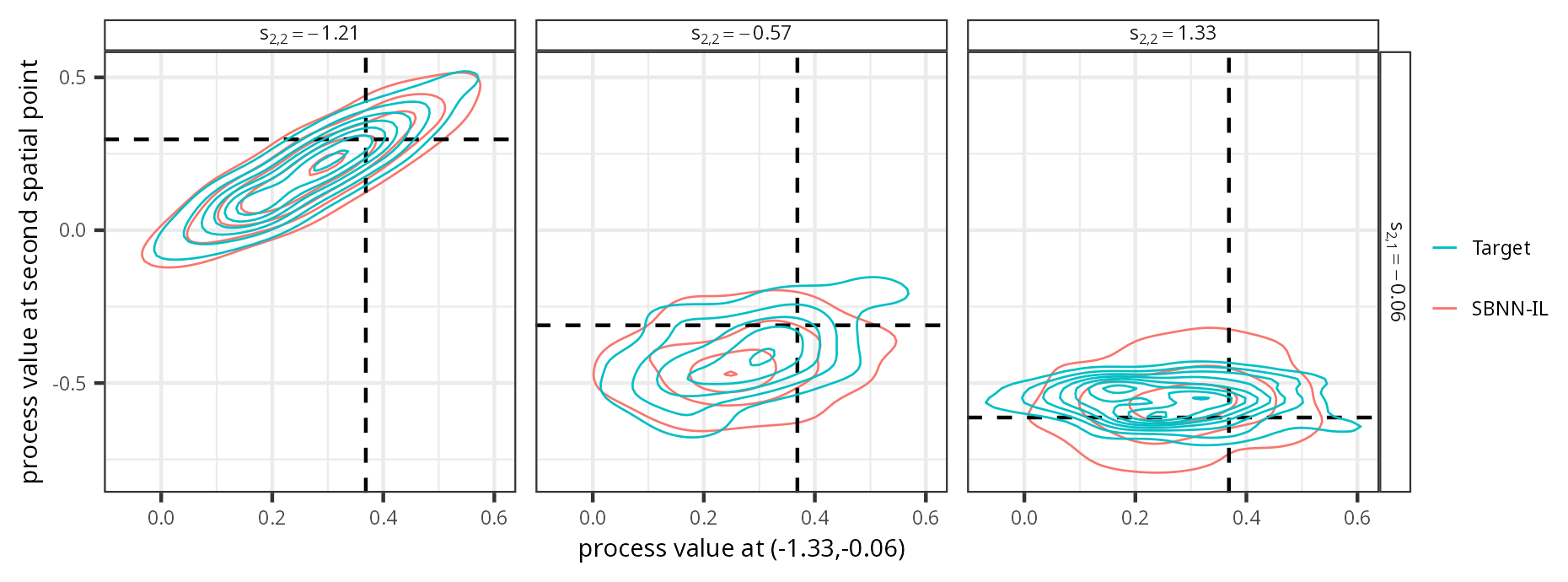} 

    \caption{Kernel density plots from 800 samples from the predictive distribution of the SBNN-IL and that of the target process (stationary GP). (Top panel) Marginal predictive densities of the two processes at eight spatial locations arranged on a $2 \times 4$ grid in $D$ (with coordinates as indicated by the labels of the sub-panels). (Bottom panel) Overlayed bivariate predictive densities of the two processes at $\svec = (-1.33, -0.06)'$ and three other locations in $D$ (with coordinates as indicated by the labels of the sub-panels). \input{./figures/Section4_3_SBNN-VP_contour_levels.tex} True process values are denoted by the black dashed lines. \label{fig:posterior_densities_stationaryGP}}
\end{figure}

In this section we consider the SBNN-IL $Y(\cdot)$ that was calibrated to the stationary GP $\widetilde{Y}(\cdot)$ in Section~\ref{sec:StatGP}. The top two panels of Figure~\ref{fig:posteriors_2d} show the predictive mean (posterior mean of $Y(\cdot)$) and the predictive standard error (posterior standard deviation of $Y(\cdot)$) of the calibrated SBNN-IL, when conditioned on simulated data $\Zvec$ with sample locations denoted by the black dots, as obtained using SGHMC. Here, the data model for $\Zvec$ is given by $Z_i = Y(\tilde\svec_i) + \epsilon_i$, for $i= 1,\dots,m$, where $m = 100$, $\epsilon_i \sim \textrm{Gau}(0, 0.001)$, and where $\tilde\svec_i \in D$ are the sample locations. The predictive means and the predictive standard errors under the true model (from which the data are simulated) are shown in the bottom two panels of Figure~\ref{fig:posteriors_2d}. Note how these quantities are similar in pattern; this was somewhat expected since, as shown in Section~\ref{sec:StatGP}, the SBNN-IL is well-calibrated to the true underlying process.  In Figure~\ref{fig:posterior_densities_stationaryGP} we plot kernel density estimates from 800 MCMC samples taken from the predictive distribution under the calibrated SBNN-IL and the true (also target) stationary GP. The top and bottom panels show empirical marginal predictive densities and joint bivariate predictive densities corresponding to the predictive distributions of the processes at the same spatial locations considered in Figure~\ref{fig:prior_densities_stationaryGP}. The marginal and joint predictive densities of the SBNN-IL appear to be unimodal and Gaussian; this is reassuring since any finite-dimensional distribution of the target process conditioned on the data $\Zvec$ is indeed Gaussian. On the other hand, there is a small discrepancy in the shapes of these posterior densities, with the SBNN-IL appearing to yield slightly larger posterior variances. This discrepancy could be due to several reasons. First, it could be because not all the finite-dimensional distributions of the stochastic process being targeted were considered when minimising Wasserstein distance. Second, it is possible that the SBNN needs to be made more flexible (e.g., have a larger embedding layer) to more faithfully represent the finite-dimensional distributions of the underlying process. Third, it could be that the MCMC chains need more time to converge: Although we ran four parallel MCMC chains for 300,000 iterations in each chain, excluding 100,000 samples as burn-in and thinning by a factor of 1,000, our effective sample size ranged between 50--400 at several prediction locations. Despite these drawbacks however, the predictive distribution from the SBNN-IL is clearly very similar to what one would have obtained under the true model.

We repeated the experiment above for $m = 1,000$ and $m = 5,000$. In Table~\ref{tab:Scoring} we score the SBNN-IL using conventional prediction diagnostics, and compare them to those of the true process and a BNN-IL that is uncalibrated (i.e., with standard normal priors on the weights and biases). The differences in Table~\ref{tab:Scoring} between the calibrated SBNN-IL and the uncalibrated BNN-IL are largest for the small $m = 100$ case, where the (prior) model plays a bigger role in prediction than in the large $m$ case.  In Figure~\ref{fig:MCMC_traces_SBNN-IL} we plot samples from the predictive distribution at spatial locations regularly spaced on a $12\times 12$ gridding of our spatial domain; the 800 samples constitute 200 from each of the four MCMC chains. There are no obvious problems with convergence of the chains, and the predictive distributions for the three cases of $m$ have similar expectations, as one would expect. Since SGHMC uses mini-batches of a fixed size, and because we kept the SBNN architecture unchanged, the time needed to obtain these samples for different $m$'s was the same. This insensitivity to $m$ and, indeed, the underlying `true' model, is a major strength of this approach. The time required to obtain representative samples in this case was just under one hour, which one might argue is substantial given that the underlying process is a GP. However, as we show next, obtaining predictive distributions of SBNNs can be done in the same amount of time for spatial models where prediction is computationally difficult or impossible.

\begin{table}[t!]
  \caption{Scoring the prediction distribution of the true model (GP), the calibrated SBNN-IL, and an uncalibrated BNN-IL using the mean absolute prediction error (MAPE), the root mean squared prediction error (RMSPE), and the continuous ranked probability score (CRPS). Scores are calculated using samples from the predictive distribution, and the true (simulated) process, for different sample size, $m$.\label{tab:Scoring}}
  \centering
  \begin{tabular}{ccccc}
    Model & $m$ & MAPE & RMSPE & CRPS \\ 
  \hline
GP (True Model) &  100 & 0.087 & 0.146 & 0.061 \\ 
  SBNN-IL (Calibrated) &  100 & 0.094 & 0.155 & 0.067 \\ 
  BNN-IL (Uncalibrated) &  100 & 0.144 & 0.270 & 0.112 \\ 
\hline
  GP (True Model) & 1000 & 0.011 & 0.014 & 0.008 \\ 
  SBNN-IL (Calibrated) & 1000 & 0.013 & 0.017 & 0.010 \\ 
  BNN-IL (Uncalibrated) & 1000 & 0.016 & 0.021 & 0.012 \\ 
\hline
  GP (True Model) & 5000 & 0.005 & 0.007 & 0.004 \\ 
  SBNN-IL (Calibrated) & 5000 & 0.007 & 0.009 & 0.005 \\ 
  BNN-IL (Uncalibrated) & 5000 & 0.008 & 0.011 & 0.007 \\ 
   \hline

  \end{tabular}
\end{table}

\subsection{Case study 2: Max-stable target process}\label{sec:CaseStudy2}

In this section we consider Schlather's max-stable model \citep{Schlather_2002} constructed using a stationary power exponential covariance function with range parameter equal to 3 and smoothness parameter equal to 3/2. We generate samples from the process using the \texttt{SpatialExtremes} package in \texttt{R} \citep{SpatialExtremes}, log-transform the data so that they have Gumbel marginals (for variance stabilisation), and calibrate an SBNN-IP with the log-transformed data. As data model we use $\log Z_i = \log Y(\tilde\svec_i) + \epsilon_i$, for $i= 1,\dots,m$, where $m = 50$, $\epsilon_i \sim \textrm{Gau}(0, 0.001)$, and where $\tilde\svec_i \in D$ are sample locations. This data model is multiplicative on the original scale.

\begin{figure}[t!]
        \centering
    \includegraphics{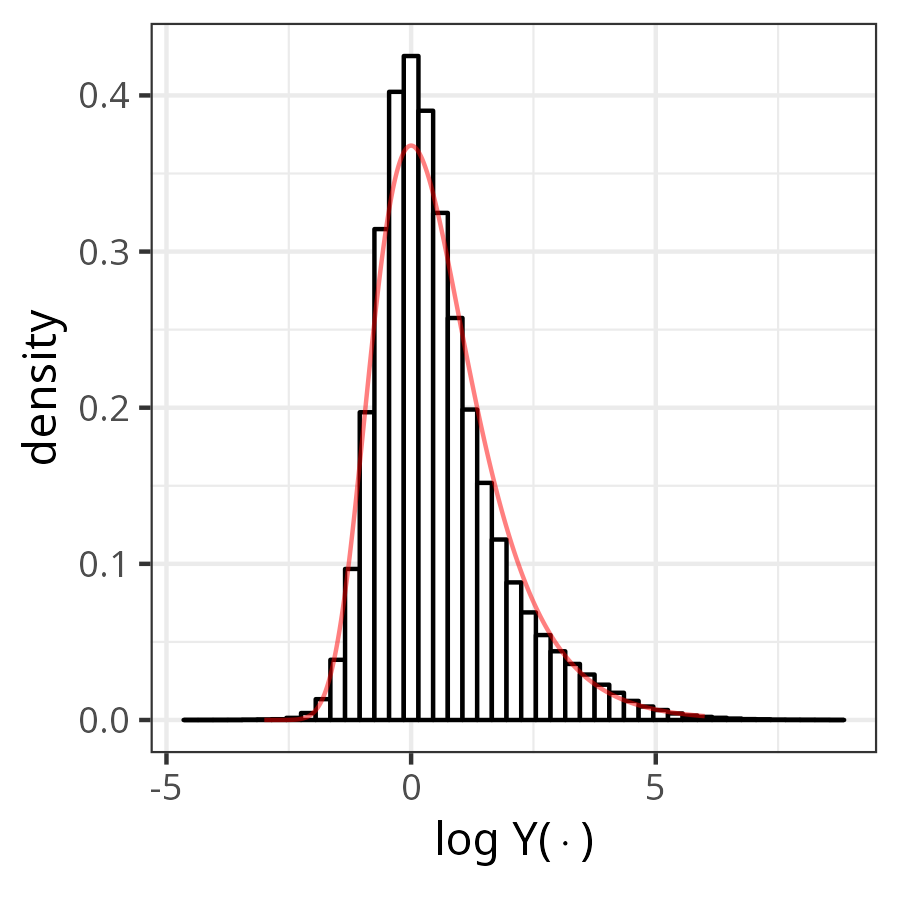} \hfill    \includegraphics{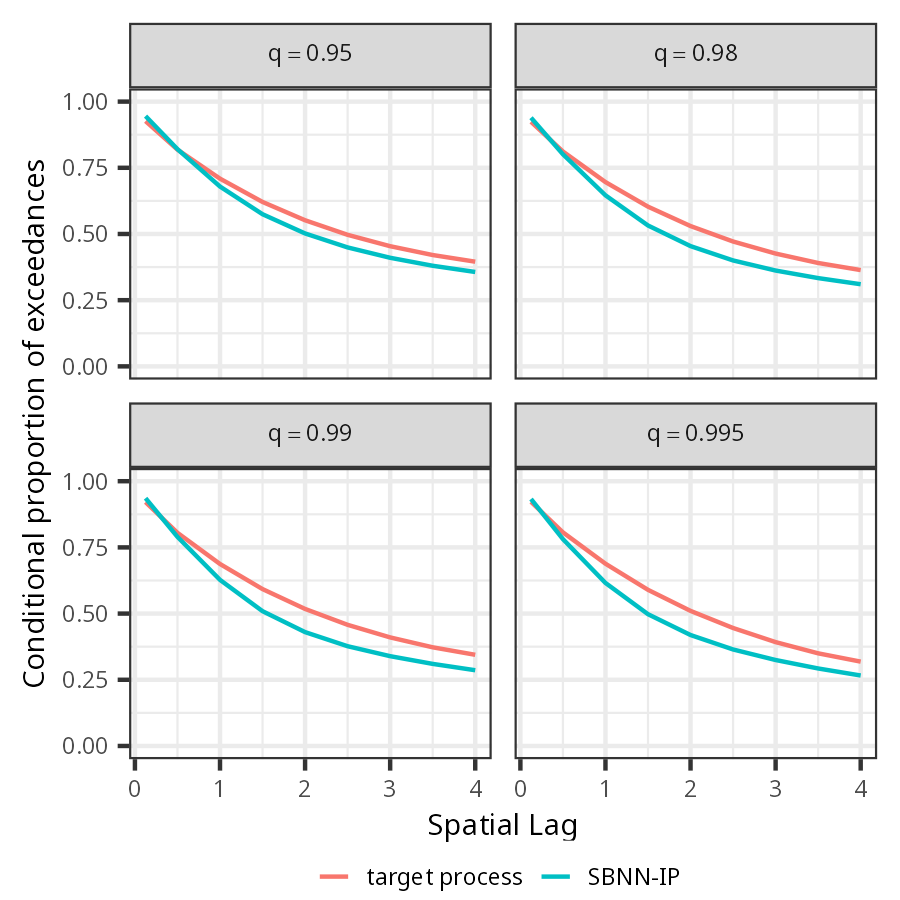} 

    \caption{(Left panel) Histogram of samples of $\log Y(\cdot)$ from the calibrated SBNN-IP (bins) and the standard Gumbel density (red curve).  (Right panel) Conditional probability of exceedance ${P(\log Y(\svec_i) > y\mid \log Y(\svec_j) > y)}$ against the spatial lag $\| \svec_j - \svec_i\|$, where $y$ corresponds to the 0.95, 0.98, 0.99 and the 0.995 quantiles $q$ of the standard Gumbel distribution.\label{fig:extremes_EDA}}
\end{figure}

For this case study we increased the number of basis functions in the embedding layer from ${K = 15^2}$ to ${K = 20^2}$, which led to better fits (all other settings were left unchanged). In Figure~\ref{fig:extremes_EDA}, left panel, we show the marginal distribution of the calibrated SBNN-IP, which follows the standard Gumbel distribution, as one would expect following calibration. On the right panel we plot empirical conditional exceedance probabilities as a function of spatial lag. Specifically, we plot empirical estimates of $P(\log Y(\svec_i) > y\mid \log Y(\svec_j) > y)$ against the spatial lag $\| \svec_j - \svec_i\|$, where $y$ corresponds to the 0.95, 0.98, 0.99 and the 0.995 quantiles of the standard Gumbel distribution; such a plot is often used to diagnose whether a model appropriately captures extremal dependence. The calibrated SBNN-IP appears to model the spatial dependence in the extrema reasonably well.

We proceed now to make inference with the calibrated SBNN-IP. Because it was not possible to generate conditional simulations using the package \texttt{SpatialExtremes} for large sample sizes, in Table~\ref{tab:Max-Stable_results} we only give results for $m = 50$ for the calibrated SBNN-IP, an uncalibrated BNN-IL, and the true model. We see that in this small-sample-size scenario the calibrated SBNN-IP is considerably better than an uncalibrated BNN-IL, and it is very close in performance to the true model. Importantly, since the compute time required to make predictions with (S)BNNs is largely independent of $m$ and the true underlying model, we again only need approximately one hour to obtain representative samples from the predictive distribution with the SBNN-IP and BNN-IL (we tested up to $m = 5,000$). For comparison, \texttt{SpatialExtremes} required several hours of compute time to obtain 30 samples from the predictive distribution with $m = 50$, and we were unable to draw any such samples using \texttt{SpatialExtremes} with $m \ge 100$.

\begin{table}[t!]\caption{Scoring the predictive distribution of the true model (max-stable process with Gumbel marginals), the calibrated SBNN-IP, and an uncalibrated BNN-IL using the mean absolute prediction error (MAPE), the root mean squared prediction error (RMSPE), and the continuous ranked probability score (CRPS). Scores are calculated using samples from the predictive distribution, and the true (simulated) process, for sample size $m = 50$.\label{tab:Max-Stable_results}}
  \centering
  
  \begin{tabular}{c|c|c|c|c}
  Model   &         $m$   & MAPE & RMSPE &  CRPS \\ \hline
Max-Stable (True Model)& 50 & 0.129 & 0.175 & 0.0938 \\ 
SBNN-IP (Calibrated) & 50 & 0.129 & 0.173 & 0.102 \\
    BNN-IL (Uncalibrated)  & 50 & 0.148 & 0.208 & 0.120 \\ \hline                                               

  \end{tabular}

\end{table}

\section{Conclusion}\label{sec:conclusion}

The proposed spatial-statistical methodology that calibrates SBNNs to a target spatial process differs starkly from current approaches where one uses a parametric model as a starting point, estimates the model parameters, and then uses the fitted model to predict at unobserved locations. Instead, with SBNNs, one first calibrates the prior distribution of the weights and biases, then finds the posterior distribution of these weights and biases using SGHMC, and then uses this to obtain predictive distributions of the process. We show that an SBNN can be used to model a large variety of spatial processes that could be non-stationary and/or non-Gaussian. This concluding section focusses on the advantages and disadvantages of SBNNs, shedding light on situations where they will likely be useful and where they will not.

\subsection{Calibration requires replicated realisations from the underlying stochastic process}

A notable limitation of SBNNs is that their calibration requires a considerable number of realisations from an underlying stochastic process. In many applications of interest, the spatial statistician only has a single realisation at hand. In such cases, one could calibrate an SBNN to a process model that is easy to simulate from, but it is unclear whether this approach will ultimately lead to any computational or inferential benefit. For example, several methods exist for parameter estimation and prediction with Gaussian processes, and there is little reason why one needs to calibrate SBNNs to Gaussian processes other than for software verification. On the other hand, there are other processes, such as the max-stable process of Section~\ref{sec:CaseStudy2} (and some classes of stochastic partial differential equations), where parameter estimation, prediction, and conditional simulation are notoriously difficult, and yet the process is relatively easy to (unconditionally) simulate from. In such cases, SBNNs are well-positioned to yield computational benefits over the classical approach of fitting and predicting using a parametric model.

When a large quantity of data is available, and where these data can be reasonably treated as replicates from some underlying stochastic process, SBNNs present a natural way forward to modelling and predicting. They place few assumptions on the underlying process and relieve the modeller from having to make a judgement call on which class of models is most suited to their application. Calibration data could be available, for example, in the form of re-analysis data of some geophysical quantity (such as sea-surface temperature). For example, temporally stationary spatio-temporal data could provide the spatial replicates indexed by time.

\subsection{Computing resources}

Both calibrating SBNNs and finding the posterior distribution of their parameters requires considerable computing resources and sophisticated algorithms. However, once these routines have been developed, they are widely  applicable; this leads to several advantages from a computing perspective. First, since SBNNs are process agnostic, calibrating and fitting them to data is likely to require a similar amount of computing resources, irrespective of what the underlying target process is. This is not the case with classical approaches, where the adopted spatial process model and data model largely determine the computational complexity involved when estimating and predicting. Second, while SBNNs can be used in a wide range of settings, the same algorithms for calibration and for making inference can be used. This is in contrast to traditional likelihood-based approaches where the model is application-dependent with parameter spaces of varying sizes, and where complex algorithms are generally needed for complex models. Finally, with SBNNs, model calibration can be done using gradient descent with mini-batches (i.e., a small set of realisations at each optimisation step); this allows one to calibrate SBNNs using a large number of realisations on memory-limited devices.

\subsection{Computational tools used for facilitating calibration}

Our approach to calibration substitutes the 1-Lipschitz function, which is a function required for establishing the Wasserstein distance between two distributions, with a function $\phi_{NN}(\cdot\,;\lambdab)$ that is only approximately 1-Lipschitz. The most straightforward way to remove this approximation, and to ensure 1-Lipschitz continuity of $\phi_{NN}(\cdot\,;\lambdab)$, is to construct a neural network $\phi_{NN}(\cdot\,;\lambdab)$ that is 1-Lipschitz by design. For instance, as 1-Lipschitz functions are closed under composition, a 1-Lipschitz multi-layer perceptron can be constructed by composing 1-Lipschitz affine transformations with 1-Lipschitz activations that are themselves 1-Lipschitz (e.g., hyperbolic tangent functions, ReLUs, and sigmoid functions). Affine transformations can be made 1-Lipschitz by enforcing norm constraints on the weight matrices \citep{anil2019}. This can be implemented using, for example, {spectral normalisation}, which re-scales the weight matrices using their dominant singular values, or {Bj\"ork orthonormalisation}, which projects the weight matrices onto their nearest orthonormal neighbours \citep{ducotterd2022}. These alternative approaches were tried and led to computational difficulties in our setting, and thus were put aside in favour of the more computationally-efficient approach of \citet{gulrajani2017}.

Even if we were successful in establishing the exact Wasserstein distance between two distributions, our approach only aims at matching one selected finite-dimensional distribution of the SBNN to that of the target process. However, when generating samples for \eqref{eq:wasserstein1_dual_MC}, one may also randomly sample the spatial locations (and the number thereof) over which both $Y(\cdot)$ and $\widetilde Y(\cdot)$ are evaluated. One would then be targeting a wide range of finite-dimensional distributions using the SBNN, and not just one; the optimised hyper-parameters $\psib^*$ would then lead to an SBNN $Y(\cdot)$ that approximates the process $\widetilde Y(\cdot)$ in distribution. Note that even if all the finite-dimensional distributions of the two processes are identical, their sample paths may differ; see \citet[Section 2.2]{Lindgren_2012} for discussion and examples. This caveat is unlikely to cause issues in practice, unless the sample paths of $Y(\cdot)$ or $\widetilde{Y}(\cdot)$ are discontinuous.

\subsection{SBNN architecture and model interpretation}

Incorporating an embedding layer and spatially-varying network parameters was needed for our SBNNs to be able to reproduce realistic covariances and covariance non-stationarity/anisotropy. We have yet to explore what spatial processes our SBNN architectures are \emph{not} suitable for. Future work may reveal that our SBNNs are too inflexible in some settings, even with a `prior-per-parameter' scheme, and that further modifications are needed. \Copy{Limitation}{A drawback of our SBNNs is that they are by-and-large uninterpretable: Whereas in classical modelling one obtains estimates or distributions over parameters that typically have a clear interpretation, with an SBNN one only has access to posterior distributions over weights and biases that are related to the output $Y(\cdot)$ in a highly nonlinear fashion. This limitation alone may preclude SBNNs from being used in certain settings where parameter interpretability is paramount. On the other hand, where prediction and uncertainty quantification through predictive variances is the main goal, SBNNs are likely to be of high practical value.}

\subsection{Open questions and future research directions}

We consider our work as a first step in the development of using SBNNs to model spatial stochastic processes. While we have demonstrated their versatility, there are several questions that remain to be answered, of which three stand out. First, it is not clear how to make predictive inferences with SBNNs that have spatially-varying parameters; while there is software available for BNNs, these typically assume that the weights and biases do not have prior distributions that are input-dependent. It is for this reason that we have restricted our analyses in Section~\ref{sec:inference} to stationary processes, and we have not looked at prediction in a real-data setting (where a spatially-varying parameterisation of the SBNN would be important for modelling non-stationarity). Second, we have not explored the use of SBNNs for prediction in scenarios where there are covariates. We consider this issue to be relatively benign, as one may incorporate other variables in a larger model incorporating the SBNN in a relatively straightforward manner when doing MCMC (e.g., by adding a Gibbs step). Third, it would be useful to know what the representational capacity of the SBNN is; what classes of spatial processes are they able to represent and what are their limitations? How does the representational capacity scale with the size of the SBNN? Answers to these questions are needed to make SBNNs a viable instrument for spatial-statistical modelling.

\section{Acknowledgements}
We would like to thank the reviewers and editors for the care they took reading and commenting on the initial submission, leading to many improvements. The authors also thank Rapha{\"e}l Huser for feedback on the manuscript.
AZ-M was partially supported by the Australian Research Council (ARC) Discovery Early Career Research Award DE180100203. Further, this material is based upon work supported by the Air Force Office of Scientific Research under award number FA2386-23-1-4100 (AZ-M and NC).

\bibliography{Bibliography}

\ifarxiv

\newpage
\renewcommand{\thefigure}{S\arabic{figure}}
\setcounter{figure}{0}

~
\vspace{-0.6in}

\begin{center}
  \LARGE{Supplementary Material for ``Spatial Bayesian Neural Networks''}

\vspace{0.1in}
  
\small  Andrew Zammit-Mangion, Michael D. Kaminski, Ba-Hien Tran, Maurizio Filippone, Noel Cressie
\end{center}

\begin{figure}[h!]
    \begin{center}
    \includegraphics{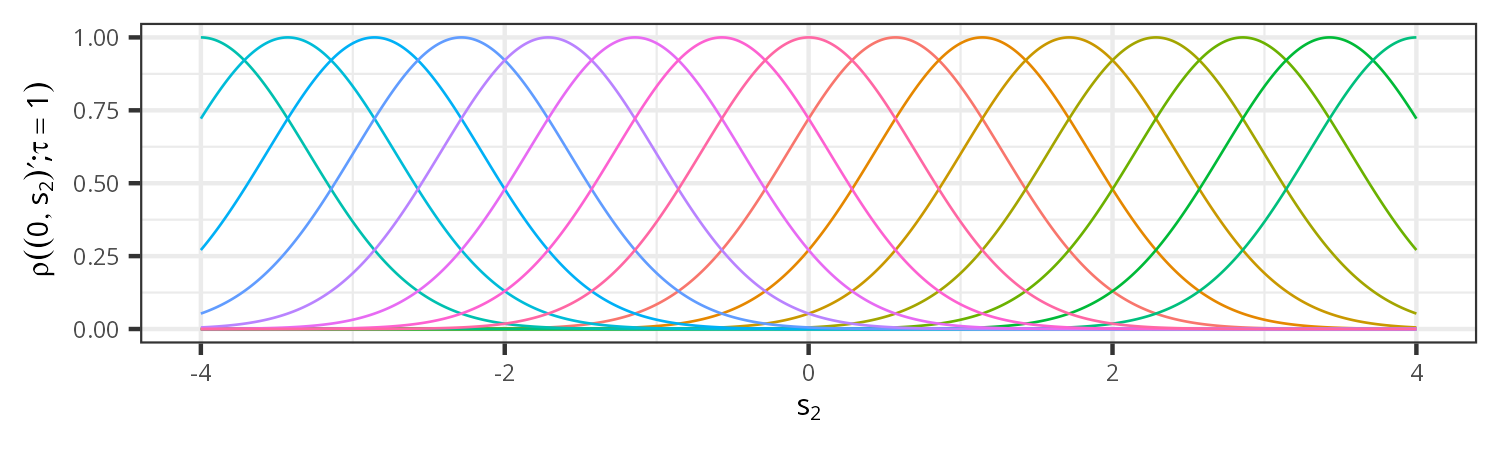}
    \end{center}
     \caption{Evaluations of the embedding-layer basis functions in $\rhob(\cdot\,;\tau = 1)$, which have their centroid at $s_1 = 0$, along an $s_2$-transect at $s_1 = 0$. The different colours denote the different basis functions, of which there are $\sqrt{K} = 15$. \label{fig:basis_functions}}
 \end{figure}

\begin{figure}[h!]
        
    \includegraphics{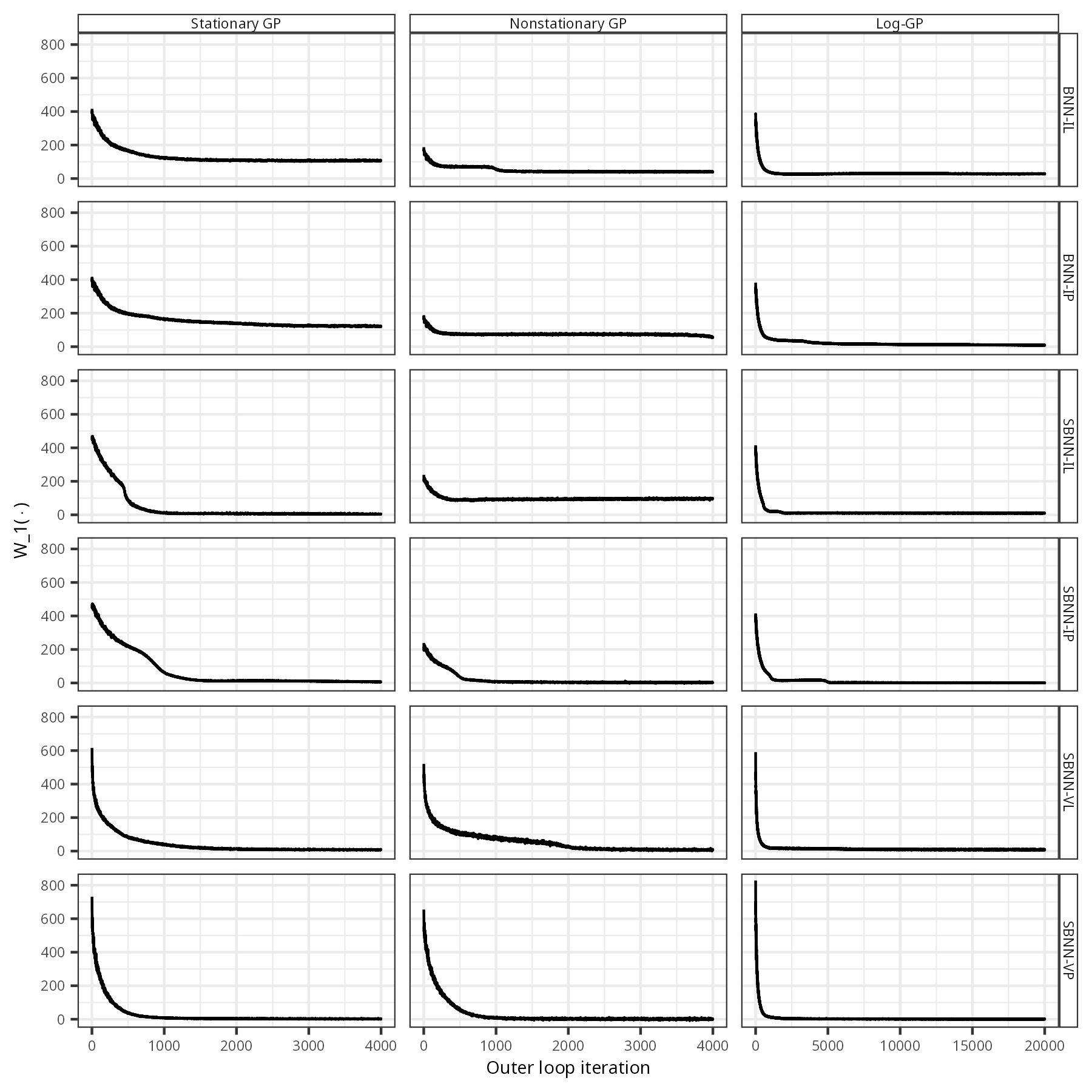}
 
     \caption{Wasserstein distances $W_1(\cdot)$ as a function of outer-loop iteration number when training the six (S)BNN models for the three simulation experiments described in Section~\ref{sec:results}.\label{fig:wass_distances}}
 \end{figure}

\begin{figure}[h!]
        
    \includegraphics{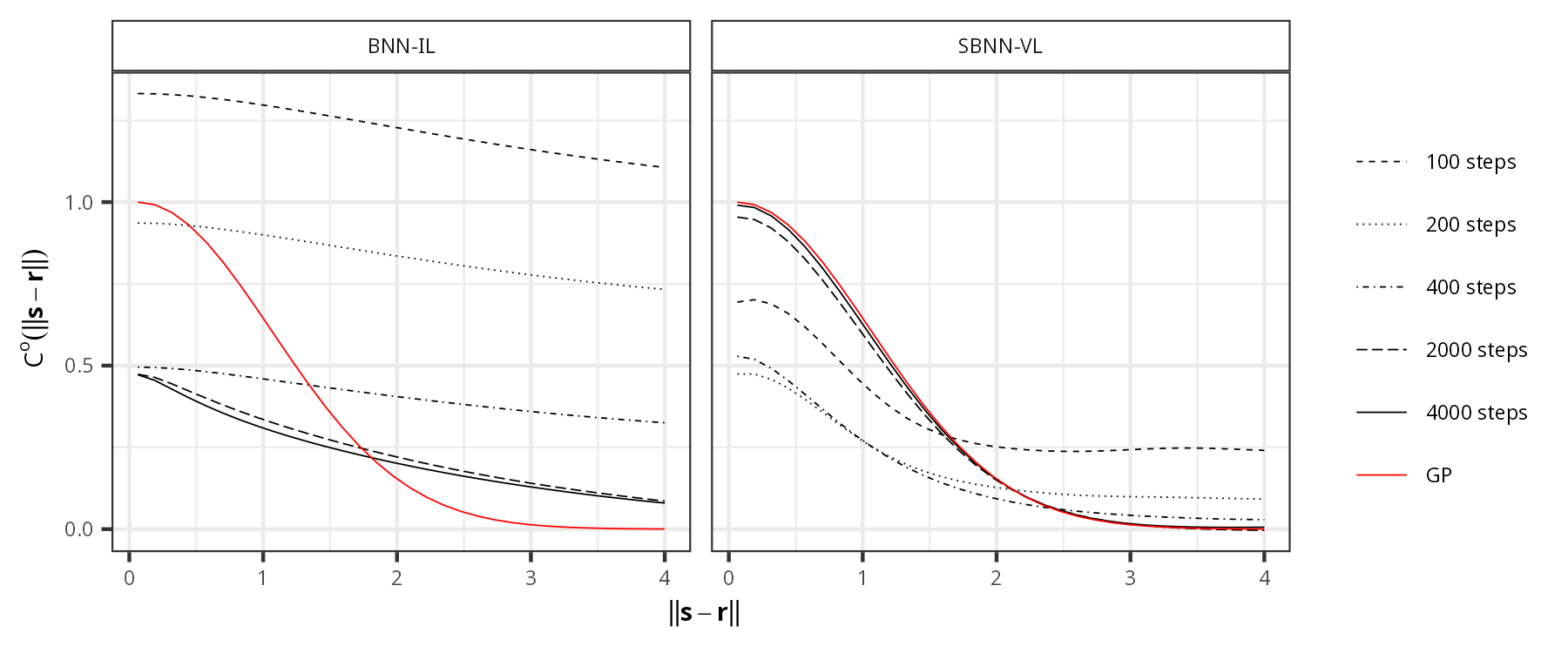}
 
     \caption{Empirical covariogram of the (S)BNN at different stages during the optimisation (different line-styles denote the empirical covariogram after 100, 200, 400, 1000, 2000, and 4000 gradient steps, respectively), and the true covariogram of the target Gaussian process (red). Left: BNN-IL. Right: SBNN-VL. \label{fig:gp_bnn_covs-sv}}
 \end{figure}

\begin{figure}[t!]

    \includegraphics{./figures/Section4_1_GP_cov.png}    
    \includegraphics{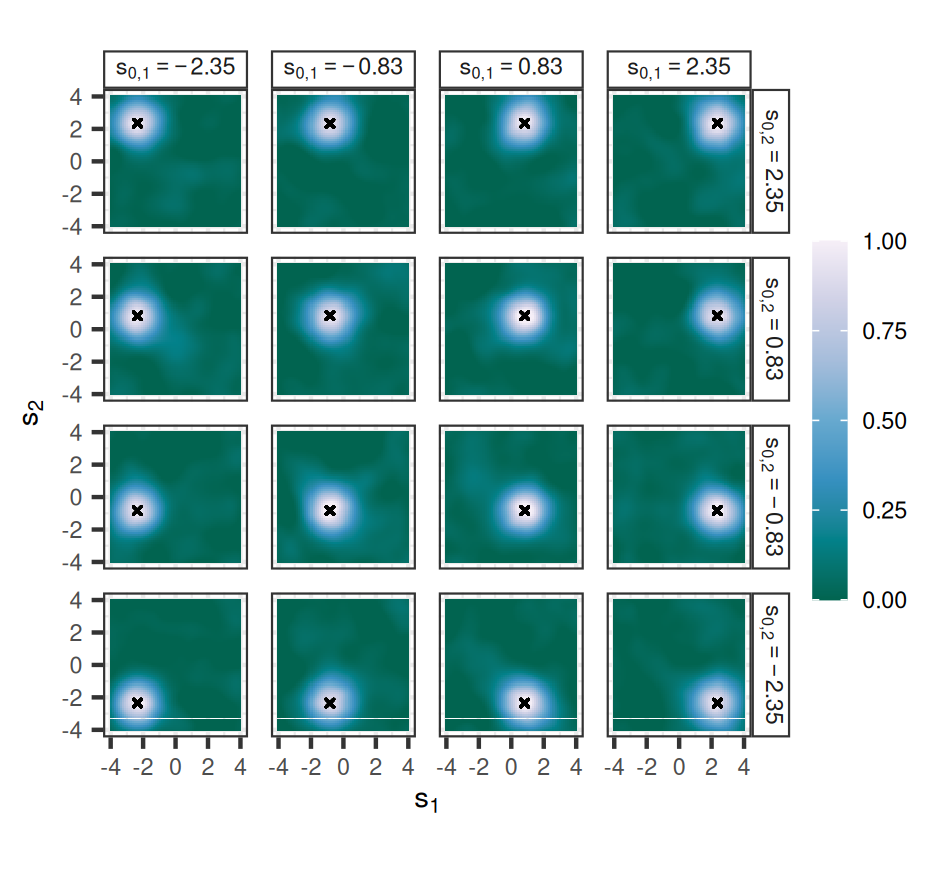}
    \includegraphics{./figures/Section4_1_GP_prior_samples.png}
    \includegraphics{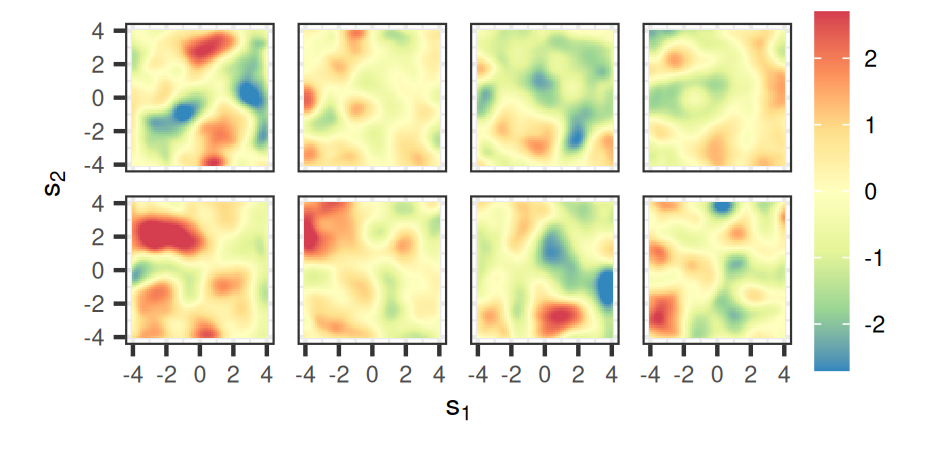}
    
    \caption{(Top-left panel) Covariance between the target process (stationary GP) at 16 grid points (crosses), with coordinates as indicated by the labels at the top and the right of the sub-panels, and the target process on a fine gridding $(64 \times 64)$ of $D = [-4, 4] \times [-4, 4]$. (Bottom-left panel) Eight realisations of the underlying target process on the same fine gridding of $D$. (Right panels) Same as left panels but for the calibrated SBNN-VL. \label{fig:gp_bnn_heatmaps_samples-sv}}

\end{figure}

 \newcommand{\contourcaption}{Contour levels (from the outer to the inner contours) correspond to 0.05, 0.10, 0.15, 0.20, 0.25, 0.30, 0.35, 0.40 and 0.45.
}
 \begin{figure}[h!]
\begin{center}         
     \includegraphics[width=0.9\linewidth]{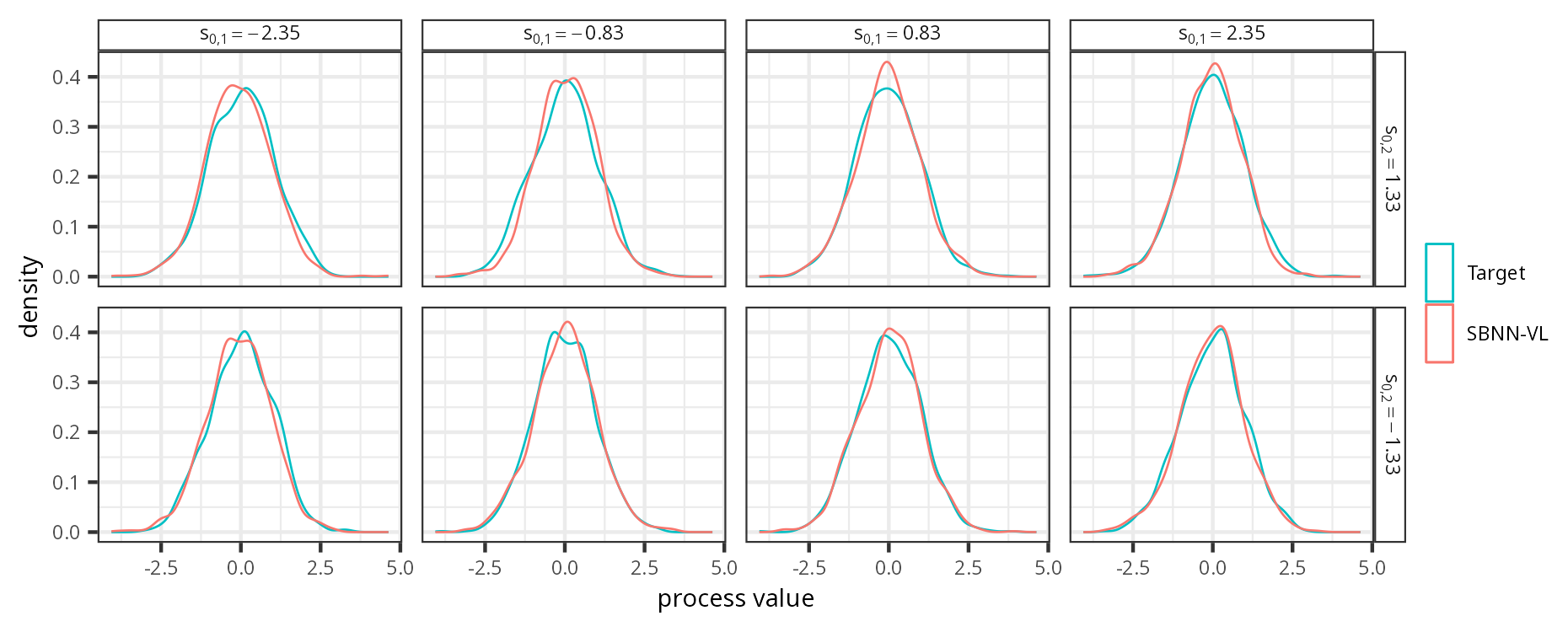} \\
     \includegraphics[width=0.9\linewidth]{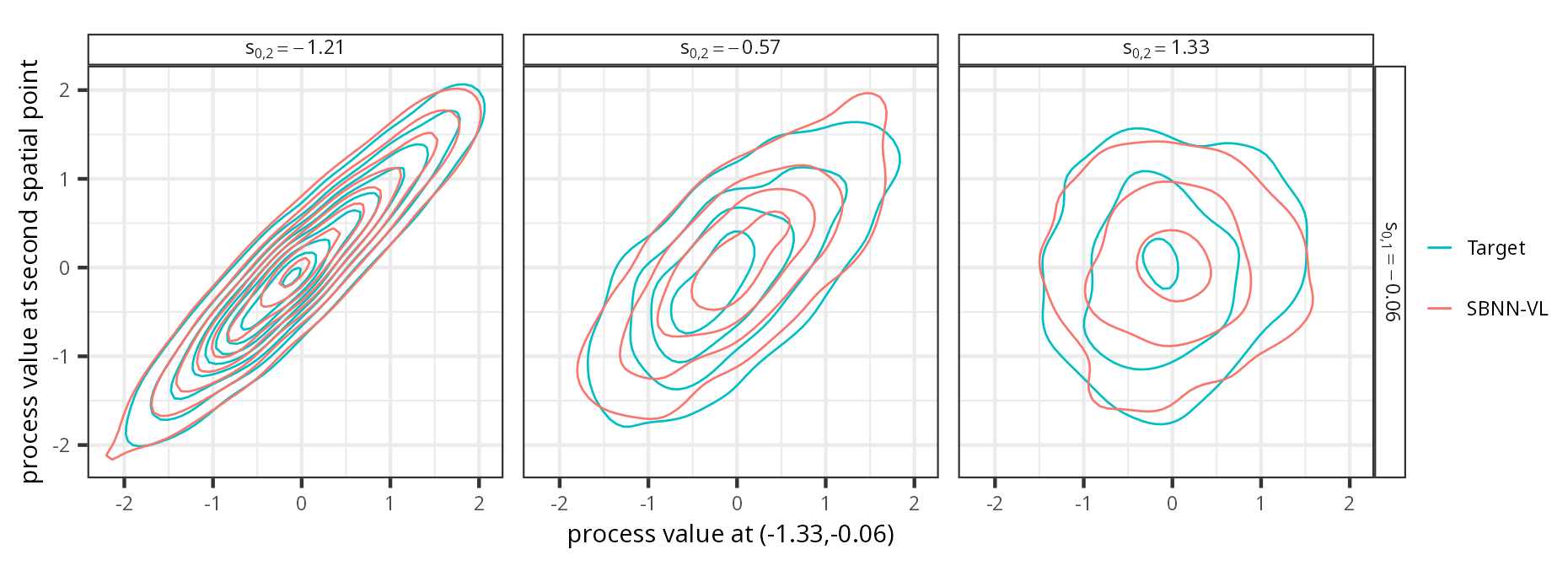} 
 \end{center}
     \caption{Kernel density plots from 1000 samples of the SBNN-VL and the target process (stationary GP). (Top panel) Univariate densities of the two processes at eight spatial locations arranged on a $2 \times 4$ grid in $D$ (with coordinates as indicated by the labels of the sub-panels). (Bottom panel) Overlayed bivariate densities of the two processes at ${\tilde\svec_0  = (-1.33, -0.06)'}$ and three other locations in $D$ (with coordinates as indicated by the labels of the sub-panels). \input{./figures/Section4_3_SBNN-VP_contour_levels.tex} \label{fig:prior_densities_stationaryGP-sv}}
 \end{figure}

\begin{figure}[t!]
        \includegraphics{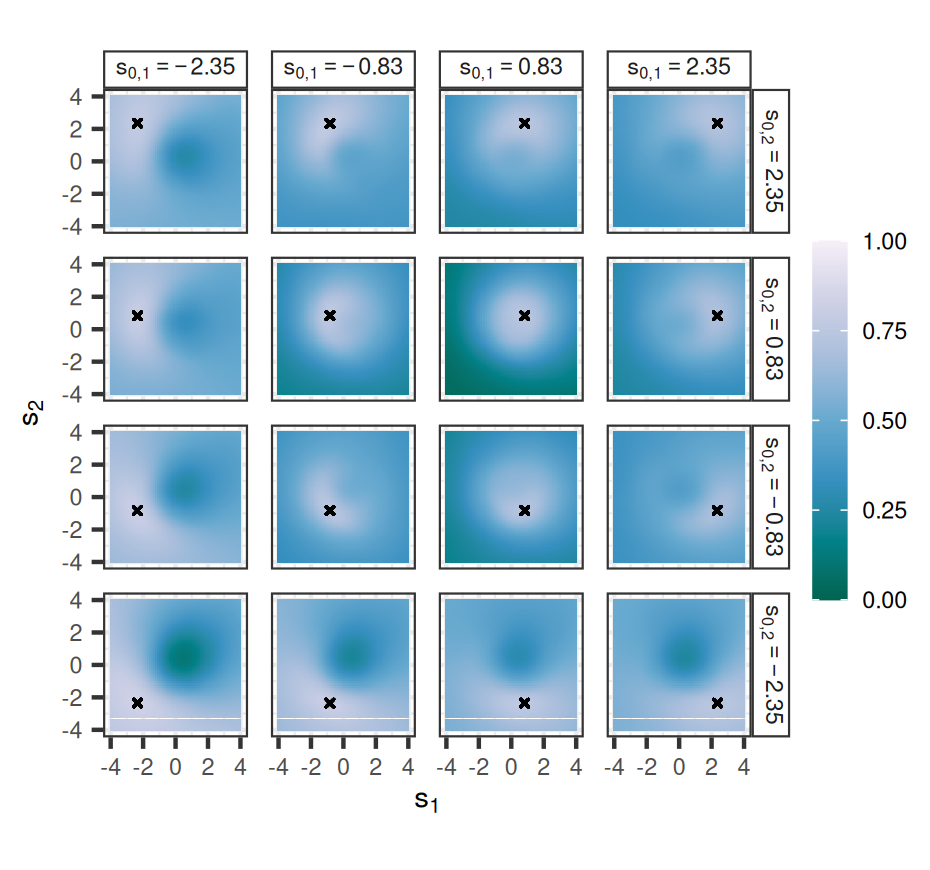}
        \includegraphics{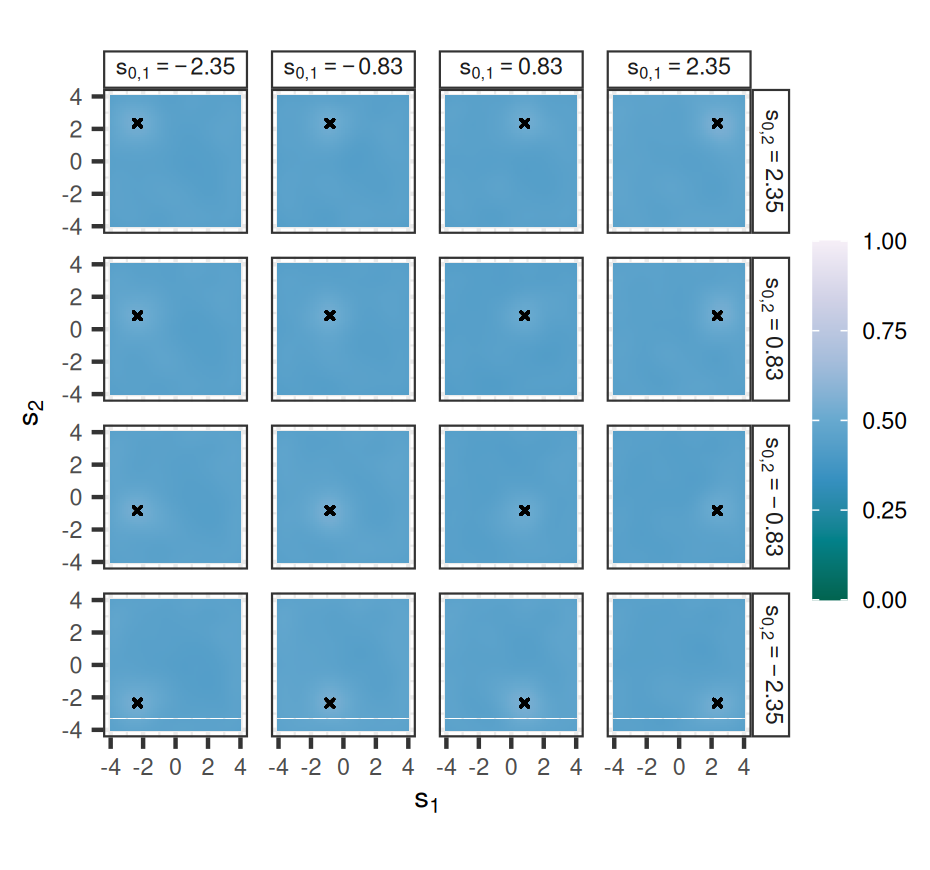}

        \caption{(Left panel) Covariance between the calibrated BNN-IL process at 16 grid points (crosses) arranged on a 4 × 4 grid (see Figure~\ref{fig:gp_bnn_heatmaps_samples}), and the process on a fine gridding $(64 \times 64)$ of $D = [-4, 4] \times [-4,  4]$. (Right panel) Same as left panel, but for the calibrated SBNN-IL.}
    \label{fig:nonstat_cov2} 
\end{figure}

\newcommand{\contourcaption}{Contour levels (from the outer to the inner contours) correspond to 0.20, 0.40, 0.60, 0.80, 1.00, 1.20, 1.40, 1.60, 1.80 and 2.00.
}

\begin{figure}[t!]
        
    \includegraphics{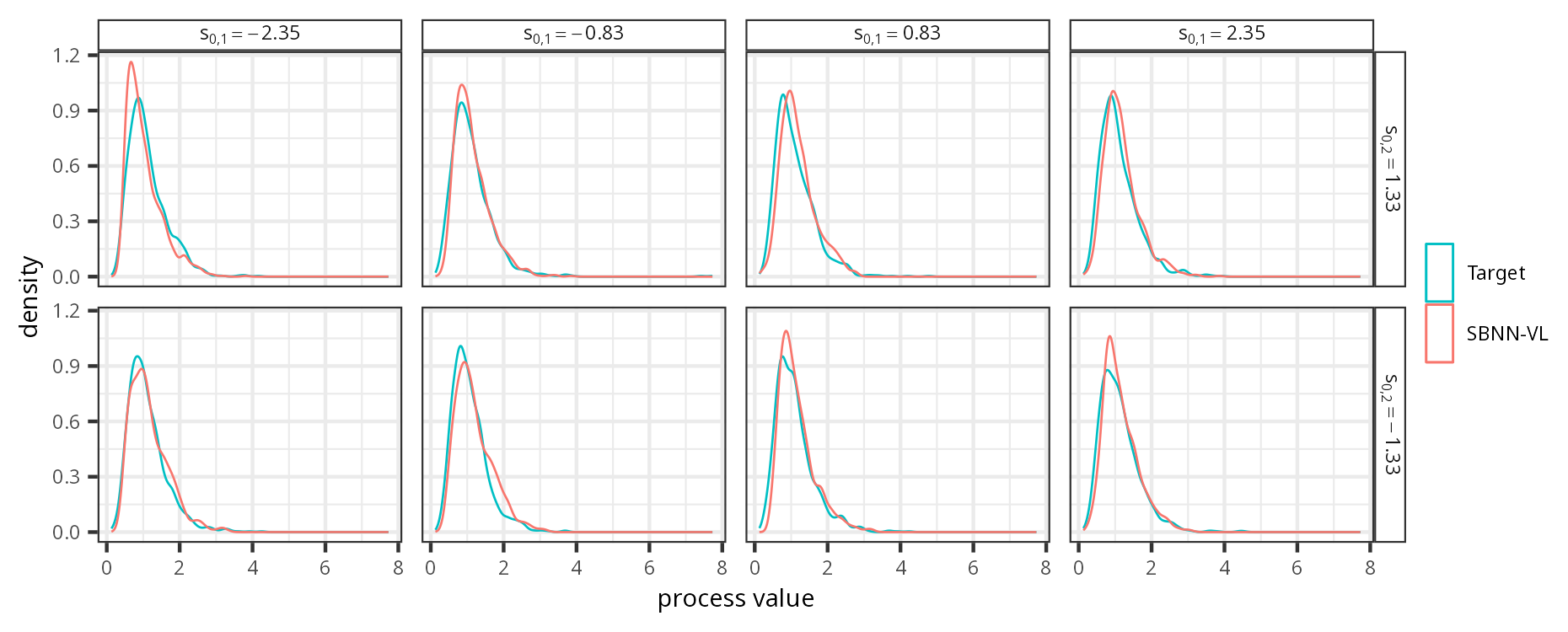} \\
    \includegraphics{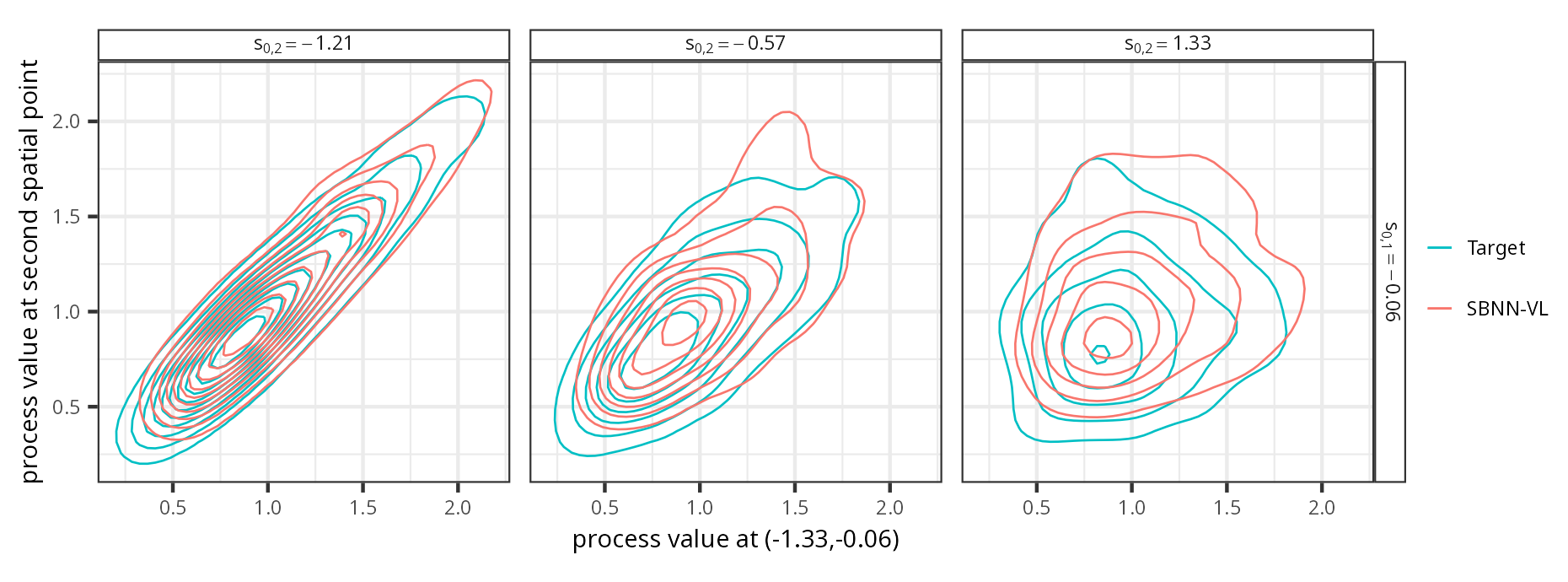} 

    \caption{Kernel density plots from 1000 samples of the SBNN-VL and the target process (stationary lognormal spatial process). (Top panel) Univariate densities of the two processes at eight spatial locations arranged on a $2 \times 4$ grid in $D$ (with coordinates as indicated by the labels of the sub-panels). (Bottom panel) Overlayed bivariate densities of the two processes at ${\tilde\svec_0 = (-1.33, -0.06)'}$ and three other locations in $D$ (with coordinates as indicated by the labels of the sub-panels). \input{./figures/Section4_3_SBNN-VP_contour_levels.tex} \label{fig:prior_densities_lognormalGP-sv}}
  \end{figure}

\newcommand{\contourcaption}{Contour levels (from the outer to the inner contours) correspond to 0.20, 0.40, 0.60, 0.80, 1.00, 1.20, 1.40, 1.60, 1.80 and 2.00.
}

\begin{figure}[t!]
        
    \includegraphics{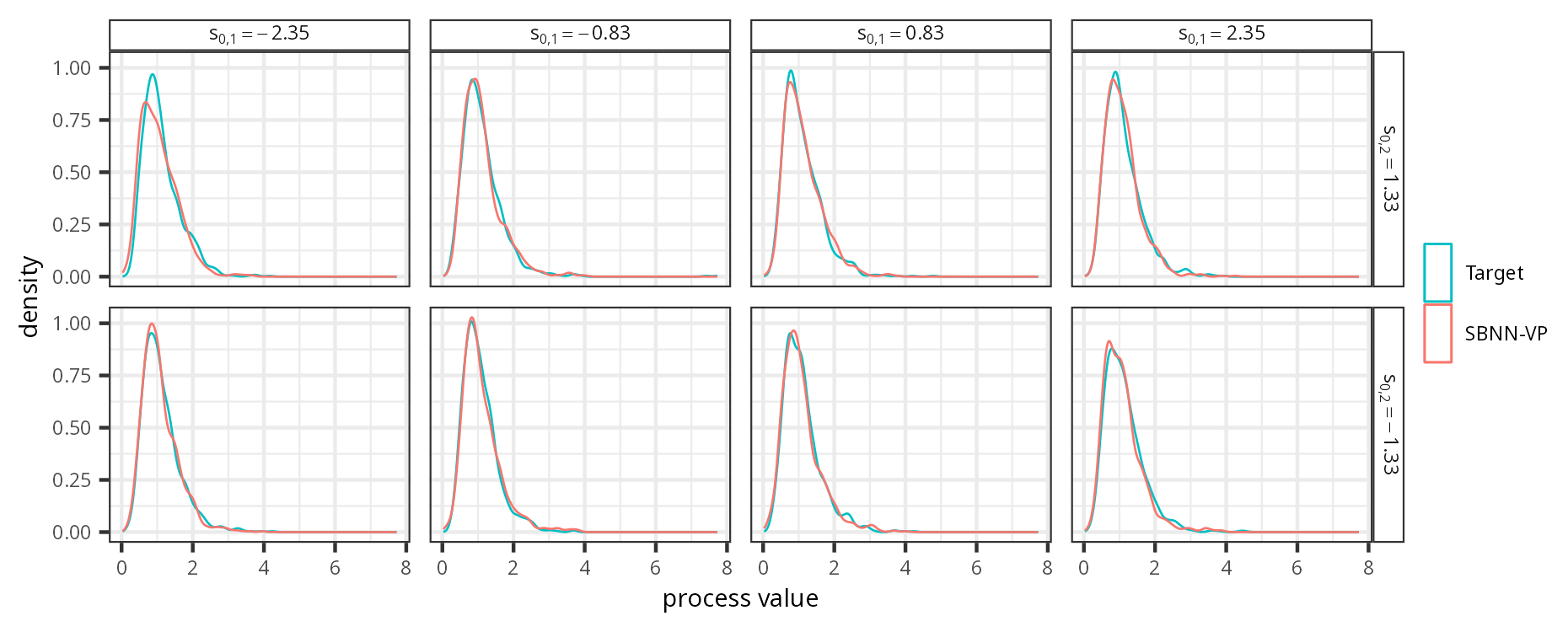} \\
    \includegraphics{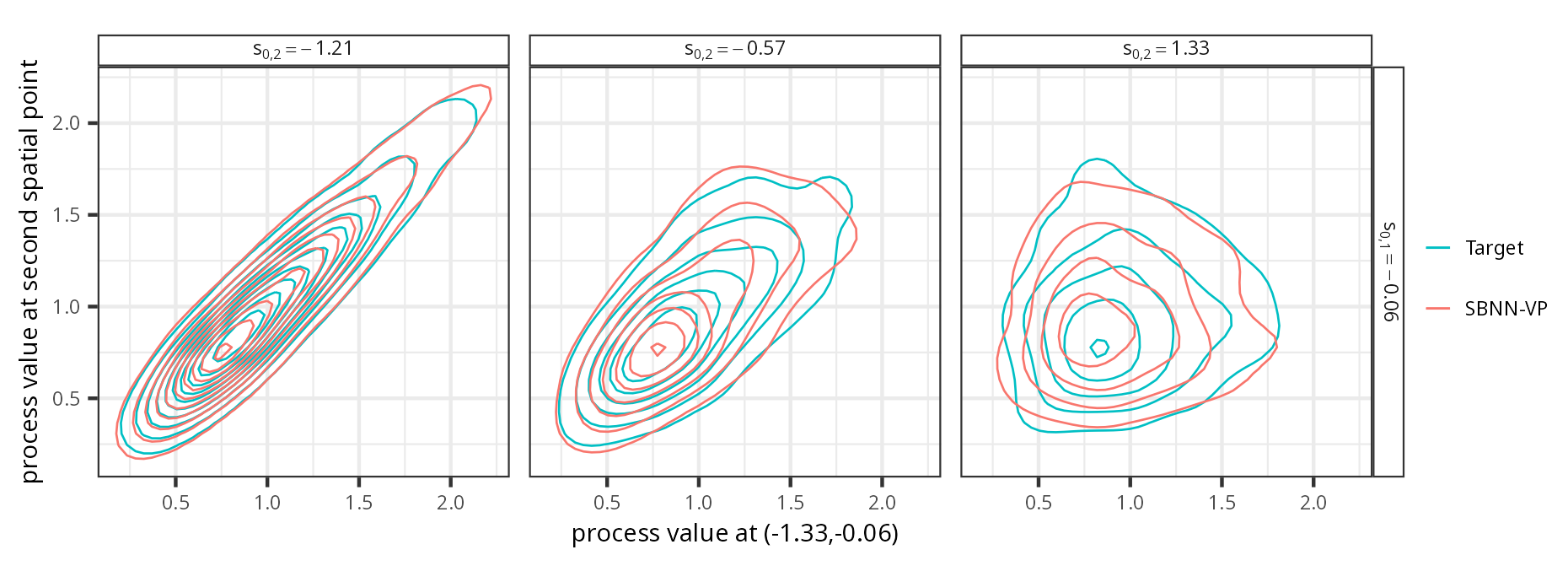} 

    \caption{Kernel density plots from 1000 samples of the SBNN-VP and the target process (stationary lognormal spatial process). (Top panel) Univariate densities of the two processes at eight spatial locations arranged on a $2 \times 4$ grid in $D$ (with coordinates as indicated by the labels of the sub-panels). (Bottom panel) Overlayed bivariate densities of the two processes at ${\tilde\svec_0 = (-1.33, -0.06)'}$ and three other locations in $D$ (with coordinates as indicated by the labels of the sub-panels). \input{./figures/Section4_3_SBNN-VP_contour_levels.tex} \label{fig:prior_densities_lognormalGP-svp}}
  \end{figure}

  \begin{figure}[t!]
        
 \hspace{-0.5in}    \includegraphics{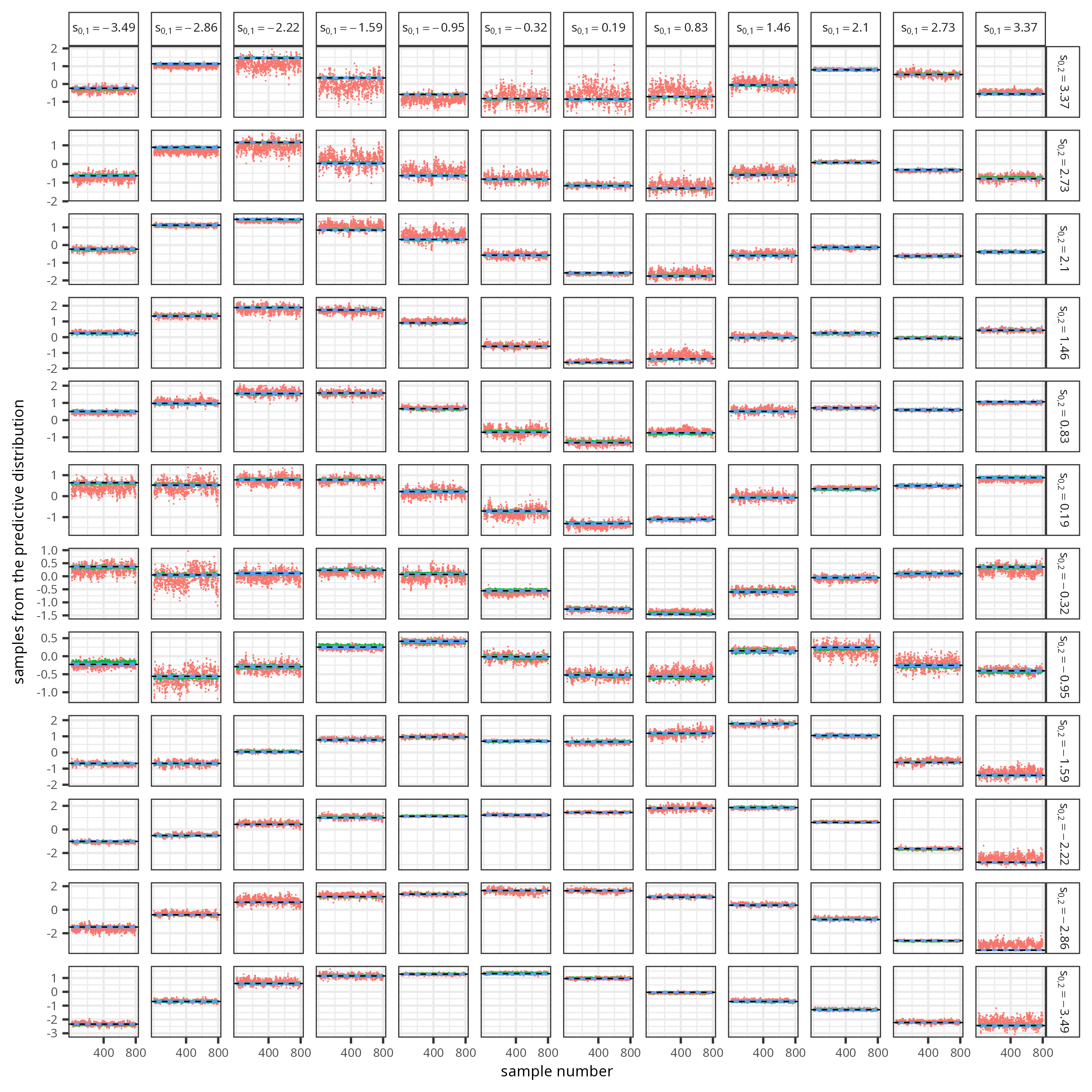} \\

    \caption{Samples from the predictive distribution of $Y(\svec)$, for $\svec$ in a $12\times 12$ gridding of $D$, obtained by SGHMC on data of size $m = 100$ (red dots), $m = 1,000$ (green dots) and $m = 5,000$ (blue dots) using the calibrated SBNN-IP in Section~\ref{sec:CaseStudy1}. The 800 samples comprise 200 samples drawn from four separate chains concatenated together. The true values generated from the stationary GP $\widetilde{Y}(\cdot)$ are denoted by the horizontal black dashed lines. Note that the samples corresponding to $m \ge 1000$ are barely visible because they concentrate around the true value. \label{fig:MCMC_traces_SBNN-IL}}
  \end{figure}

\fi

\end{document}